\documentclass{article}

\usepackage{microtype}
\usepackage{graphicx}
\usepackage{subcaption}
\usepackage{booktabs} 

\usepackage{multirow}
\usepackage{amssymb}
\usepackage{amsmath}
\usepackage{wrapfig}

\usepackage{hyperref}

\usepackage[preprint]{icml2026}

\usepackage{mathtools}
\usepackage{amsthm}
\usepackage{xcolor}
\usepackage{colortbl}
\usepackage{color}

\usepackage[capitalize,noabbrev]{cleveref}

\theoremstyle{plain}
\newtheorem{theorem}{Theorem}[section]

\theoremstyle{definition}

\newtheorem{assumption}[theorem]{Assumption}
\theoremstyle{remark}

\usepackage[textsize=tiny]{todonotes}

\icmltitlerunning{Chain-based Distillation for Effective Initialization of Variable-Sized Small Language Models}

\begin{document}

\twocolumn[
  \icmltitle{Chain-based Distillation for Effective Initialization of Variable-Sized \\
   Small Language Models}

  \begin{icmlauthorlist}
    \icmlauthor{Boyu Shi}{seu1,seu2}
    \icmlauthor{YiCheng Jiang}{seu1,seu2}
    \icmlauthor{Chang Liu}{seu1,seu2}
    \icmlauthor{Qiufeng Wang}{seu1,seu2}
    \icmlauthor{Xu Yang\textsuperscript{*}}{seu1,seu2}
    \icmlauthor{Xin Geng\textsuperscript{*}}{seu1,seu2}
  \end{icmlauthorlist}

  \icmlaffiliation{seu1}{School of Computer Science and Engineering, Southeast University, Nanjing, China}
  \icmlaffiliation{seu2}{Key Laboratory of New Generation Artificial Intelligence Technology and Its Interdisciplinary Applications (Southeast University), Ministry of Education, China}

  \icmlcorrespondingauthor{Xu Yang}{xuyang\_palm@seu.edu.cn}
  \icmlcorrespondingauthor{Xin Geng}{xgeng@seu.edu.cn}

  \icmlkeywords{Machine Learning, ICML}

  \vskip 0.3in
]



\printAffiliationsAndNotice{$^*$ Co-corresponding authors.}

\begin{abstract}
  Large language models (LLMs) achieve strong performance but remain costly to deploy in resource-constrained settings. Training small language models (SLMs) from scratch is computationally expensive, while conventional knowledge distillation requires repeated access to large teachers for different target sizes, leading to poor scalability.
  To solve these problems, we propose \textbf{Chain-based Distillation (CBD)}, a scalable paradigm for efficiently initializing variable-sized language models. A sparse and limited sequence of intermediate models (called anchors) is constructed via stepwise distillation, forming a distillation chain that progressively transfers knowledge from the source LLMs. To support heterogeneous settings, we introduce \emph{bridge distillation} for cross-architecture and cross-vocabulary transfer. Models of variable sizes are initialized via parameter interpolation between adjacent anchors, eliminating repeated large teacher inference.
  Experiments show that the proposed method substantially improves efficiency and downstream performance. A 138M-parameter SLM without recovery pre-training, outperforms scratch-trained models on a 10B-token corpus on the specific task. 
  CBD also demonstrates versatility in heterogeneous settings for initialize models with different architectures and vocabularies.
\end{abstract}

\section{Introduction}
Large language models (LLMs) \citep{touvron2023llama2,team2024qwen2,yang2025qwen3} have demonstrated remarkable generalization capabilities across diverse domains \citep{cui2023chatlaw,chen2023disc}. However, their immense parameter scales pose significant challenges for deployment in resource-constrained environments, such as mobile devices or embedded systems \citep{jia2025scaling}. The variety of real-world applications necessitates a spectrum of smaller language models (SLMs) tailored to specific hardware constraints and latency requirements.

\begin{figure}
    \centering
    \includegraphics[width=0.92\linewidth]{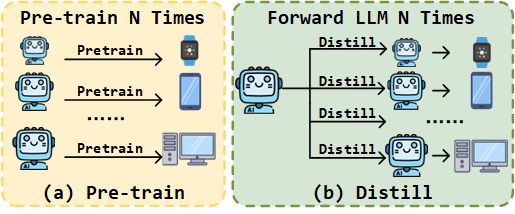}
    \caption{Conceptual comparison between traditional pre-training from scratch, direct knowledge distillation for constructing $N$ variable-sized SLMs. Both methods have repetitive, high-overhead distillation or training processes.}
    \label{fig:motivation}
    \vspace{-0.6cm}
\end{figure}

To efficiently populate this model spectrum, the most intuitive approach is to construct a dense set of SLMs covering various sizes. However, traditional paradigms struggle to balance performance, efficiency, and architectural flexibility. As illustrated in \cref{fig:motivation}a, training $N$ models from scratch \citep{devlin2019bert,singer2024h2o} incurs a prohibitive $\mathcal{O}(N)$ computational cost and often yields suboptimal results for smaller scales due to limited model capacity. While knowledge distillation (KD) \citep{yao2023minirbt,hu2025large} can enhance SLM performance, it requires repeated, computationally expensive access to the teacher LLM for every target size (\cref{fig:motivation}b). Furthermore, in the LLM era, the \textbf{massive capacity gap} between the hundred-billion-parameter teacher and the million-parameter student often creates a "distillation bottleneck," where the student fails to capture the complex distribution of the teacher, resulting in significant performance degradation \citep{muralidharan2024compact,zhang2023towards}.


\begin{figure*}[t]
  \centering
  \includegraphics[width=0.92\linewidth]{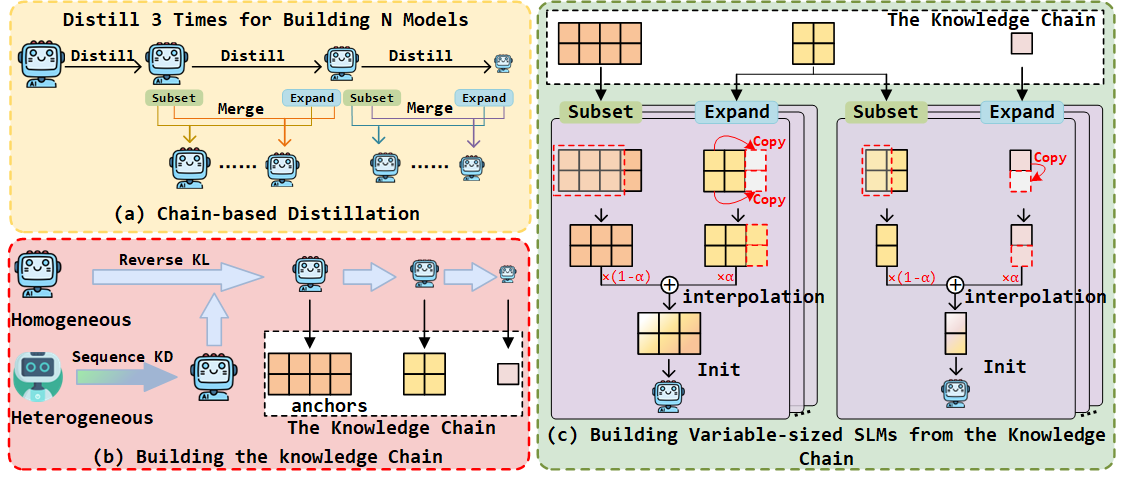}
  \caption{The overall framework of the proposed CBD. (a) CBD achieves a significant reduction in cumulative computational cost by leveraging a structured knowledge chain. (b) \textbf{Knowledge chain construction}: propagating knowledge from the source LLM to a sequence of anchors via stepwise distillation (homogeneous case) or bridge distillation (heterogeneous case). (c) \textbf{Variable-sized SLM building}: rapid initialization of target SLMs through parameter interpolation between adjacent anchors.}
  \label{fig:overall}
  \vspace{-0.6cm}
\end{figure*}

Beyond capacity concerns, a critical yet underexplored challenge is \textbf{architectural and structural heterogeneity}. In many practical scenarios, high-performance teacher models (e.g., Qwen3 \cite{yang2025qwen3}) and target deployment models (e.g., legacy GPT-based architectures \cite{radford2019language}) exhibit significant \textbf{architectural divergence}, including mismatched vocabulary sizes, hidden dimensions, and structural configurations. This mismatch creates a structural barrier that standard homogeneous KD cannot bridge. 
Furthermore, heterogeneous distillation is not merely an optional feature but a necessity to leverage state-of-the-art LLM knowledge for hardware-specific optimized architectures. 
Therefore, a scalable method that can efficiently generate high-quality, well-initialized models across variable scales and heterogeneous architectures is highly desirable for both industrial deployment.

To achieve this, we propose \textbf{Chain-based-Distillation (CBD)}, a novel paradigm for efficiently constructing variable-sized SLMs through a knowledge chain. First, CBD establishes a sparse sequence of intermediate models, termed \textit{anchors}, via \textbf{stepwise distillation}. Unlike traditional KD, each anchor in CBD is distilled from its immediate, slightly larger predecessor, effectively decomposing the massive capacity gap between a giant teacher and a tiny student into manageable steps, as shown in \cref{fig:overall}a. 
To handle the aforementioned heterogeneity, we introduce a \textbf{bridging mechanism}: the source LLM is first distilled into a structurally aligned proxy model that matches the target chain's architecture. This proxy serves as the "anchor zero," ensuring that subsequent distillation steps focus on knowledge compression rather than resolving alignment conflicts (e.g., vocabulary mismatches), as in \cref{fig:overall}b.

Finally, to achieve $\mathcal{O}(1)$ efficiency for $N$ models, we employ a parameter interpolation strategy between adjacent anchors, as shown in \cref{fig:overall}c. By subsetting and expanding weights with a scaling coefficient $\alpha$, CBD can rapidly generate well-initialized SLMs of variable size within the anchor range without further access to the source LLM, significantly reducing computational overhead while maintaining high initialization quality.

Extensive experiments demonstrate that CBD dramatically accelerates convergence across multiple scales. For instance, a 138M SLM initialized via CBD, even without additional training, outperforms scratch-trained models on 10B tokens. Furthermore, CBD achieves up to a \textbf{200$\times$} speedup in reaching target performance levels compared to training from scratch. Our theoretical analysis confirms that stepwise distillation reduces approximation errors and improves convergence rates, providing a robust foundation for building scalable and efficient SLMs.

\section{Related Work}
\subsection{LLM Knowledge Distillation}
Knowledge distillation \citep{hinton2015distilling} transfers knowledge from a large teacher model to a smaller student, enabling efficiency without sacrificing performance. In LLMs, methods are generally categorized as \textit{black-box} or \textit{white-box} distillation based on teacher accessibility.
\textbf{Black-box distillation} targets closed-source models where internal states are unavailable. Student models are trained on outputs generated by the teacher, such as chain-of-thought (CoT) reasoning \citep{Wang2022SelfConsistencyIC,hsieh2023distilling,wang2023scott} or in-context learning (ICL) examples \citep{Li2024MENDMD,Duan2024InContextLD}.
\textbf{White-box distillation} \citep{gu2023minillm} uses open-source teachers. Beyond aligning output logits \citep{agarwal2024policy}, students can learn from intermediate features \citep{liang2023less}, enabling fine-grained transfer.

\subsection{Learngene and Chain-of-Model}
Learngene \citep{wang2022learngene} and its extensions \citep{wang2023learngene, shi2024building, xia2024transformer} focus on identifying and transferring compact, reusable subnetworks from large pretrained models to initialize smaller targets. While effective for knowledge reuse, they primarily rely on subnetwork extraction and subsequent structural expansion. Chain-of-Model (CoM) \citep{wangchain}, alternatively, employs a single backbone to support multi-scale inference through dynamic activation.

In contrast, our CBD introduces a distinct \textbf{stepwise distillation chain}. Unlike Learngene's subnetwork-based initialization or CoM's dynamic inference, CBD constructs a sequence of intermediate models that progressively transfer knowledge across scales. This distillation-driven chain enables flexible initialization of models at variable sizes via parameter interpolation, effectively bypassing the need for repeated, high-cost teacher LLMs access while ensuring scalable model generation.

\section{Methodology}

To facilitate the rapid construction of Small Language Models (SLMs) across a continuous spectrum of resource constraints, we propose \textbf{Chain-based Distillation (CBD)}. The framework is illustrated in \cref{fig:overall}.

\subsection{Building the Knowledge Chain}
The knowledge chain is defined as a sequence of pre-trained language models, termed \textit{anchors}, $\{A_1, A_2, \dots, A_M\}$, arranged in descending order of parameter scale. These anchors serve as discrete "milestones" of knowledge density within the chain.
Once the structure of the knowledge chain is established, it is essential to propagate effective knowledge from the source LLM to form a coherent anchors. We propose two distillation strategies to accommodate both homogeneous and heterogeneous source LLMs, as illustrated in \cref{fig:overall}a.

In this work, we instantiate the chain using three representative families to demonstrate architectural breadth: 
(1) \textbf{GPT2-based chain}: \{GPT2-L (762M), GPT2-M (345M), GPT2-B (117M)\}; 
(2) \textbf{Pythia-based chain}: \{Pythia-1b, 410m, 160m, 70m\}; 
(3) \textbf{Qwen3-based chain}: \{Qwen3-1.7B, 0.6B\}. 
For brevity, we take the GPT2-based chain as the primary reference in the following sections. The selection of these anchors is guided by two principles: \textit{structural consistency}, ensuring a smooth transition of hidden dimensions and layers to facilitate interpolation; and \textit{practical accessibility}, utilizing publicly available checkpoints to bypass the prohibitive cost of training-from-scratch.

\subsection{Knowledge Propagation: Stepwise Distillation}
The core of CBD lies in propagating knowledge from a high-capacity source LLM through the anchor sequence. We categorize this into two scenarios based on the structural alignment between the source and the target chain.

\textbf{Scenario 1: Homogeneous Distillation.} 
When the source LLM (e.g., GPT2-XL, 1.5B) shares the same architecture and vocabulary as the anchors, we employ a \textit{stepwise distillation} strategy. Instead of distilling the source model into each anchor independently, knowledge is transferred sequentially: $LLM \to A_1 \to A_2 \dots \to A_M$. For each step, the preceding larger model acts as the teacher, and the subsequent smaller model as the student. We optimize the student using the reverse Kullback-Leibler (KL) divergence on a general supervised fine-tuning (SFT) dataset $\mathcal{D}$:
\begin{equation}
\mathcal{L}_{\text{distill}} = \mathbb{E}_{x \sim \mathcal{D}} \left[ \sum_{t} P_{S}(x_t \mid x_{<t}) \log \frac{P_{S}(x_t \mid x_{<t})}{P_{T}(x_t \mid x_{<t})} \right]
\end{equation}
where $P_T$ and $P_S$ represent the token probability distributions of the teacher and student, respectively. This stepwise approach decomposes the massive capacity gap into a series of manageable transitions, ensuring stable convergence and superior knowledge retention.

\textbf{Scenario 2: Heterogeneous Distillation \& Bridging.}
To leverage knowledge from state-of-the-art but structurally different LLMs (e.g., Llama3-8B \cite{dubey2024llama} or Qwen3-4B \cite{yang2025qwen3}), we must address two critical challenges: (1) the massive parameter gap between the source scale and the student, and (2) the alignment noise arising from mismatched vocabularies and hidden dimensions. To mitigate these issues, we introduce a bridge distillation mechanism that employs an intermediate model (e.g., GPT2-XL) mirroring the chain's architecture.
This bridge model alleviates the parameter gap through staged compression and eliminates alignment noise by providing architectural consistency. Specifically, we employ Sequence-level Knowledge Distillation (SeqKD) to bypass vocabulary mismatches:
\begin{equation}
\mathcal{L}_{\text{seqkd}} = \mathbb{E}_{(x, \hat{y}) \sim \mathrm{LM}_{\text{source}}}\left[-\log P_{\text{bridge}}(\hat{y} \mid x)\right]
\end{equation}
where $\hat{y}$ is the output sequence sampled from the source LLM. Once the bridge model is optimized, the bridge model serves as a homogeneous starting point, effectively decoupling architectural alignment from knowledge compression.

\subsection{Theoretical Underpinning of the Knowledge Chain}
The superiority of stepwise distillation over direct distillation is grounded in the reduction of both statistical and approximation errors. We formalize this via the following theorems:

\begin{theorem}[Homogeneous stepwise distillation]
  \label{thm:homo}
  Consider a $k$-level distillation chain 
  $\mathcal{H}_0 \to \mathcal{H}_1 \to \cdots \to \mathcal{H}_k$, 
  where all $\mathcal{H}_i$ belong to the same hypothesis family. 
  The generalization error of the final student satisfies
  \[
  \begin{aligned}
  R(\hat f_{\mathcal{H}_k}) - R(f^\star)
  \;\le\;&
  \sum_{i=1}^k 
  O\!\left(\tfrac{C_{\mathcal{H}_i}}{n^{\rho_{\mathcal{H}_i \leftarrow \mathcal{H}_{i-1}}}}\right) \\
  &\;+\;
  \sum_{i=1}^k 
  \varepsilon_{\mathcal{H}_i \leftarrow \mathcal{H}_{i-1}}.
  \end{aligned}
  \]
\end{theorem}
  
\begin{theorem}[Heterogeneous stepwise distillation]
  \label{thm:hetero}
  Consider a $k$-level distillation chain 
  $\mathcal{G} \to \mathcal{H}_1 \to \cdots \to \mathcal{H}_k \to \mathcal{S}$, 
  where $\mathcal{G}$ and $\mathcal{S}$ may belong to different hypothesis families. 
  The generalization error of the final student satisfies
  \[
  \begin{aligned}
  R(\hat f_{\mathcal{S}}) - R(f^\star)
  \;\le\;&
  \sum_{i=1}^k 
  O\!\left(\tfrac{C_{\mathcal{H}_i}}{n^{\rho_{\mathcal{H}_i \leftarrow \mathcal{H}_{i-1}}}}\right) \\
  &\;+\;
  O\!\left(\tfrac{C_{\mathcal{S}}}{n^{\rho_{\mathcal{S} \leftarrow \mathcal{H}_k}}}\right) \\
  &\;+\;
  \sum_{i=1}^k 
  \varepsilon_{\mathcal{H}_i \leftarrow \mathcal{H}_{i-1}}
  \;+\;
  \varepsilon_{\mathcal{S} \leftarrow \mathcal{H}_k}.
  \end{aligned}
  \]
\end{theorem}

These bounds suggest that by breaking down a large capacity gap into $k$ smaller steps, the convergence rate is stabilized and the cumulative approximation error $\varepsilon$ is minimized compared to a single-step jump. Detailed proofs are provided in \cref{app:theoretical}.

\subsection{Rapid Building of Variable-Sized Models}
A key utility of the CBD framework is the ability to build high-quality initializations for variable model sizes within the anchor range. Given a target SLM with size $S_{target}$, we identify the adjacent anchors $A_{small}$ and $A_{large}$ such that $Size(A_{small}) < S_{target} < Size(A_{large})$. 

We propose a \textbf{parameter interpolation} strategy (\cref{fig:overall}c). Specifically, for structural parameters (layers, heads, dimensions), we conduct two structural transformations:
\begin{itemize}
    \item \textbf{Expansion}: Parameters of $A_{small}$ are expanded via layer replication or zero-padding.
    \item \textbf{Subset}: Parameters of $A_{large}$ are subsetted via structural pruning.
\end{itemize}
The target model $\Theta_{target}$ is then initialized as:
\begin{equation}
\Theta_{target} = \alpha \cdot \text{Trans}(\Theta_{small}) + (1-\alpha) \cdot \text{Trans}(\Theta_{large})
\end{equation}
where $\text{Trans}(\cdot)$ denotes the structural transformation, and $\alpha$ is a scaling coefficient determined by the relative distance in the parameter space. Note that, we assign $\alpha$ to the smaller anchor.
This provides a "warm-start" for the target model, significantly reducing the tokens required for convergenc and improving the perfromance with the knowledge propagated from the source LLMs.

\section{Experiments}
\label{sec:ex}
In this section, we conduct a comprehensive evaluation of the proposed \textbf{Chain-based Distillation (CBD)} framework. We demonstrate its effectiveness across multiple source LLMs, target architectures, and diverse downstream tasks.

\textbf{Experimental Setup.} We utilize four source LLMs to construct our knowledge chains: \textbf{GPT2-XL} (1.5B) \citep{radford2019language}, \textbf{Llama3-8B} \citep{touvron2023llama2}, \textbf{Qwen3-4B} \citep{yang2025qwen3}, and \textbf{Pythia-1.4B} \citep{biderman2023pythia}. Using these as the "knowledge source," we generate SLMs across five specific scales ranging from 138M to 537M parameters. The evaluation suite comprises a broad spectrum of benchmarks: \textbf{MMLU} \citep{Hendrycks2020MeasuringMM} for general knowledge, \textbf{English\_XLSum} (XLsum) \citep{Narayan2018DontGM} for summarization, \textbf{HellaSwag} (HellaS) \citep{zellers2019hellaswag} and \textbf{WinoGrande} (WinG) \citep{ai2:winogrande} for commonsense reasoning, \textbf{BoolQ} \citep{Clark2019BoolQET} for reading comprehension, and \textbf{Dolly} \citep{DatabricksBlog2023DollyV2} for instruction following.

\textbf{Implementation Details.} All anchors within the knowledge chain undergo 5,000 steps of distillation. The constructed SLMs are then fine-tuned on target downstream tasks for 10 epochs. For the GPT2-based and Pythia-based chains, we utilize the OpenWebText and Pile datasets respectively for anchor construction. Further details regarding hyperparameters, prompt templates, and architectural configurations of the SLMs are provided in \cref{app:ex_details,app:datasets,app:architecture}.

\begin{figure*}[t]
  \centering
  \begin{minipage}{0.31\textwidth}
      \centering
      \begin{subfigure}{\linewidth}
          \includegraphics[width=\linewidth]{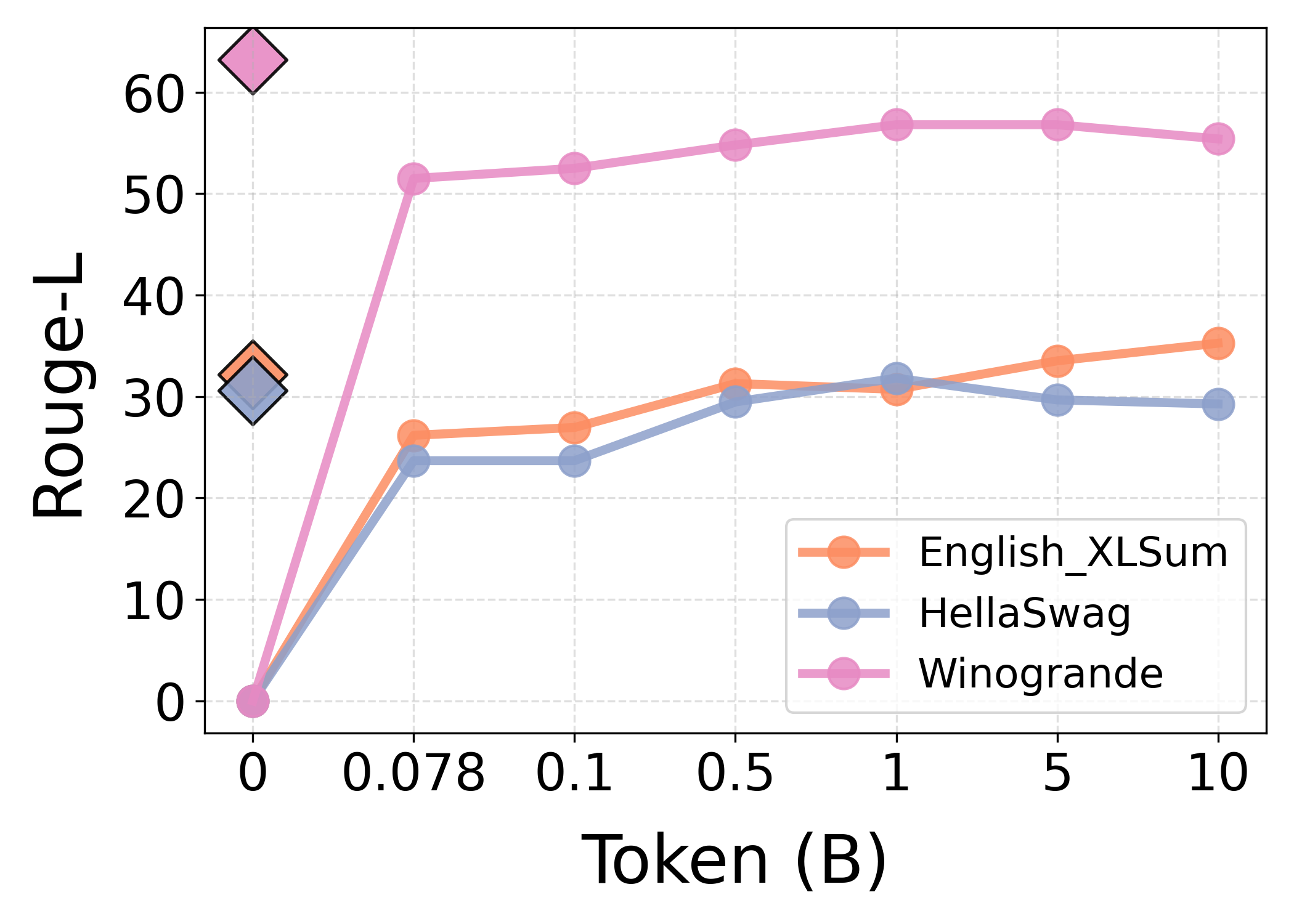}
      \end{subfigure}
      \begin{subfigure}{\linewidth}
          \includegraphics[width=\linewidth]{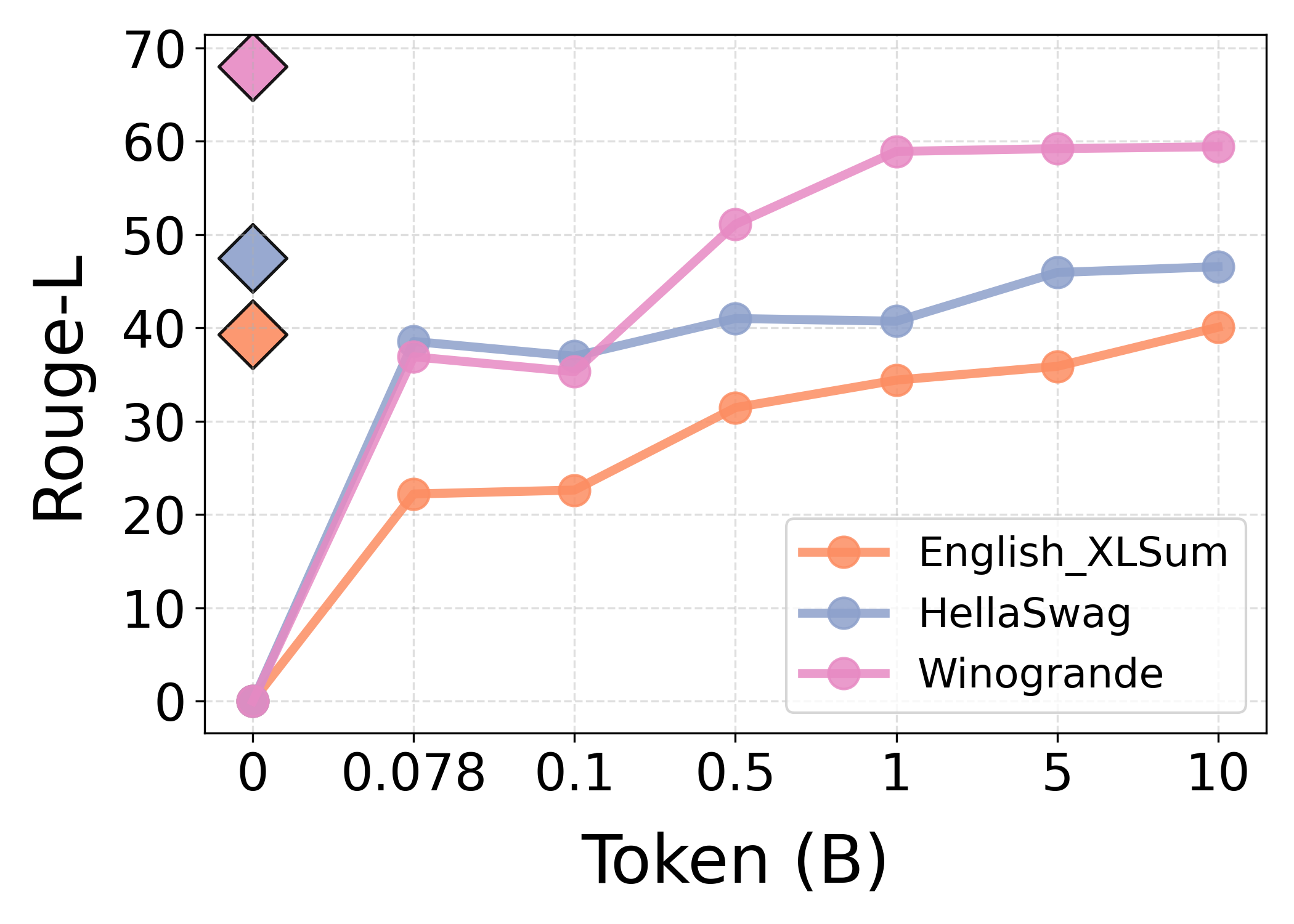}
          \caption{GPT2-XL}
      \end{subfigure}
  \end{minipage}
  \begin{minipage}{0.31\textwidth}
      \centering
      \begin{subfigure}{\linewidth}
          \includegraphics[width=\linewidth]{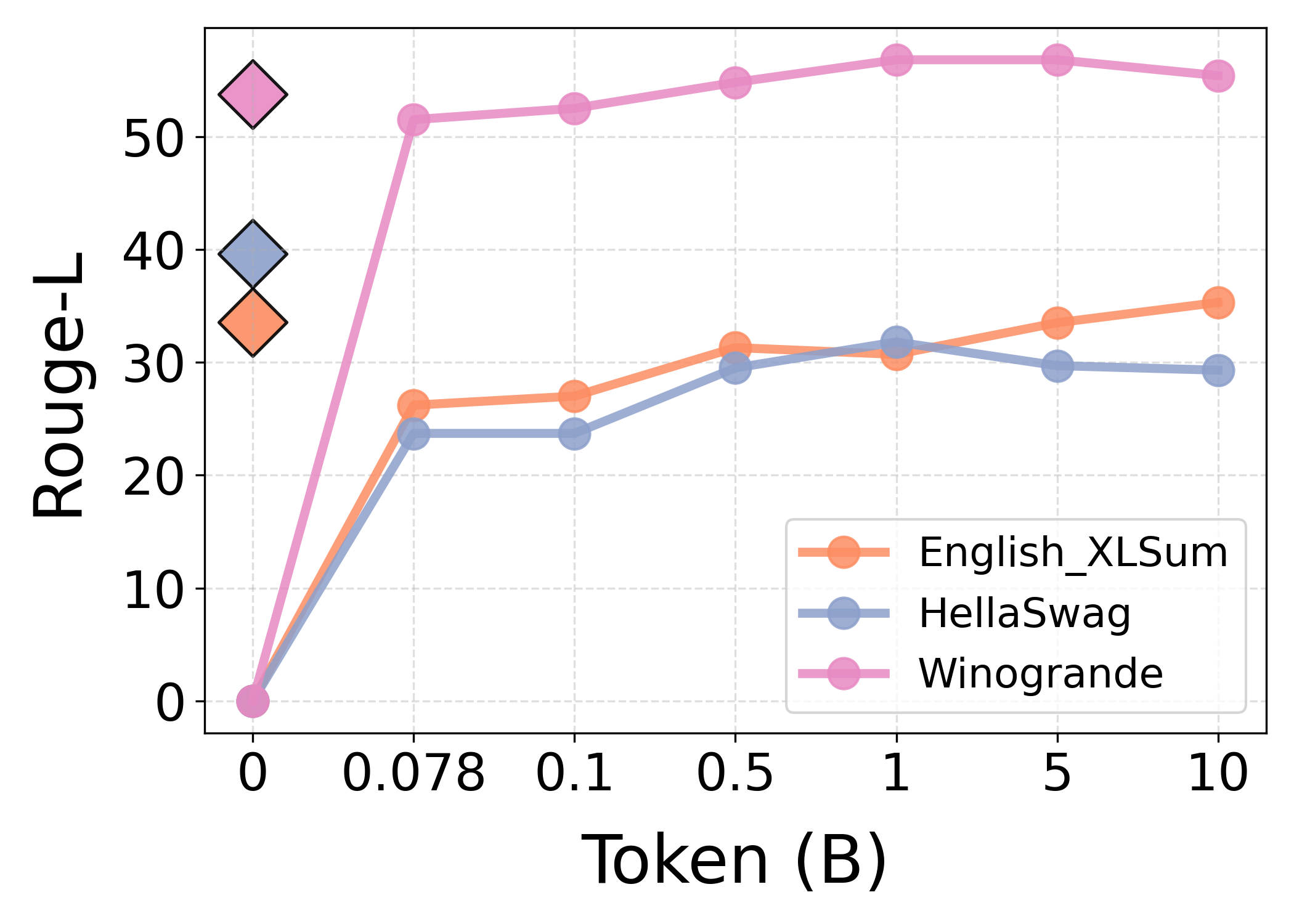}
      \end{subfigure}
      \begin{subfigure}{\linewidth}
          \includegraphics[width=\linewidth]{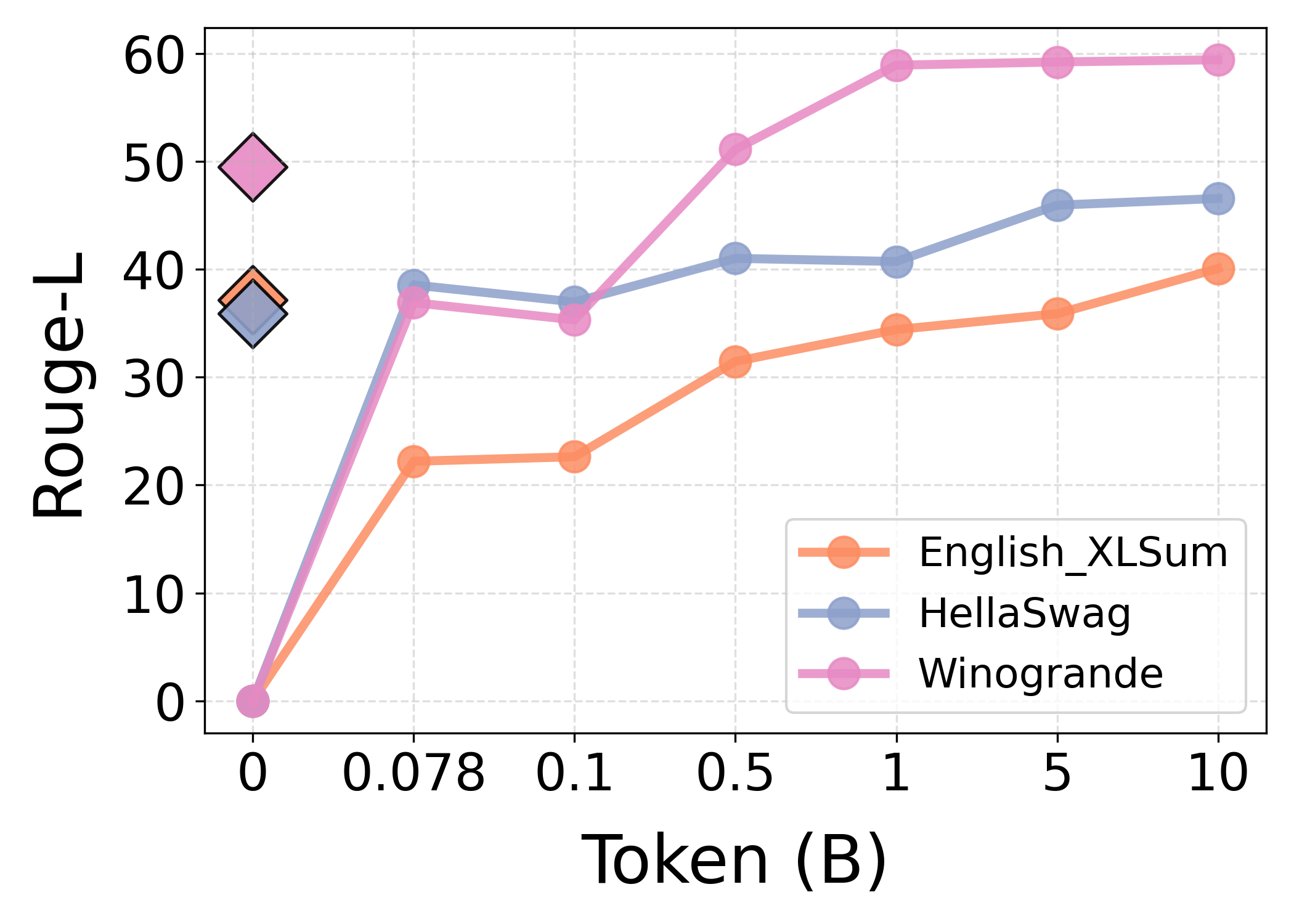}
          \caption{Llama3-8B}
      \end{subfigure}
  \end{minipage}
  \begin{minipage}{0.31\textwidth}
      \centering
      \begin{subfigure}{\linewidth}
          \includegraphics[width=\linewidth]{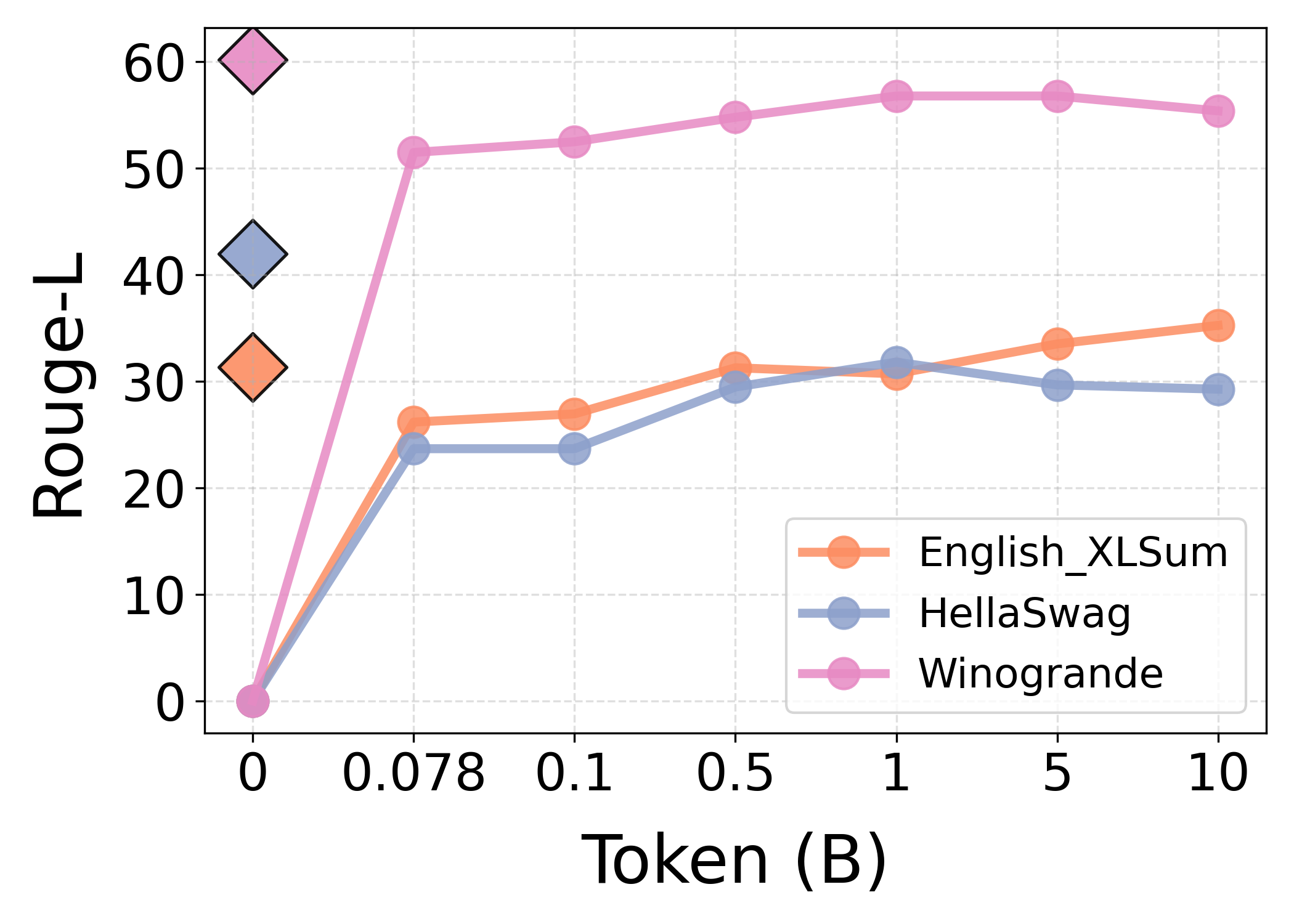}
      \end{subfigure}
      \begin{subfigure}{\linewidth}
          \includegraphics[width=\linewidth]{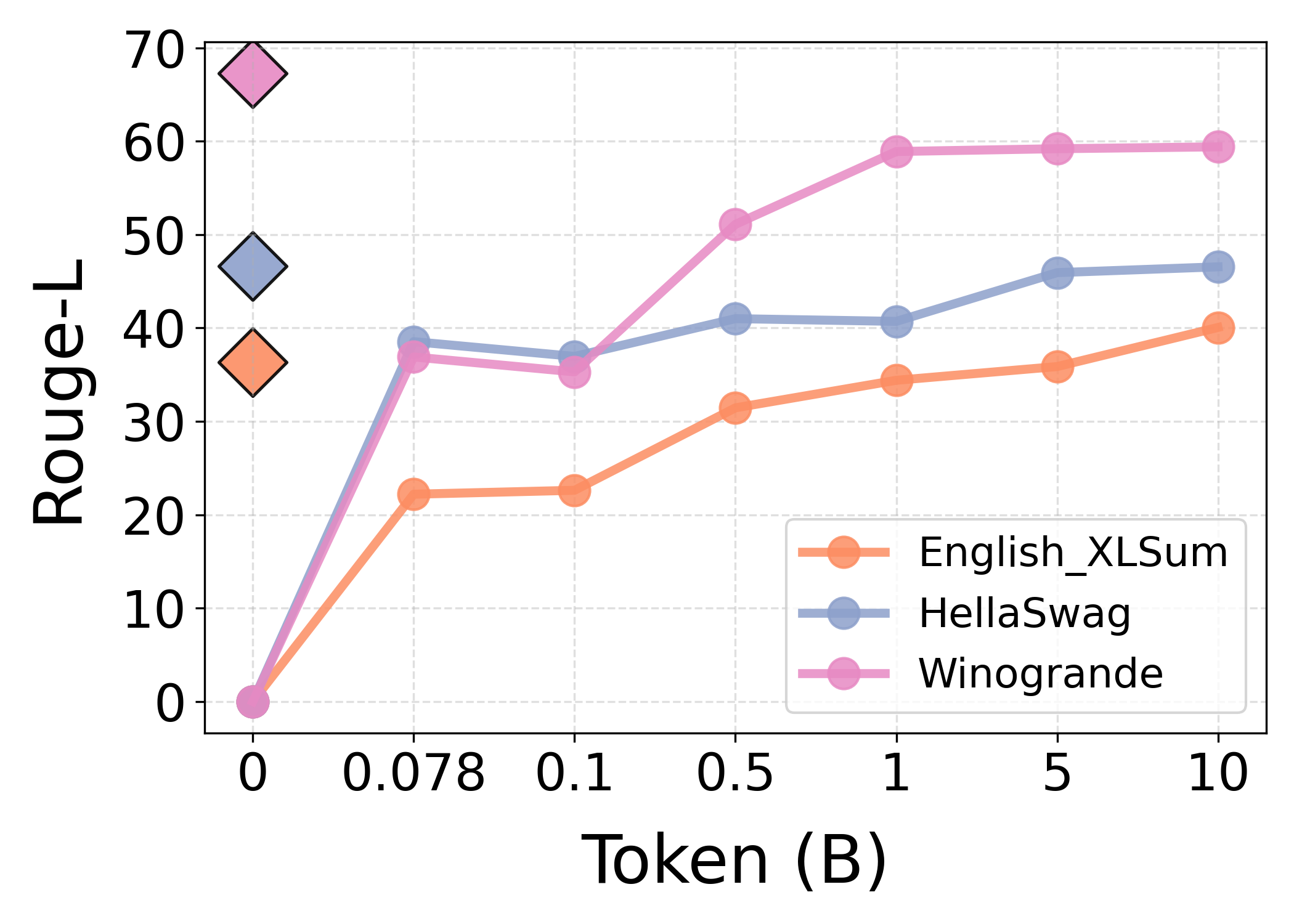}
          \caption{Qwen3-4B}
      \end{subfigure}
  \end{minipage}
  \vspace{-0.2cm}
  \caption{Quantification of training corpora savings enabled by CBD. Models initialized via our method (diamonds) without any recovery training consistently surpass those trained from scratch across billions of tokens. Line1: the 138M-parameter SLM, Line2: the 380M-parameter SLM.}
  \vspace{-0.4cm}
  \label{fig:token_saving_comparison}
\end{figure*}

\begin{figure*}[t]
    \centering
    \begin{minipage}{0.31\textwidth}
        \centering
        \begin{subfigure}{\linewidth}
             \includegraphics[width=\linewidth]{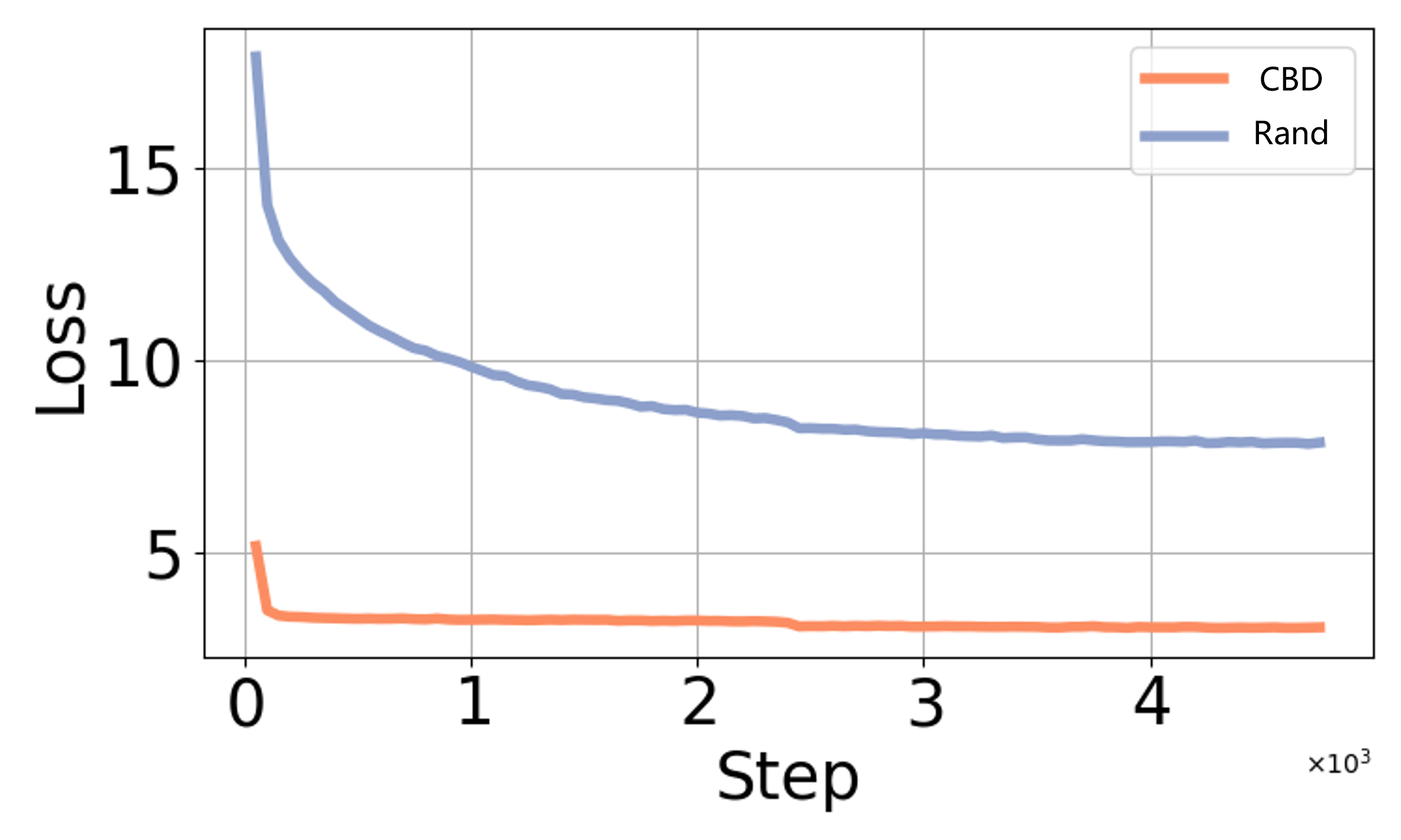}
        \caption{GPT2-XL}
        \end{subfigure}
            
    \end{minipage}
    \begin{minipage}{0.31\textwidth}
        \centering
        \begin{subfigure}{\linewidth}
            \includegraphics[width=\linewidth]{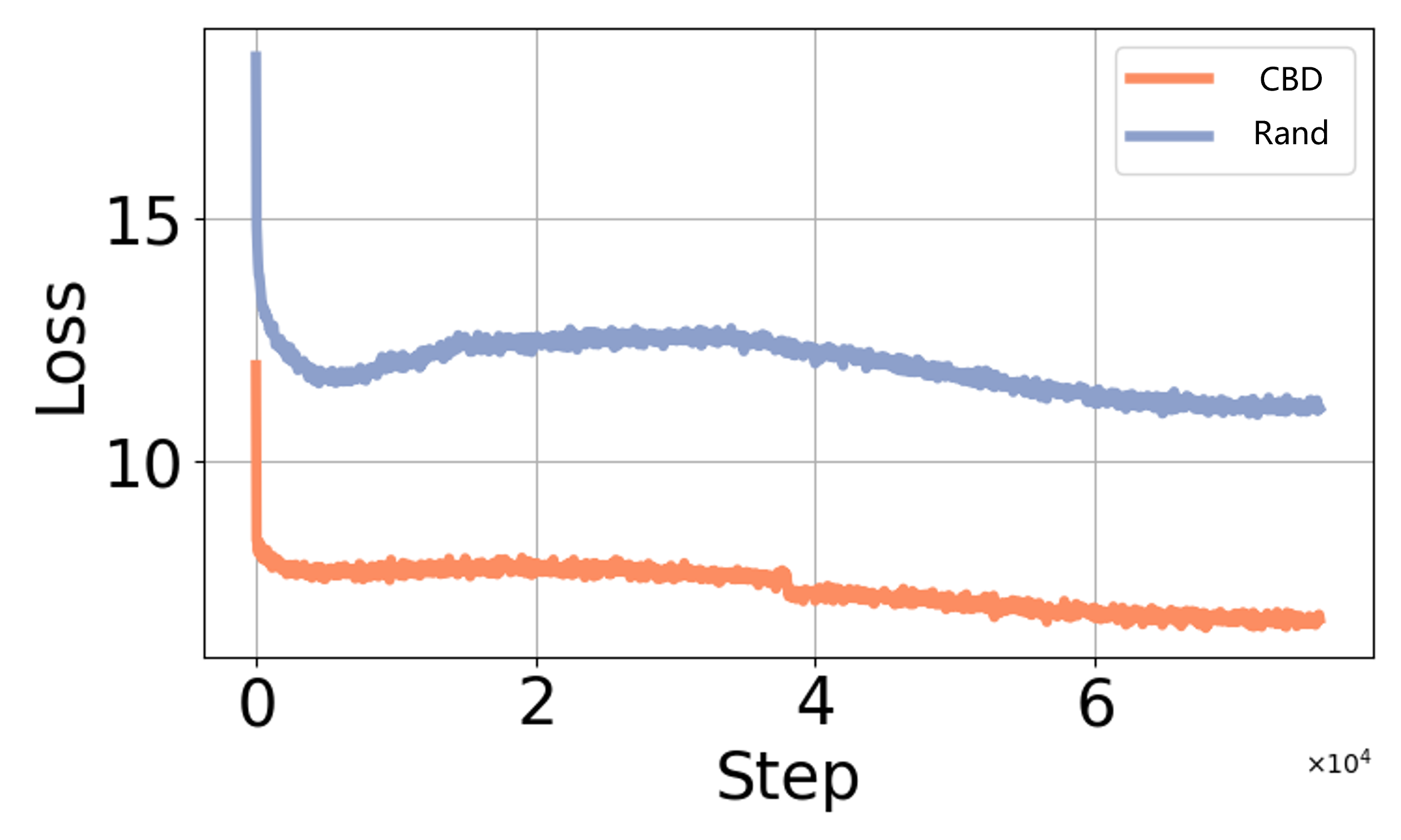}
            \caption{Llama3-8B}
        \end{subfigure}
            
    \end{minipage}
    \begin{minipage}{0.31\textwidth}
        \centering
        \begin{subfigure}{\linewidth}
            \includegraphics[width=\linewidth]{Qwen3-4B-DesNet-138M-LInit_vs_Pre-78Mtoken-NewName.png}
            \caption{Qwen3-4B}
        \end{subfigure}
            
    \end{minipage}
    \vspace{-0.2cm}
    \caption{Comparison of convergence trajectories between SLM-138M initialized via CBD and random initialization (Rand) on a 78M token budget. CBD exhibits a prominent step-zero advantage, where its initial loss is already lower than the final converged loss of the Rand baseline, demonstrating a substantial acceleration in optimization efficiency.}
    \label{fig:convergence}
    \vspace{-0.4cm}
\end{figure*}

\begin{table*}[t]
  \caption{Performance comparison between CBD and random initialization (Rand) under identical token budgets (78M and 100M). Across multiple SLM scales and benchmarks, CBD consistently achieves a higher accuracy ceiling, underscoring the effective inheritance of high-level semantic knowledge from the source LLM (GPT2-XL).}
  \label{tab:desnet_ansnet_token_comparison}
  \resizebox{\textwidth}{!}{
    \begin{tabular}{c|c|cccccc|cccccc}
      \toprule
      \multirow{2}{*}{Token Num} & \multirow{2}{*}{Init} & \multicolumn{6}{c|}{SLM-138M}                                                & \multicolumn{6}{c}{SLM-220M}                                                 \\ \cmidrule{3-14} 
                                 &                       & MMLU  & XLsum & HellaS & WinoG & \multicolumn{1}{c|}{BoolQ} & Avg            & MMLU  & XLsum & HellaS & WinoG & \multicolumn{1}{c|}{BoolQ} & Avg            \\ \midrule
      \multirow{2}{*}{78M}       & Rand                   & 29.00 & 26.18 & 23.68  & 51.50 & \multicolumn{1}{c|}{68.00} & 39.67          & 28.40 & 28.48 & 23.87  & 49.70 & \multicolumn{1}{c|}{66.50} & 39.39          \\
                                 & CBD                   & 31.00 & 30.68 & 31.15  & 62.00 & \multicolumn{1}{c|}{76.80} & \textbf{46.33} & 28.50 & 26.68 & 26.84  & 58.00 & \multicolumn{1}{c|}{69.50} & \textbf{41.90} \\ \midrule
      \multirow{2}{*}{100M}      & Rand                   & 29.10 & 26.95 & 23.68  & 52.50 & \multicolumn{1}{c|}{67.80} & 40.01          & 29.30 & 26.87 & 23.67  & 52.80 & \multicolumn{1}{c|}{69.10} & 40.35          \\
                                 & CBD                   & 30.90 & 31.54 & 30.94  & 61.90 & \multicolumn{1}{c|}{76.60} & \textbf{46.38} & 28.20 & 28.27 & 28.76  & 59.20 & \multicolumn{1}{c|}{70.70} & \textbf{43.03} \\ \midrule
      \multirow{2}{*}{Token Num} & \multirow{2}{*}{Init} & \multicolumn{6}{c|}{SLM-277M}                                                & \multicolumn{6}{c}{SLM-380M}                                                 \\ \cmidrule{3-14} 
                                 &                       & MMLU  & XLsum & HellaS & WinoG & \multicolumn{1}{c|}{BoolQ} & Avg            & MMLU  & XLsum & HellaS & WinoG & \multicolumn{1}{c|}{BoolQ} & Avg            \\ \midrule
      \multirow{2}{*}{78M}       & Rand                   & 26.40 & 27.30 & 23.77  & 50.20 & \multicolumn{1}{c|}{67.50} & 39.03          & 26.70 & 22.20 & 38.54  & 36.90 & \multicolumn{1}{c|}{67.90} & 38.45          \\
                                 & CBD                   & 28.20 & 32.36 & 29.57  & 61.49 & \multicolumn{1}{c|}{71.80} & \textbf{44.68} & 30.90 & 34.14 & 46.04  & 68.60 & \multicolumn{1}{c|}{74.90} & \textbf{50.92} \\ \midrule
      \multirow{2}{*}{100M}      & Rand                   & 27.90 & 27.68 & 24.39  & 47.50 & \multicolumn{1}{c|}{66.70} & 38.83          & 28.50 & 22.62 & 36.96  & 35.30 & \multicolumn{1}{c|}{65.10} & 37.70          \\
                                 & CBD                   & 31.30 & 35.31 & 29.57  & 58.47 & \multicolumn{1}{c|}{73.00} & \textbf{45.53} & 28.80 & 35.53 & 43.23  & 66.20 & \multicolumn{1}{c|}{75.00} & \textbf{49.75} \\ \bottomrule
      \end{tabular}}
      \vspace{-0.2cm}
\end{table*}

\begin{table*}[ht]
  \centering
  \caption{Performance comparison between the CBD and baseline initilalization methods (learngene). CBD outperforms the learngene methods across the Tasks on the source LLM (GPT2-XL).}
  \begin{tabular}{c|ccccccc|c}
  \toprule
  Method         & Params  & MMLU           & XLsum          & HellaS      & WinoG     & BoolQ          & Dolly          & Avg   \\ \midrule
  Auto-Learngene & 174.28 & 25.47          & \textbf{33.91} & 34.98          & 44.72          & 65.65          & 20.30          & 37.50 \\
  Van-Learngene  & 174.28 & 24.62          & 32.90          & \textbf{36.24} & 41.83          & 63.79          & 19.62          & 36.50 \\
  CBD            & 134.00 & \textbf{31.00} & 30.68          & 31.15          & \textbf{62.00} & \textbf{76.80} & \textbf{22.99} & \textbf{42.44} \\ \midrule
  Auto-Learngene & 235.76 & 24.66          & \textbf{35.18} & \textbf{36.45} & 44.12          & 60.88          & \textbf{20.03} & 36.89 \\
  Van-Learngene  & 235.76 & 24.46          & 33.30          & 35.85          & 42.23          & 55.86          & 19.92          & 35.27 \\
  CBD            & 220.00 & \textbf{28.50} & 26.68          & 26.84          & \textbf{58.00} & \textbf{69.50} & 19.82          & \textbf{38.22} \\ \midrule
  Auto-Learngene & 297.24 & 24.53          & 31.86          & 35.02          & 44.28          & 61.94          & 20.43          & 36.34 \\
  Van-Learngene  & 297.24 & 23.77          & \textbf{32.70} & 34.28          & 44.48          & 62.62          & 20.67          & 36.42 \\
  CBD            & 277.00 & \textbf{28.20} & 32.36          & \textbf{43.03} & \textbf{61.49} & \textbf{71.80} & \textbf{20.82} & \textbf{42.95} \\ \bottomrule
  \end{tabular}
  \label{tab:col_vs_previous_learngene}
  \vspace{-0.6cm}
\end{table*}

\subsection{Main Results and Analysis}
We evaluate the CBD-based initialization (hereafter referred to as CBD) from three critical dimensions: (1) training corpora efficiency, (2) convergence acceleration, and (3) performance gains under constrained data budgets.

\subsubsection{Significant Savings in Training Corpora} 
A primary motivation for CBD is to reduce the massive tokens requirements of pretraining SLMs from scratch. We compare CBD \textbf{without any recovery training} against conventional pre-training from scratch (\textit{i.e.,} Random Initialization, Rand) using increasing volumes of the OpenWebText corpus.

As illustrated in \cref{fig:token_saving_comparison}, SLMs initialized via CBD consistently outperform their pre-trained counterparts even when the latter are trained on billions of tokens. Specifically, on the \textbf{HellaSwag} benchmark, an SLM-138M initialized by CBD surpasses a model trained from scratch on over \textbf{10B tokens} across all source LLM configurations, as in \cref{fig:token_saving_comparison}a (Line1). This implies that CBD provides a knowledge-rich starting point that effectively "bypasses" the first 10B tokens of general pre-training. Similarly, for the \textbf{English\_XLSum} task, SLM-380M achieves comparable or superior performance while saving at least 5B tokens of pre-training, as in \cref{fig:token_saving_comparison}a (Line2). These results underscore that CBD does not merely provide a better starting point but captures essential linguistic patterns that would otherwise require massive computational overhead to learn from scratch. The detailed values in \cref{fig:token_saving_comparison} are presented in \cref{app:token_saving_value}.

\subsubsection{Accelerated Convergence during Pre-training} 
To further investigate the optimization benefits, we analyze the convergence curves of CBD compared to random initialization (Rand). Using a fixed budget of 78M tokens for SLM-138M, we observe a dramatic disparity in learning efficiency.

As shown in \cref{fig:convergence}, regardless of whether the source LLM is GPT2-XL, Llama3-8B, or Qwen3-4B, the \textbf{initial loss} of CBD at step zero is already lower than the \textbf{final converged loss} of the Rand baseline. This step-zero advantage suggests that the parameter interpolation from the knowledge chain successfully preserves the manifold structure of the teacher's latent space. 
Additional convergence analyses across varying token budgets (detailed in \cref{app:convergence}) confirm that this acceleration is robust and not limited to specific tasks.
Especially, CBD enables the model to reach a target loss value up to \textbf{200$\times$ faster} than traditional random initialization. 

\subsubsection{Performance Superiority under Equal Token Budgets} 
In resource-constrained research or industrial settings, the ability to maximize performance under a fixed token budget is vital. \cref{tab:desnet_ansnet_token_comparison} summarizes the performance across five benchmarks under equal budgets of 78M and 100M tokens.

The empirical evidence is compelling: CBD consistently outperforms the Rand baseline across all tasks and SLM scales. For instance, using \textbf{GPT2-XL} as the source, an SLM-380M initialized with CBD achieves an average score of 35.53 on XLsum, a substantial \textbf{12.91 improvement} over the 22.62 achieved by Rand. This gap indicates that the knowledge chain successfully distills and preserves high-level semantic features from the source LLM, which are effectively inherited by the SLMs during interpolation. 
The additional comparision of CBD against Rand using \textbf{Llama3-8B} and \textbf{Qwen3-4B} as the source LLMs are listed in \cref{app:cbd_vs_rand_llama3_qwen3}, where the similar results are shown. This consistently high average performance across all SLM scales and the source LLMs highlights the versatility and robustness of CBD under varying resource constraints.

\textbf{Comparison with Initialization Methods.} We also benchmark CBD against earlier architectural transfer approaches, namely \textbf{Auto-Learngene} \citep{wang2023learngene} and \textbf{Vanilla-Learngene} \citep{wang2022learngene}. While these methods were originally designed for ViT architectures, we adapted them to the LLM setting. As detailed in \cref{tab:col_vs_previous_learngene}, CBD consistently yields superior results. We attribute this to the \textbf{Chain} structure, which is unlike previous methods that attempt a single "jump" from teacher to student, CBD's stepwise approach mitigates the catastrophic information loss caused by the massive capacity gap in LLMs. 
Our evaluation focuses on representative initialization baselines, as other learngene methods specifically optimized for Vision Transformers (ViTs) lack the structural compatibility required for LLMs. A detailed analysis regarding the technical challenges for this cross-domain transfer is provided in \cref{app_previous_learngene}.

\textbf{Generality Across Architectures.} Finally, to demonstrate the universality of CBD, we extend our evaluation to the \textbf{Pythia} and \textbf{Qwen3} chains. As shown in \cref{tab:pythia_chain} and \cref{app_qwen3_col}, the performance gains remain consistent. The 113.63M SLM based on the Pythia chain demonstrates that even with different tokenizers and architectural nuances, the principle of knowledge chain interpolation holds, making CBD a versatile tool for various LLM families.

\subsection{Ablation Studies}
\label{sec:ablation}
In this section, we dissect the core components of CBD to elucidate their individual contributions to the framework's overall efficacy. Our analysis focuses on why the knowledge chain structure and the interpolation mechanism outperform traditional paradigms in both stability and performance.

\begin{figure*}[t]
    \centering
        \begin{minipage}{0.45\textwidth}
            \centering
            \begin{subfigure}{\linewidth}
                 \includegraphics[width=\linewidth]{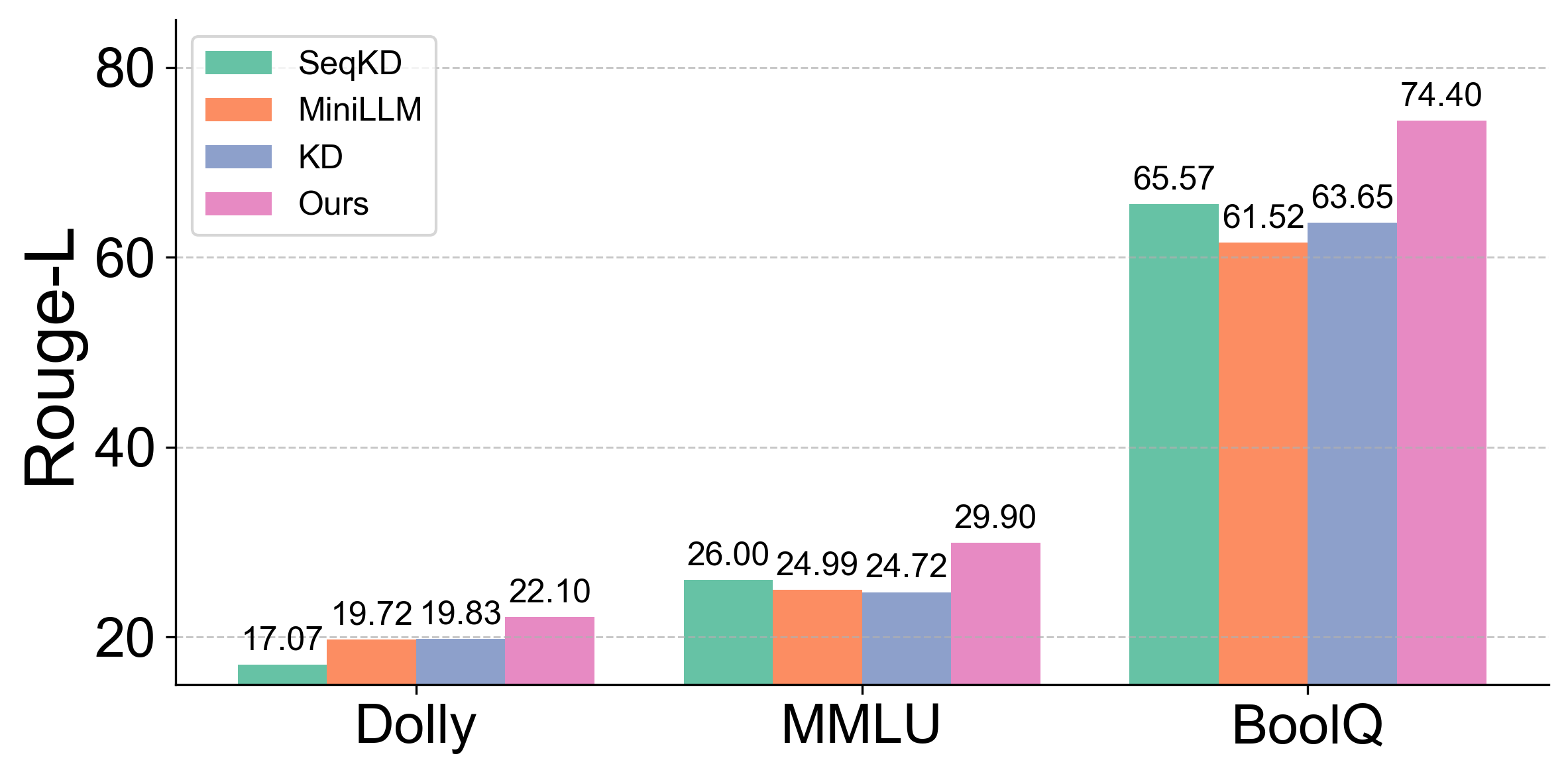}
            \caption{}
            \label{fig:direct_vs_ours_performance}
            \end{subfigure}
        \end{minipage}
        \begin{minipage}{0.42\textwidth}
            \centering
            \begin{subfigure}{\linewidth}
                \includegraphics[width=\linewidth]{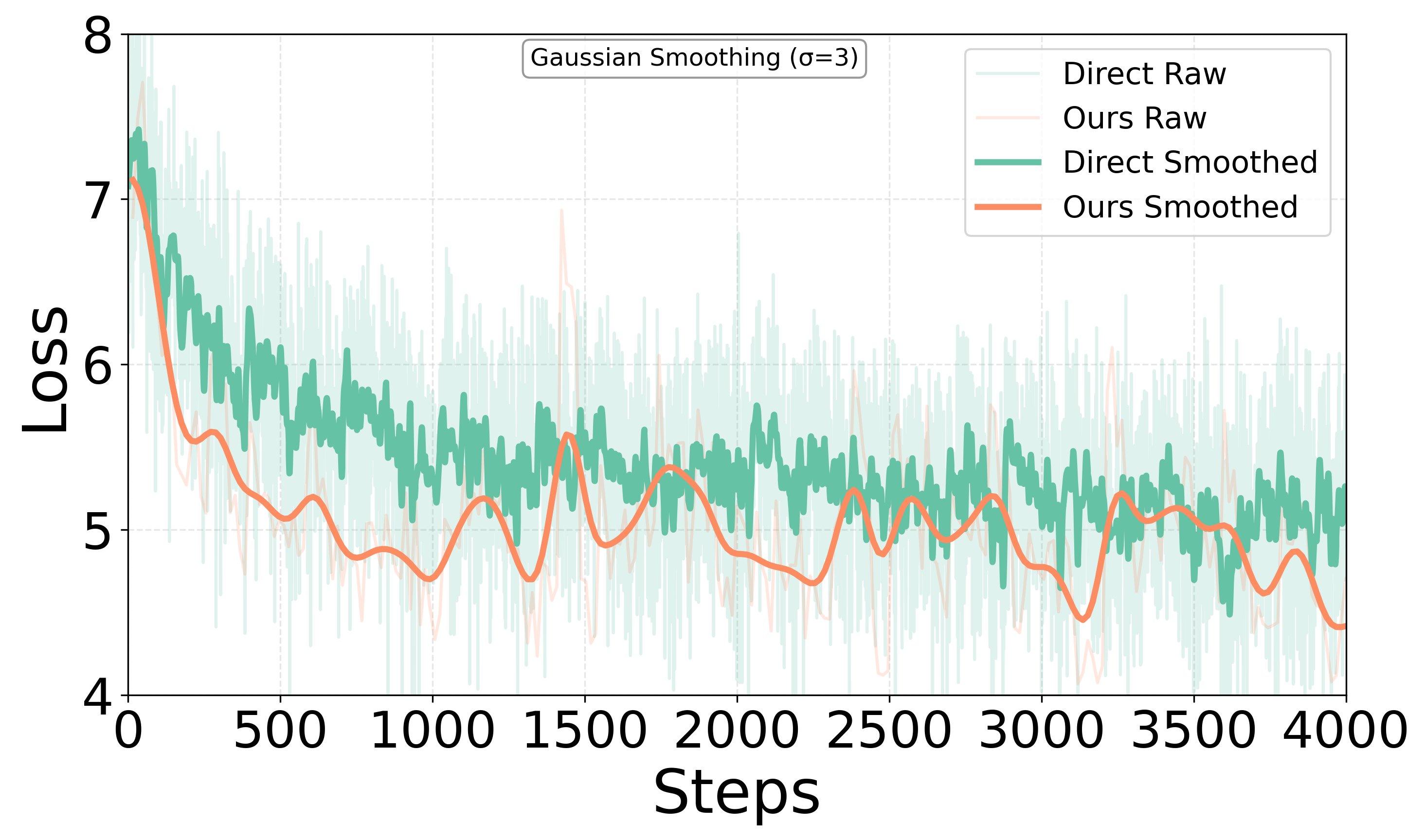}
                \caption{}
                \label{fig:direct_vs_ours_stable}
            \end{subfigure}
        \end{minipage}
      \vspace{-0.3cm}
      \caption{(a) Downstream performance of SLM-138M comparing CBD against state-of-the-art distillation baselines. (b) Training stability analysis. CBD's stepwise approach produces a significantly smoother and more stable loss curve by effectively bridging the semantic distance through intermediate knowledge buffers.}
    \label{fig:direct_vs_ours}
    \vspace{-0.2cm}
  \end{figure*}
  
  \begin{figure*}[t]
    \centering
    \begin{minipage}{0.3\textwidth}
      \centering
      \begin{subfigure}{\linewidth}
          \includegraphics[width=\linewidth]{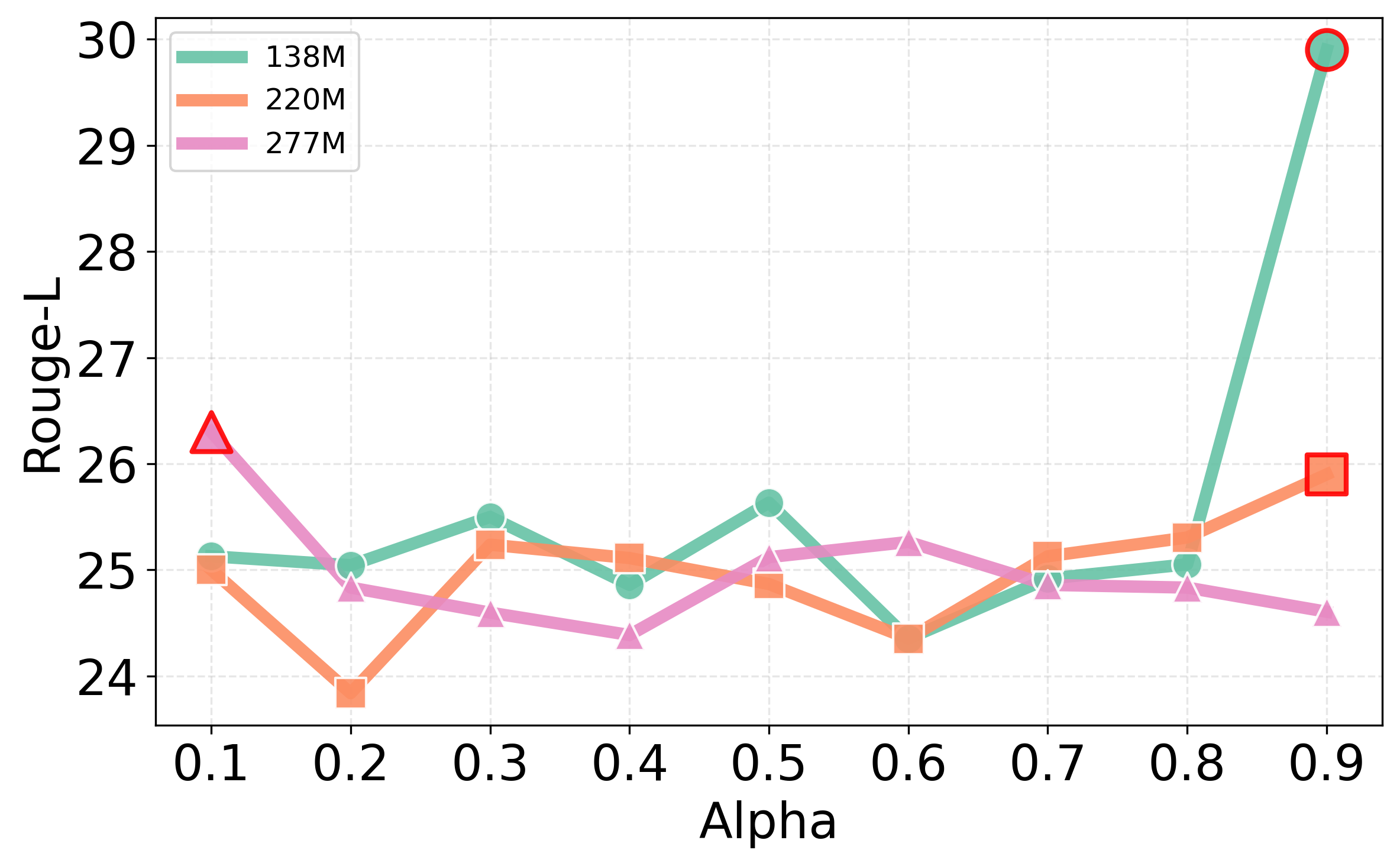}
          \caption{GPT2-XL}
      \end{subfigure}
    \end{minipage}
    \begin{minipage}{0.3\textwidth}
      \centering
      \begin{subfigure}{\linewidth}
          \includegraphics[width=\linewidth]{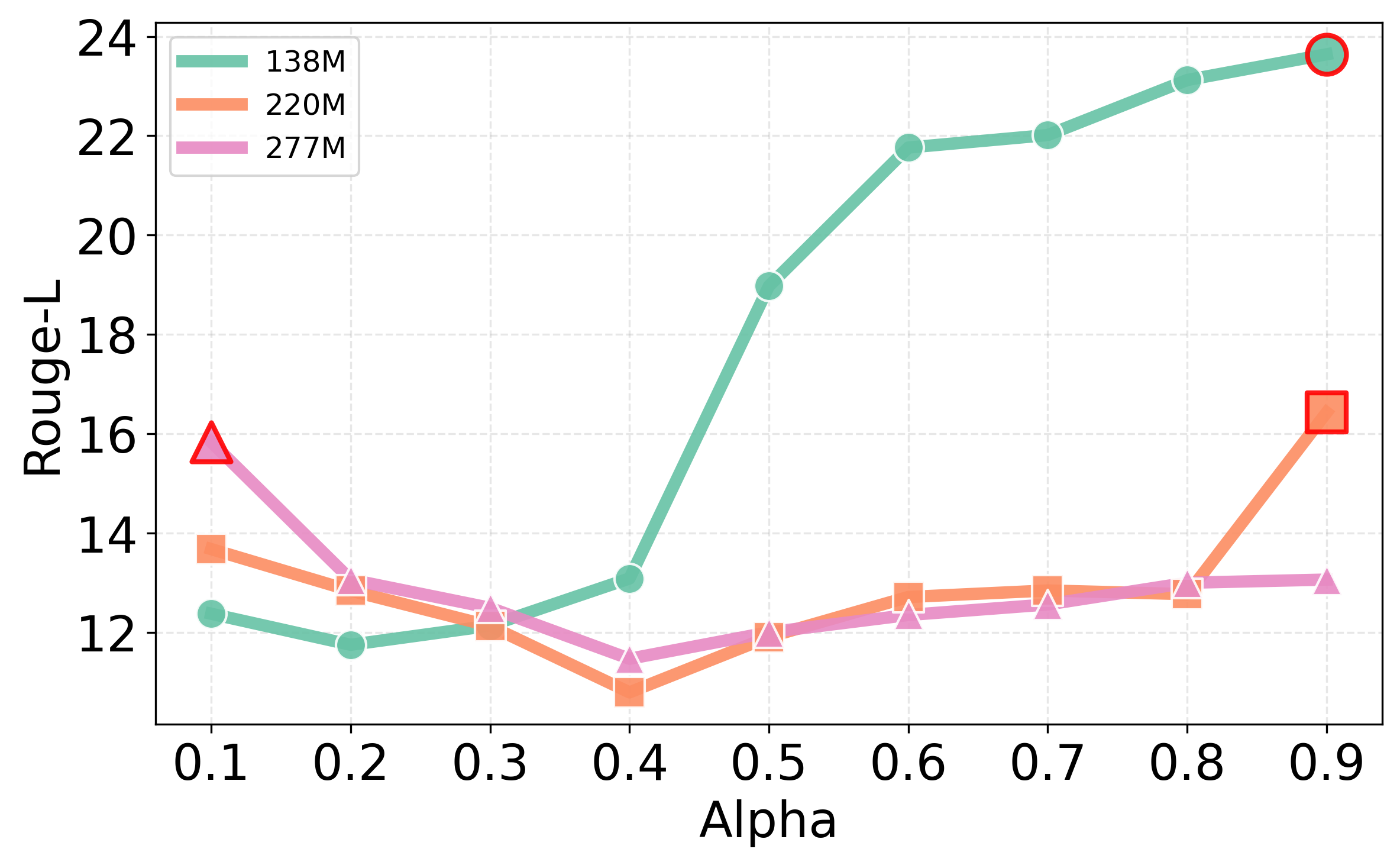}
          \caption{Llama3-8B}
      \end{subfigure}
    \end{minipage}
    \begin{minipage}{0.3\textwidth}
      \centering
      \begin{subfigure}{\linewidth}
          \includegraphics[width=\linewidth]{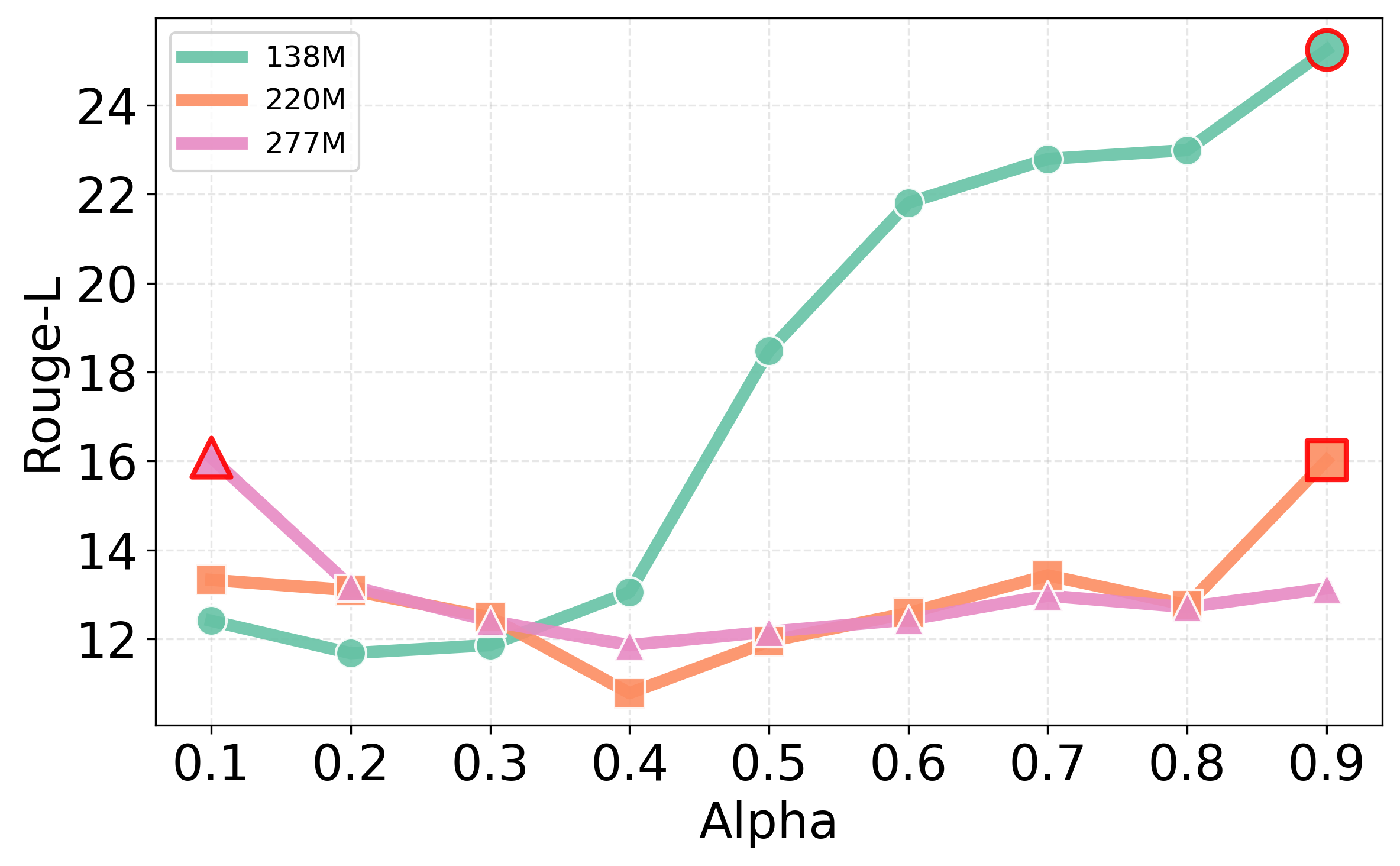}
          \caption{Qwen3-4B}
      \end{subfigure}
    \end{minipage}
    \caption{Sensitivity analysis of the interpolation coefficient $\alpha$ across various \textbf{source LLMs}. The results reveal a strong correlation between the optimal $\alpha$ and the architectural proximity of the target \textbf{SLM} to its adjacent anchors, validating the continuity of the parameter manifold within the \textbf{knowledge chain}.}
    \label{fig:alpha_ablation}
    \vspace{-0.6cm}
  \end{figure*}  

  \begin{table}[t]
    \centering
    \small
    \caption{Comprehensive comparison between CBD and the single-expansion baseline across two source LLMs and two SLM scales. CBD's multi-anchor interpolation yields increasingly significant gains as the target model size expands, highlighting the necessity of cross-scale structural alignment.}
    \resizebox{0.48\textwidth}{!}{
    \begin{tabular}{cccccccccc}
    \toprule
    \multicolumn{1}{c|}{}                           & \multicolumn{4}{c|}{SLM-220M}                                                                                                                        & \multicolumn{4}{c}{SLM-277M}                                                                                                    \\
    \multicolumn{1}{c|}{Method}                     & Dolly                        & MMLU                         & BoolQ                        & \multicolumn{1}{c|}{Avg}                                   & Dolly                        & MMLU                         & BoolQ                        & Avg                                   \\ \midrule
    \multicolumn{9}{c}{ GPT2-XL}                                                                                                                                                                                                                                                                                                                                                                                                                                                                     \\ \midrule
    \multicolumn{1}{c|}{single}                     & 12.33                        & 25.26                        & 65.54                        & \multicolumn{1}{c|}{34.38}                                 & 11.98                        & 25.45                        & 66.71                        & 34.71                                 \\
   \multicolumn{1}{c|}{{\textbf{CBD}}} & {15.89} & {25.90} & {66.70} & \multicolumn{1}{c|}{{\textbf{36.16}}} & {14.72} & {26.30} & {68.40} & {\textbf{36.47}} \\ \midrule
    \multicolumn{9}{c}{ Llama3-8B}                                                                                                                                                                                                                                                                                                                                                                                                                                                                \\ \midrule
    \multicolumn{1}{c|}{single}                     & 13.03                        & 25.51                        & 65.50                        & \multicolumn{1}{c|}{34.68}                                 & 13.26                        & 24.88                        & 67.39                        & 35.17                                 \\
   \multicolumn{1}{c|}{{\textbf{CBD}}} & 16.43                        & 25.53                        & 67.32                        & \multicolumn{1}{c|}{\textbf{36.43}}                        & 15.83                        & 24.58                        & 66.64                        & \textbf{35.69}                        \\ \midrule
    \multicolumn{9}{c}{ Qwen3-4B}                                                                                                                                                                                                                                                                                                                                                                                                                                                                    \\ \midrule
    \multicolumn{1}{c|}{single}                     & 13.41                        & 25.33                        & 66.54                        & \multicolumn{1}{c|}{35.09}                                 & 12.69                        & 24.50                        & 66.25                        & 34.48                                 \\
    \multicolumn{1}{c|}{\textbf{CBD}}                        & 16.02                        & 25.75                        & 64.60                        & \multicolumn{1}{c|}{\textbf{35.46}}                        & 16.08                        & 28.90                        & 67.40                        & \textbf{37.46}           \\ \bottomrule            
    \end{tabular}
    }
    \label{tab:single_vs_col}
    \vspace{-0.7cm}
\end{table}

\subsubsection{CBD vs. Conventional Knowledge Distillation} 
\textbf{Performance Dynamics across the Capacity Gap.} 
To investigate whether the complexity of a knowledge chain is truly necessary, we benchmark CBD against state-of-the-art distillation baselines: SeqKD, MiniLLM \citep{gu2023minillm}, and vanilla KD. As illustrated in \cref{fig:direct_vs_ours_performance}, across the Dolly, MMLU, and BoolQ benchmarks, CBD consistently yields improvements over the strongest distillation baseline. 

The pivotal observation occurs in the extreme compression regime. As shown in \cref{tab:col_vs_direct}, when attempting to distill knowledge from a source LLM (e.g., GPT2-XL) directly into a tiny student (e.g., GPT2-B), conventional KD suffers from a performance collapse due to the overwhelming \textbf{information density mismatch}. In contrast, CBD maintains high fidelity by decomposing this massive capacity gap into several incremental, manageable steps through the knowledge chain. This confirms that intermediate anchors act as knowledge buffers, preventing the student from being overwhelmed by complex teacher distributions it lacks the capacity to model.

\textbf{Optimization Stability and Gradient Variance.} 
Beyond final accuracy, we examine the optimization landscape. We compare the distillation of GPT2-B via two paths: a direct path (GPT2-XL $\to$ GPT2-B) and the CBD path (via the GPT2-M anchor). \cref{fig:direct_vs_ours_stable} visualizes the raw and smoothed training losses. The direct distillation exhibits high-frequency oscillations, suggesting that the semantic distance between the teacher and student creates a rugged loss surface. Conversely, CBD provides a significantly smoother trajectory. By ensuring that each distillation step occurs between models of comparable complexity, we effectively mitigate gradient stochastics and facilitate a more stable convergence to a superior local optimum.

\begin{table}[t]
    \centering
    \begin{minipage}[t]{0.48\textwidth}
    \centering
    \caption{Performance of SLM-113.63M built from the Pythia-based knowledge chain.}
    \begin{tabular}{c|c|c|c|c}
    \toprule
    & BoolQ & Arc-E & WinoG & Avg \\
    \midrule
    Rand & 66.36 & 23.68 & 47.72 & 45.92 \\
    CoL  & 65.82 & 28.07 & 48.84 & \textbf{47.58} \\
    \bottomrule
    \end{tabular}
    \label{tab:pythia_chain}
    \end{minipage}
    \vspace{-0.3cm}
\end{table}

\begin{table}[t]
    \centering
    \begin{minipage}[t]{0.48\textwidth}
    \centering
    \caption{Zero-shot performance of intermediate anchors on the Dolly task. CBD mitigates the performance reduce typically observed in direct distillation when a massive capacity gap exists between the source LLM and a small student. `L', `M', `B' are GPT2-L, GPT2-M, and GPT2-B.}
    \begin{tabular}{c|c|c|c|c}
    \toprule
    & LLM & L & M & B \\
    \midrule
    Direct & 27.94 & - & - & 23.30 \\
    CoL    & 27.94 & 27.77 & 27.63 & \textbf{26.02} \\
    \bottomrule
    \end{tabular}
    \label{tab:col_vs_direct}
    \end{minipage}
    \vspace{-0.5cm}
\end{table}


\subsubsection{Optimal Choice of Interpolation Coefficient $\alpha$} 
The coefficient $\alpha$ is a hyperparameter that governs the linear combination of adjacent anchors in the parameter manifold. To determine its impact, we evaluate SLMs of different sizes (138M, 220M, 277M) with $\alpha$ varying from 0.1 to 0.9.

As shown in \cref{fig:alpha_ablation}, the optimal $\alpha$ is intrinsically linked to the \textbf{architectural proximity} of the target SLM to its neighbors in the chain. For instance, the 138M-parameter SLM, which is structurally closer to the GPT2-B (117M) anchor, achieves peak performance at $\alpha = 0.9$. Conversely, larger SLMs exhibit higher sensitivity to the larger anchor, favoring smaller $\alpha$ values. This empirical trend validates our hypothesis that the knowledge chain constructs a continuous and searchable parameter space. It suggests that knowledge is not just stored in individual anchors but resides in the \textbf{interpolated trajectory} between them, allowing CBD to sample highly-effective initializations for variable scale.

\subsubsection{The Power of Multi-Anchor Alignment} 
To highlight the advantage of CBD's multi-anchor interpolation, we compare it with a \textit{single-expansion} baseline that initializes SLMs only from the smallest anchor.
As shown in \cref{tab:single_vs_col}, CBD consistently outperforms the baseline across all configurations. For GPT2-XL, CBD improves Dolly performance by +3.56 (SLM-220M) and +2.74 (SLM-277M), with average gains of +1.78 and +1.76 points respectively. 
This advantage demonstrates that multi-anchor interpolation provides superior initialization quality. By leveraging structural information from both larger and smaller neighbors, CBD positions parameters in more optimal regions of the loss landscape, enabling more effective knowledge transfer than single-anchor approaches.

\textbf{Addressing Compute Fairness.} 
We further compare CBD against a stronger \textit{single-expansion} baseline that receives an additional 100M recovery tokens before downstream fine-tuning. As shown in \cref{tab:single_vs_col_recovery}, even with substantially more training data, the Single model still performs worse than CBD, which requires \textbf{no} extra tokens. 

This result demonstrates that the initialization quality of CBD is \textbf{qualitatively superior} to that of Single-expansion. The multi-anchor strategy provides a cross-scale alignment. By leveraging information from both a larger and a smaller neighbor, CBD ensures that the initialized parameters are already situated in a region of the loss landscape that is both high-performing and generalizable. 

\begin{table}[t]
    \centering
    \begin{minipage}[t]{0.48\textwidth}
    \centering
    \caption{Stress test for compute fairness: even when the Single-expansion baseline is augmented with an additional 100M recovery tokens, zero-shot CBD (0 tokens) maintains a significant performance margin.}
    \begin{tabular}{c|c|c|c|c}
    \toprule
    & BoolQ & Arc-E & WinoG & Avg \\
    \midrule
    Single & 66.11 & 26.04 & 58.06 & 50.07 \\
    CoL    & 74.40 & 25.70 & 63.20 & \textbf{54.43} \\
    \bottomrule
    \end{tabular}
    \label{tab:single_vs_col_recovery}
    \end{minipage}
    \vspace{-0.5cm}
\end{table}
    
\begin{table}[t]
    \centering
    \begin{minipage}[t]{0.48\textwidth}
    \centering
    \caption{Impact of knowledge chain density: a denser chain (4 anchors) provides more precise parameter interpolation and smoother transitions, resulting in superior downstream accuracy.}
    \begin{tabular}{c|c|c|c|c}
    \toprule
    Length & BoolQ & Arc-E & WinoG & Avg \\
    \midrule
    2 & 65.60 & 24.21 & 48.92 & 46.24 \\
    4 & 65.82 & 28.07 & 48.84 & \textbf{47.58} \\
    \bottomrule
    \end{tabular}
    \label{tab:pythia_chain_length}
    \end{minipage}
    \vspace{-0.5cm}
\end{table}

\subsubsection{Impact of Knowledge Chain Density} 
Finally, we investigate the granularity of the knowledge chain. By fixing the target SLM size at 113.63M and varying the number of intermediate anchors, we observe a clear correlation between chain density and performance.
As shown in \cref{tab:pythia_chain_length}, a 4-anchor chain (with inter-model capacity gaps of $\approx$30\%) outperforms a sparse 2-anchor chain (gaps of $\approx$70\%) by 1.34 on average. Denser chains provide a more "fine-grained" interpolation, reducing the \textbf{approximation error} that occurs during the linear combination of parameters. This confirms that the knowledge chain effectively discretizes the continuous knowledge space of the source LLM, and increasing the number of anchors leads to a more precise approximation of the optimal parameter state for any target SLM.

\section{Conclusion}
In this paper, we introduced \textbf{Chain-based Distillation (CBD)}, a transformative paradigm for the scalable initialization of variable-sized SLMs. By building a structured \textbf{knowledge chain} and utilizing a \textbf{bridging mechanism} for heterogeneous architectures, CBD successfully addresses the twin challenges of capacity gaps and structural mismatches. Our interpolation-based technique allows for the generation of high-quality, well-initialized models at $\mathcal{O}(1)$ cost relative to the number of target models. Empirical results across multiple LLM families and benchmarks demonstrate that CBD not only saves billions of training tokens but also significantly accelerates convergence and enhances performance, which is robust for rapid deployment.

\section*{Impact Statement}
This work does not involve human subjects, personal data, or other sensitive information. 
All datasets used in the experiments are publicly available, and we adhered to standard practices for data usage and reporting. 
The methods and findings are intended solely for academic research and are not designed for harmful applications. 
We ensured that the experimental protocols, theoretical derivations, and released source code conform to the principles of fairness, transparency, and reproducibility. 
No conflicts of interest or ethical concerns are associated with this work.

\bibliography{example_paper}
\bibliographystyle{icml2026}

\newpage
\appendix
\onecolumn

We organize the appendix as follows.
\begin{itemize}
    \item In \cref{app:theoretical}: the theoretical analysis for the stepwise distillation under the homogeneous and heterogeneous source LLMs. 
     \item In \cref{app:ex_details}: the experimental details.
     \item In \cref{app:datasets}: the detailed information of datasets employed in the main paper. 
     \item In \cref{app:architecture}: detailed architectures of the initialized SLMs.
     \item In \cref{app:token_saving_value}: the detailed values in \cref{fig:token_saving_comparison} in the main paper.
     \item In \cref{app:cbd_vs_rand_llama3_qwen3}: more comparison of CBD and Rand on the source LLMs: Llama3-8B and Qwen3-4B, and the results of SLM with 537M parameters.
     \item In \cref{app:convergence}: an additional comparison of the convergence speed between SLMs initialized with CBD and those trained from scratch (Rand) under various token budgets.
     \item In \cref{app_performance_gap}: the performance gap between the CBD and random initialization (Rand) with the increase of the pre-training tokens on the Dolly task.
     \item In \cref{app_previous_learngene}: a comprehensive discussion regarding the technical constraints and the empirical challenges for transferring the learngene methods into the setting of LLMs.
     \item In \cref{app_qwen3_col}: the comparison of Qwen3-Based CBD and Rand.
     \item In \cref{app_compare_hyper_fslm}: the similarities and differences with other scalable models.
     \item In \cref{app_extreme_scales}: the performance of CBD at very small SLM scales.
     \item In \cref{app_role_bridge}: the performance comparison of SLMs built from the GPT2-XL and Qwen3-4B on the generation tasks.
     \item In \cref{app_limit}, the limitation and future work of the proposed initialization way.

\end{itemize}

\section{Theoretical Analysis}
\label{app:theoretical}

We provide a unified theoretical analysis for $k$-level stepwise distillation 
as employed in Chain-based Distillation. 
This section first introduces notation and the general single-step learning bound, 
then analyzes the homogeneous (same-architecture) case, 
and finally extends the results to the heterogeneous (cross-architecture) case.

\subsection{Unified notation and preliminaries}

Let the ground-truth target function be $f^\star$. 
We denote hypothesis classes in a distillation chain as
\[
\mathcal{G} \;\;\to\;\; \mathcal{H}_1 \;\;\to\;\; \mathcal{H}_2 \;\;\to\;\; \cdots \;\;\to\;\; \mathcal{H}_k \;\;\to\;\; \mathcal{S},
\]
where $\mathcal{G}$ is the teacher (possibly heterogeneous), $\mathcal{H}_i$ are intermediate assistants, 
and $\mathcal{S}$ is the final student.

For any hypothesis class $\mathcal{X}$ we denote:
\begin{itemize}
  \item $C_{\mathcal{X}}$: a capacity measure;
  \item $\rho_{\mathcal{X}\leftarrow \mathcal{Y}}$: the statistical convergence rate when $\mathcal{X}$ learns from $\mathcal{Y}$;
  \item $\varepsilon_{\mathcal{X}\leftarrow \mathcal{Y}}$: the approximation error when $\mathcal{X}$ mimics $\mathcal{Y}$.
\end{itemize}

\textbf{General single-step bound.}
For any pair $(\mathcal{X},\mathcal{Y})$,
\begin{equation}
R(\hat{f}_{\mathcal{X}}) - R(f_{\mathcal{Y}})
\;\;\le\;\; O\!\left(\frac{C_{\mathcal{X}}}{n^{\rho_{\mathcal{X}\leftarrow\mathcal{Y}}}}\right) + \varepsilon_{\mathcal{X}\leftarrow\mathcal{Y}}.
\label{eq:basic-bound}
\end{equation}

\subsection{Stepwise distillation with homogeneous source LLMs}
\label{app:theory-homo}

\textbf{Direct distillation.}
Training $\mathcal{S}$ directly from $\mathcal{G}$ gives
\begin{equation}
R(\hat{s}) - R(f^\star)
\;\;\le\;\;
O\!\left(\frac{C_{\mathcal{G}}}{n^{\rho_{\mathcal{G}\leftarrow f^\star}}}
+ \frac{C_{\mathcal{S}}}{n^{\rho_{\mathcal{S}\leftarrow \mathcal{G}}}}\right)
+ \varepsilon_{\mathcal{G}\leftarrow f^\star} + \varepsilon_{\mathcal{S}\leftarrow \mathcal{G}}.
\label{eq:direct-homo}
\end{equation}

\textbf{Stepwise distillation.}
Proceeding through $\mathcal{H}_1,\dots,\mathcal{H}_k$ yields
\begin{equation}
\begin{aligned}
R(\hat{s}) - R(f^\star)
\;\;\le\;\;&
O\!\Bigg(
\frac{C_{\mathcal{G}}}{n^{\rho_{\mathcal{G}\leftarrow f^\star}}}
+ \sum_{i=1}^k \frac{C_{\mathcal{H}_i}}{n^{\rho_{\mathcal{H}_i\leftarrow \mathcal{H}_{i-1}}}}
+ \frac{C_{\mathcal{S}}}{n^{\rho_{\mathcal{S}\leftarrow \mathcal{H}_k}}}
\Bigg) \\
&+ \varepsilon_{\mathcal{G}\leftarrow f^\star}
+ \sum_{i=1}^k \varepsilon_{\mathcal{H}_i\leftarrow \mathcal{H}_{i-1}}
+ \varepsilon_{\mathcal{S}\leftarrow \mathcal{H}_k},
\end{aligned}
\label{eq:multi-homo}
\end{equation}
where $\mathcal{H}_0 := \mathcal{G}$.

\begin{assumption}[Rate improvement for homogeneous case]
For all $i$,
\[
\rho_{\mathcal{S}\leftarrow \mathcal{G}} \;\le\; 
\min\!\left(\rho_{\mathcal{H}_i\leftarrow \mathcal{H}_{i-1}},\, \rho_{\mathcal{S}\leftarrow \mathcal{H}_k}\right).
\]
\end{assumption}

\begin{assumption}[Approximation decomposition for homogeneous case]
\[
\sum_{i=1}^k \varepsilon_{\mathcal{H}_i\leftarrow \mathcal{H}_{i-1}}
+ \varepsilon_{\mathcal{S}\leftarrow \mathcal{H}_k}
\;\;\le\;\; \varepsilon_{\mathcal{S}\leftarrow \mathcal{G}}.
\]
\end{assumption}

\textbf{Conclusion.}
Under these assumptions, the multi-step bound \eqref{eq:multi-homo} is asymptotically tighter than the direct bound \eqref{eq:direct-homo}, establishing the theoretical advantage of homogeneous stepwise distillation.

\subsection{Stepwise distillation with heterogeneous source LLMs}
\label{app:theory-hetero}

Now consider the case where the original large teacher belongs to a different architecture family $\mathcal{G}^{(A)}$ (e.g., Llama3-8B, Qwen3-4B).  
We insert a GPT2-XL ancher, denoted $\mathcal{G}^{(\mathrm{GPT2})}$, as a bridging assistant before continuing with GPT-2 style anchors.

\textbf{Additional notation.}
\begin{itemize}
  \item Mapping operator $\Pi_{\mathrm{map}}$: projects $\mathcal{G}^{(A)}$ outputs into the GPT-2 output space.
  \item Mapping residual $\delta_{\mathrm{map}}$: additional approximation error from projection mismatch.
\end{itemize}
The error from $\mathcal{G}^{(A)}$ to $\mathcal{G}^{(\mathrm{GPT2})}$ decomposes as
\begin{equation}
\varepsilon_{\mathcal{G}^{(\mathrm{GPT2})}\leftarrow \mathcal{G}^{(A)}}
= \varepsilon_{\mathcal{G}^{(\mathrm{GPT2})}\leftarrow \Pi_{\mathrm{map}}(\mathcal{G}^{(A)})}
+ \delta_{\mathrm{map}}.
\label{eq:map-decomp}
\end{equation}

\textbf{Heterogeneous stepwise bound.}
Let $\mathcal{H}_{-1}:=\mathcal{G}^{(A)}$ and $\mathcal{H}_0:=\mathcal{G}^{(\mathrm{GPT2})}$, then
\begin{equation}
\begin{aligned}
R(\hat{s}) - R(f^\star)
\le\;&
O\!\Bigg(\frac{C_{\mathcal{G}^{(A)}}}{n^{\rho_{\mathcal{G}^{(A)}\leftarrow f^\star}}}
+ \sum_{i=0}^{k} \frac{C_{\mathcal{H}_i}}{n^{\rho_{\mathcal{H}_i\leftarrow \mathcal{H}_{i-1}}}}
+ \frac{C_{\mathcal{S}}}{n^{\rho_{\mathcal{S}\leftarrow \mathcal{H}_k}}}\Bigg) \\
&+ \varepsilon_{\mathcal{G}^{(A)}\leftarrow f^\star}
+ \sum_{i=0}^{k} \varepsilon_{\mathcal{H}_i\leftarrow \mathcal{H}_{i-1}}
+ \varepsilon_{\mathcal{S}\leftarrow \mathcal{H}_k},
\end{aligned}
\label{eq:multi-hetero}
\end{equation}
where $\varepsilon_{\mathcal{H}_0\leftarrow \mathcal{H}_{-1}}$ includes $\delta_{\mathrm{map}}$.

\begin{assumption}[Rate improvement across mapped chain]
\[
\rho_{\mathcal{S}\leftarrow \mathcal{G}^{(A)}} \;\le\; 
\min\!\big(\rho_{\mathcal{H}_i\leftarrow \mathcal{H}_{i-1}} \;:\; i=0,\dots,k,\, \rho_{\mathcal{S}\leftarrow \mathcal{H}_k}\big).
\]
\end{assumption}

\begin{assumption}[Approximation decomposition with mapping error]
\[
\delta_{\mathrm{map}}
+ \sum_{i=0}^{k} \varepsilon_{\mathcal{H}_i\leftarrow \mathcal{H}_{i-1}}
+ \varepsilon_{\mathcal{S}\leftarrow \mathcal{H}_k}
\;\;\le\;\; \varepsilon_{\mathcal{S}\leftarrow \mathcal{G}^{(A)}}.
\]
\end{assumption}

\textbf{Conclusion.}
Under these assumptions, the heterogeneous bound \eqref{eq:multi-hetero} is asymptotically tighter than direct distillation from $\mathcal{G}^{(A)}$ to $\mathcal{S}$. 
This formalizes the role of GPT2-XL as a bridging assistant for cross-architecture stepwise distillation.

\section{Experimental Details}
\label{app:ex_details}

\subsection{Training Details}

Our training pipeline consists of four main stages: (1) homogeneous distillation for constructing the knowledge chain, (2) heterogeneous distillation, (3) downstream fine-tuning of SLMs, and (4) pre-training of randomly initialized SLMs on OpenWebText.

\textbf{Homogeneous Distillation.} We perform homogeneous distillation in two phases to construct the knowledge chain using architectures identical to the target model. In the first phase, both the teacher model and intermediate anchors are supervised-fine-tuned (SFT) on the Dolly dataset for 3 epochs to align model behaviors. In the second phase, we adopt a policy distillation scheme inspired by MiniLLM. Specifically, we employ reverse KL divergence between the teacher and student policies on Dolly, collecting 256 samples per batch and running 4 inner optimization epochs per batch. The clipping parameter $\epsilon$ is set to 0.2, and the sequence length is capped at 512 tokens. During distillation, we sample from the teacher with a temperature of 1. The distillation proceeds for 5000 steps, and we select the final model checkpoint based on Rouge-L score on the validation set.

\textbf{Heterogeneous Distillation.} For cross-architecture distillation, we first fine-tune two heterogeneous teacher models: Qwen-4B and LLaMA3-8B, on the Dolly dataset for 3 epochs using a batch size of 2 and a learning rate of 1e-5. We then use the final checkpoints of these teachers to generate pseudo answers given Dolly questions. These synthetic question-answer pairs are used to fine-tune a smaller proxy model (GPT2-XL) for 10 epochs, with a batch size of 8 and learning rate of 1e-5. Subsequently, we distill from the proxy teacher into the knowledge chain following the same homogeneous strategy described above.

\textbf{SLMs Fine-tuning.} Once the knowledge chain is constructed and assembled into target SLM architectures, we fine-tune the resulting SLMs on downstream tasks for 10 epochs. We use a batch size of 16 or 8 and a learning rate of 1e-5. The model parameters from the final epoch are used for performance evaluation.

\textbf{Pre-training from Scratch.} For baseline comparison, we pre-train SLMs from scratch on the OpenWebText corpus. We train for 2 epochs using a learning rate of 5e-4 and a batch size of 2. In this setting, “78M tokens” refers to the full OpenWebText-100K subset (approximately 78M tokens), while “100M”, “500M”, and “10B tokens” denote the first 100 million, 500 million, and 10 billion tokens extracted sequentially from the full OpenWebText dataset, respectively. 

\textbf{Fine-tuning Prompts.}
Here, we present the prompt templates used in all experiments. 
For clarity, we abstract away model-specific formatting tokens and show only the semantic content of the prompts.
In all experiments, we use the following instruction-style prompts.

For samples containing only the article title, the prompt is:

\begin{verbatim}
Below is an instruction that describes a task. Write a response that 
appropriately completes the request.

### Instruction:
Given the following article, and the title of the article is {instruction}, 
please summarize the article.

### Response:
\end{verbatim}

For samples that additionally contain the full article text, the prompt is:

\begin{verbatim}
Below is an instruction that describes a task, paired with an input that 
provides further context. Write a response that appropriately completes 
the request.

### Instruction:
Given the following article, and the title of the article is {instruction}, 
please summarize the article.

### Input:
{input}

### Response:
\end{verbatim}

\subsection{Evaluation Metrics}
We use Rouge-L to evaluate the performance of our model on generation tasks such as summarization. It measures the longest common subsequence (LCS) between the generated text and the reference summary, which captures sentence-level fluency and coherence better than n-gram-based metrics. Unlike Rouge-n, Rouge-L does not require consecutive matches but considers the in-sequence matches, making it more appropriate for abstractive summarization. The Rouge-L score is defined as the F-measure of the LCS-based precision and recall:
\begin{equation}
\text{Rouge-L} = \frac{(1 + \beta^2) \cdot \text{Precision} \cdot \text{Recall}}{\text{Recall} + \beta^2 \cdot \text{Precision}},
\end{equation}
where $\text{Precision} = \frac{\text{LCS}(X, Y)}{|X|}$, $\text{Recall} = \frac{\text{LCS}(X, Y)}{|Y|}$, $X$ and $Y$ represent the generated and reference summaries respectively, and $\beta$ is typically set to 1.

We also adopt Accuracy (Acc) as the evaluation metric for classification-based tasks such as MMLU and BoolQ. It measures the proportion of correctly predicted instances over the total number of examples. Given a dataset with $N$ samples, the accuracy is computed as:
\begin{equation}
\text{Accuracy} = \frac{1}{N} \sum_{i=1}^{N} \mathbb{I}(\hat{y}_i = y_i),
\end{equation}
where $\hat{y}_i$ is the predicted label, $y_i$ is the ground-truth label, and $\mathbb{I}(\cdot)$ is the indicator function that returns 1 if the argument is true and 0 otherwise. Note that, in the main paper, we replaced Accuracy (Acc) with Rouge-L to maintain consistency and conciseness across the tables. This substitution is justified as Rouge-L yields results consistent with Accuracy for multiple-choice and true/false questions.

\subsection{Hardware}
All results reported in this paper are conducted on HUAWEI ASCEND 910B (64G) NPUs. 

\section{Detailed Information of Datasets}
\label{app:datasets}
This section details the information of the datasets adopted in the main paper. As we described in the paper, we benchmark on six datasets: MMLU, English\_XLSum, HellaSwag, WinoGrande, BoolQ, and Dolly.

\subsection{MMLU}
The Massive Multitask Language Understanding (MMLU) is designed to evaluate the general knowledge and reasoning abilities of language models across 57 diverse subjects, ranging from humanities and social sciences to professional domains. Each question is multiple-choice and intended to reflect educational or professional scenarios. MMLU emphasizes zero-shot and few-shot settings, making it a widely adopted standard for assessing cross-domain generalization in foundation models.

\subsection{English\_XLSum}
English\_XLSum is a single-document summarization dataset comprising 226{,}711 BBC news articles and corresponding one-sentence summaries. The task focuses on generating concise and abstractive summaries that answer the question “What is this article about?”. The dataset covers a broad range of topics and is split into training (204{,}045), validation (11{,}332), and test (11{,}334) sets.

\subsection{HellaSwag}
HellaSwag is a benchmark for commonsense natural language inference, consisting of 70{,}000 multiple-choice sentence completion problems. While trivial for humans (95\%+ accuracy), HellaSwag poses significant challenges for language models due to its adversarial filtering (AF) construction. The dataset was introduced to expose the limitations of deep pre-trained models in commonsense reasoning and has become a cornerstone for evaluating robustness under ambiguous and nuanced scenarios.

\subsection{WinoGrande}
WinoGrande is a large-scale benchmark inspired by the Winograd Schema Challenge. It consists of 44{,}000 fill-in-the-blank problems designed to test commonsense reasoning. By scaling up the number of examples and minimizing dataset-specific biases, WinoGrande offers a more robust measure of a model’s ability to resolve coreference ambiguities using real-world knowledge.

\subsection{BoolQ}
BoolQ is a yes/no question answering dataset containing 15{,}942 examples, collected in a naturally occurring, unprompted setting. Each instance includes a question, a passage, and a binary answer. The task setup closely resembles natural language inference and is widely used to evaluate a model's capacity for contextual understanding and factual reasoning.

\subsection{Dolly}
Dolly, developed by Databricks, is an instruction-following language model based on the pythia-12B architecture and fine-tuned on approximately 15{,}000 instruction-response pairs (databricks-dolly-15k). The data covers diverse task types such as classification, generation, QA, and summarization. While not state-of-the-art, Dolly exhibits strong instruction-following behavior and is fully open-source and commercially licensed, making it a valuable model for experimentation and adaptation in real-world scenarios.

We choose the \textbf{Dolly} dataset as the training corpus for constructing the CBD due to its broad coverage of language capabilities and proven effectiveness in previous distillation research such as MiniLLM. Dolly spans a diverse range of task types, including brainstorming, classification, closed-form QA, open-ended QA, summarization, generation, and information extraction. This diversity enables CBD to capture reusable knowledge blocks that generalize across different task formats. 
In contrast, most other datasets such as MMLU, BoolQ, and English\_XLSum focus on narrower task types—e.g., closed QA or summarization—which may limit the diversity of distilled knowledge. Leveraging a multi-capability dataset like Dolly is thus critical for building a more general and robust CBD.

\section{Detailed Architectures of SLMs}
\label{app:architecture}
In this section, we list the detailed information of the SLMs shown in the main paper. Specifically, we show the number of layers, multi-head attentions, hidden dimensions and the intermediate dimensions of the FFN layers. The structure of each SLM is described in \cref{tab:struct}. 

\begin{table}[t]
\centering
\caption{The detailed structure of the SLMs in the main paper. `P' denotes the parameters of the SLM, `$\text{He}_{Dim}$' is the dimension of each multi-head attention, `$\text{Hidden}_{Dim}$' means the dimension of the hidden size, and `$\text{I}_{Dim}$' is the intermediate dimension.}
\begin{tabular}{c|c|c|c|c|c}
\toprule
P(M) & Layer & Head & $\text{He}_{Dim}$ & $\text{Hidden}_{Dim}$ & $\text{I}_{Dim}$ \\ \midrule
138       & 14    & 12   & 64       & 768        & 3072             \\ \midrule
220       & 18    & 14   & 64       & 896        & 3584             \\ \midrule
277       & 24    & 14   & 64       & 896        & 3584             \\ \midrule
380       & 26    & 16   & 64       & 1024       & 4096             \\ \midrule
537       & 30    & 18   & 64       & 1152       & 4608             \\ \bottomrule
\end{tabular}
\label{tab:struct}
\end{table}

\section{Detailed Values}
\label{app:token_saving_value}
Here, we show the detailed values of \cref{fig:token_saving_comparison} in the main paper, as shown in \cref{tab:detail_results}. The detailed results further validate our view that CBD can effectively save the training corpus required for pre-training, regardless of whether the source LLMs are homogeneous (GPT2-XL) or heterogeneous (Llama3-8B, Qwen3-4B).

\begin{table}[]
\caption{The detailed values of \cref{fig:token_saving_comparison} in the main paper.}
\resizebox{\textwidth}{!}{
\begin{tabular}{cccccccccc}
\toprule
\multicolumn{1}{c|}{\multirow{2}{*}{Task}} & \multicolumn{3}{c|}{CBD}                            & \multicolumn{6}{c}{Rand}                       \\
\multicolumn{1}{c|}{}                      & GPT2-XL & Llama3-8B & \multicolumn{1}{c|}{Qwen3-4B} & 0.078B & 0.1B  & 0.5B  & 1B    & 5B    & 10B   \\ \midrule
\multicolumn{10}{c}{SLM-138M}                                                                                                                  \\ \midrule
\multicolumn{1}{c|}{English\_XLsum}        & 32.19   & 33.57     & \multicolumn{1}{c|}{31.35}    & 26.18  & 26.95 & 31.28 & 30.71 & 33.52 & 35.26 \\
\multicolumn{1}{c|}{HellaSwag}             & 30.59   & 39.59     & \multicolumn{1}{c|}{42.01}    & 23.68  & 23.68 & 29.47 & 31.81 & 29.67 & 29.27 \\
\multicolumn{1}{c|}{WinoGrande}            & 63.20   & 53.72     & \multicolumn{1}{c|}{60.20}    & 51.50  & 52.50 & 54.80 & 56.80 & 56.80 & 55.40 \\ \midrule
\multicolumn{10}{c}{SLM-380M}                                                                                                                  \\ \midrule
\multicolumn{1}{c|}{English\_XLsum}        & 39.29   & 37.15     & \multicolumn{1}{c|}{36.31}    & 22.20  & 22.62 & 31.45 & 34.40 & 35.87 & 40.06 \\
\multicolumn{1}{c|}{HellaSwag}             & 47.47   & 35.90     & \multicolumn{1}{c|}{46.64}    & 38.54  & 36.96 & 41.00 & 40.71 & 45.93 & 46.55 \\
\multicolumn{1}{c|}{WinoGrande}            & 68.00   & 49.48     & \multicolumn{1}{c|}{67.30}    & 36.90  & 35.30 & 51.10 & 58.90 & 59.20 & 59.40 \\ \bottomrule
\end{tabular}}
\label{tab:detail_results}
\end{table}

\section{CBD \textit{vs.} Rand on Llama3-8B and Qwen3-4B}
\label{app:cbd_vs_rand_llama3_qwen3}
This section provide the extended results of SLMs intialized by CBD and Rand on the source LLM: Llama3-8B and Qwen3-4B, as in \cref{tab:cbd_vs_rand_llama3,tab:cbd_vs_rand_qwen3}. Here, we can find the similar phenomenon that CBD provides good initalization for variable-sized SLMs to achieve better performance. 
Additionally, we also show the results of SLMs with more parameters (537M) in \cref{tab:cbd_vs_rand_537M}. We can find that CBD still achieves better even in the settings that SLM has more parameters.

\begin{table}[]
  \centering
  \caption{The comparison of CBD and Random initialization of variable-sized SLMs on the source LLM: Llama3-8B.}
  \label{tab:cbd_vs_rand_llama3}
  \resizebox{\textwidth}{!}{
  \begin{tabular}{c|c|cccccc|cccccc}
  \toprule
  \multirow{2}{*}{Token Num} & \multirow{2}{*}{Init} & \multicolumn{6}{c|}{SLM-138M}                                                & \multicolumn{6}{c}{SLM-220M}                                                 \\ \cmidrule{3-14} 
  &                       & MMLU  & XLsum & HellaS & WinoG & \multicolumn{1}{c|}{BoolQ} & Avg            & MMLU  & XLsum & HellaS & WinoG & \multicolumn{1}{c|}{BoolQ} & Avg            \\ \midrule
  \multirow{2}{*}{78M}       & Rand                  & 29.00 & 26.18 & 23.68  & 51.50 & \multicolumn{1}{c|}{68.00} & 39.67          & 28.40                     & 28.48                      & 23.87                       & 49.70                      & \multicolumn{1}{c|}{66.50} & 39.39          \\
                             & CBD                   & 24.85 & 34.22 & 35.51  & 44.41 & \multicolumn{1}{c|}{64.97} & \textbf{40.79} & 24.78                     & 32.77                      & 35.37                       & 46.20                      & \multicolumn{1}{c|}{63.76} & \textbf{40.58} \\ \midrule
  \multirow{2}{*}{100M}      & Rand                  & 29.10 & 26.95 & 23.68  & 52.50 & \multicolumn{1}{c|}{67.80} & 40.01          & 29.30                     & 26.87                      & 23.67                       & 52.80                      & \multicolumn{1}{c|}{69.10} & 40.35          \\
                             & CBD                   & 24.77 & 34.20 & 35.17  & 42.90 & \multicolumn{1}{c|}{64.33} & \textbf{40.27} & 25.09                     & 34.79                      & 35.65                       & 43.22                      & \multicolumn{1}{c|}{64.90} & \textbf{40.73} \\ \midrule
  \multirow{2}{*}{Token Num} & \multirow{2}{*}{Init} & \multicolumn{6}{c|}{SLM-277M}                                                & \multicolumn{6}{c}{SLM-380M}                                                                                                                                    \\ \cmidrule{3-14} 
                             &                       & MMLU  & XLsum & HellaS & WinoG & \multicolumn{1}{c|}{BoolQ} & Avg            & MMLU                      & XLsum                      & HellaS                      & WinoG                      & \multicolumn{1}{c|}{BoolQ} & Avg            \\ \cmidrule{1-14} 
  \multirow{2}{*}{78M}       & Rand                  & 26.40 & 27.30 & 23.77  & 50.20 & \multicolumn{1}{c|}{67.50} & 39.03          & 26.70                     & 22.20                      & 38.54                       & 36.90                      & \multicolumn{1}{c|}{67.90} & 38.45          \\
                             & CBD                   & 25.86 & 33.98 & 35.65  & 43.22 & \multicolumn{1}{c|}{65.86} & \textbf{40.92} & 24.72                     & 34.51                      & 36.46                       & 40.58                      & \multicolumn{1}{c|}{63.65} & \textbf{39.99} \\ \cmidrule{3-14} 
  \multirow{2}{*}{100M}      & Rand                  & 27.90 & 27.68 & 24.39  & 47.50 & \multicolumn{1}{c|}{66.70} & 38.83          & 28.50                     & 22.62                      & 36.96                       & 35.30                      & \multicolumn{1}{c|}{65.10} & 37.70          \\
                             & CBD                   & 24.88 & 32.86 & 35.43  & 44.58 & \multicolumn{1}{c|}{64.97} & \textbf{40.54} & 24.45                     & 33.42                      & 34.42                       & 41.15                      & \multicolumn{1}{c|}{64.79} & \textbf{39.65} \\ \bottomrule
  \end{tabular}}
\end{table}

\begin{table}[]
  \centering
  \caption{The comparison of CBD and Random initialization of variable-sized SLMs on the source LLM: Qwen3-4B.}
  \label{tab:cbd_vs_rand_qwen3}
  \resizebox{\textwidth}{!}{
  \begin{tabular}{c|c|cccccc|cccccc}
  \toprule
  \multirow{2}{*}{Token Num} & \multirow{2}{*}{Init} & \multicolumn{6}{c|}{SLM-138M}                                                & \multicolumn{6}{c}{SLM-220M}                                                 \\ \cmidrule{3-14} 
                            &                       & MMLU  & XLsum & HellaS & WinoG & \multicolumn{1}{c|}{BoolQ} & Avg            & MMLU  & XLsum & HellaS & WinoG & \multicolumn{1}{c|}{BoolQ} & Avg            \\ \midrule
  \multirow{2}{*}{78M}       & Pre                   & 29.00 & 26.18 & 23.68  & 51.50 & \multicolumn{1}{c|}{68.00} & 39.67          & 28.40 & 28.48 & 23.87  & 49.70 & \multicolumn{1}{c|}{66.50} & 39.39          \\
                            & CBD                   & 26.00 & 29.83 & 42.65  & 57.60 & \multicolumn{1}{c|}{68.70} & \textbf{44.96} & 28.40 & 24.10 & 41.41  & 55.20 & \multicolumn{1}{c|}{68.80} & \textbf{43.58} \\ \midrule
  \multirow{2}{*}{100M}      & Pre                   & 29.10 & 26.95 & 23.68  & 52.50 & \multicolumn{1}{c|}{67.80} & 40.01          & 29.30 & 26.87 & 23.67  & 52.80 & \multicolumn{1}{c|}{69.10} & 40.35          \\
                            & CBD                   & 27.90 & 29.83 & 41.50  & 55.20 & \multicolumn{1}{c|}{66.00} & \textbf{44.09} & 28.00 & 30.45 & 43.49  & 56.10 & \multicolumn{1}{c|}{67.40} & \textbf{45.09} \\ \midrule
  \multirow{2}{*}{Token Num} & \multirow{2}{*}{Init} & \multicolumn{6}{c|}{SLM-277M}                                                & \multicolumn{6}{c}{SLM-380M}                                                 \\ \cmidrule{3-14} 
                            &                       & MMLU  & XLsum & HellaS & WinoG & \multicolumn{1}{c|}{BoolQ} & Avg            & MMLU  & XLsum & HellaS & WinoG & \multicolumn{1}{c|}{BoolQ} & Avg            \\ \cmidrule{1-14} 
  \multirow{2}{*}{78M}       & Pre                   & 26.40 & 27.30 & 23.77  & 50.20 & \multicolumn{1}{c|}{67.50} & 39.03          & 26.70 & 22.20 & 38.54  & 36.90 & \multicolumn{1}{c|}{67.90} & 38.45          \\
                            & CBD                   & 25.31 & 33.72 & 33.87  & 40.83 & \multicolumn{1}{c|}{68.50} & \textbf{40.44} & 24.20 & 36.31 & 44.09  & 59.80 & \multicolumn{1}{c|}{70.20} & \textbf{46.92} \\ \cmidrule{3-14} 
  \multirow{2}{*}{100M}      & Pre                   & 27.90 & 27.68 & 24.39  & 47.50 & \multicolumn{1}{c|}{66.70} & 38.83          & 28.50 & 22.62 & 36.96  & 35.30 & \multicolumn{1}{c|}{65.10} & 37.70          \\
                            & CBD                   & 28.00 & 30.45 & 42.47  & 57.60 & \multicolumn{1}{c|}{68.70} & \textbf{45.44} & 28.30 & 37.28 & 42.40  & 59.60 & \multicolumn{1}{c|}{67.40} & \textbf{47.00} \\ \bottomrule
  \end{tabular}}
\end{table}

\begin{table}[]
  \centering
  \caption{The comparison of CBD and Random initialization of SLM-537M on all source LLMs}
  \label{tab:cbd_vs_rand_537M}
  \resizebox{\textwidth}{!}{
  \begin{tabular}{c|cc|cc|cc|cc|cc|cc}
  \toprule
  \multirow{3}{*}{Task} & \multicolumn{4}{c|}{GPT2-XL}                                    & \multicolumn{4}{c|}{Llama3-8B}                                  & \multicolumn{4}{c}{Qwen3-4B}                                   \\
                        & \multicolumn{2}{c}{78M Token} & \multicolumn{2}{c|}{100M Token} & \multicolumn{2}{c}{78M Token} & \multicolumn{2}{c|}{100M Token} & \multicolumn{2}{c}{78M Token} & \multicolumn{2}{c}{100M Token} \\ \cmidrule{2-13} 
                        & CBD                & Rand     & CBD                 & Rand      & CBD                & Rand     & CBD                 & Rand      & CBD                & Rand     & CBD                & Rand      \\ \midrule
  MMLU                  & 28.10              & 26.50    & 26.30               & 25.60     & 24.75              & 26.50    & 25.52               & 25.60     & 25.30              & 26.50    & 28.40              & 25.60     \\
  XLsum                 & 30.53              & 19.14    & 34.74               & 15.15     & 34.95              & 19.14    & 34.42               & 15.15     & 35.10              & 19.14    & 35.03              & 15.15     \\
  HellaS               & 40.75              & 37.54    & 42.02               & 36.59     & 34.60              & 37.54    & 33.68               & 36.59     & 39.30              & 37.54    & 44.10              & 36.59     \\
  WinoG                & 58.90              & 33.50    & 60.20               & 34.00     & 43.08              & 33.50    & 38.20               & 34.00     & 56.90              & 33.50    & 56.80              & 34.00     \\
  BoolQ                & 69.50              & 64.70    & 71.90               & 62.80     & 62.73              & 64.70    & 64.54               & 62.80     & 68.40              & 64.70    & 73.50              & 62.80     \\ \midrule
  Avg.                  & \textbf{45.56}     & 36.28    & \textbf{47.03}      & 34.83     & \textbf{40.02}     & 36.28    & \textbf{39.27}      & 34.83     & \textbf{45.00}     & 36.28    & \textbf{47.57}     & 34.83     \\ \bottomrule
  \end{tabular}}
  \end{table}

\section{The Additional Comparison of Convergence Speed}
\label{app:convergence}
\begin{figure}[t]
\centering
\begin{minipage}{0.32\textwidth}
    \centering
    \begin{subfigure}{\linewidth}
        \includegraphics[width=\linewidth]{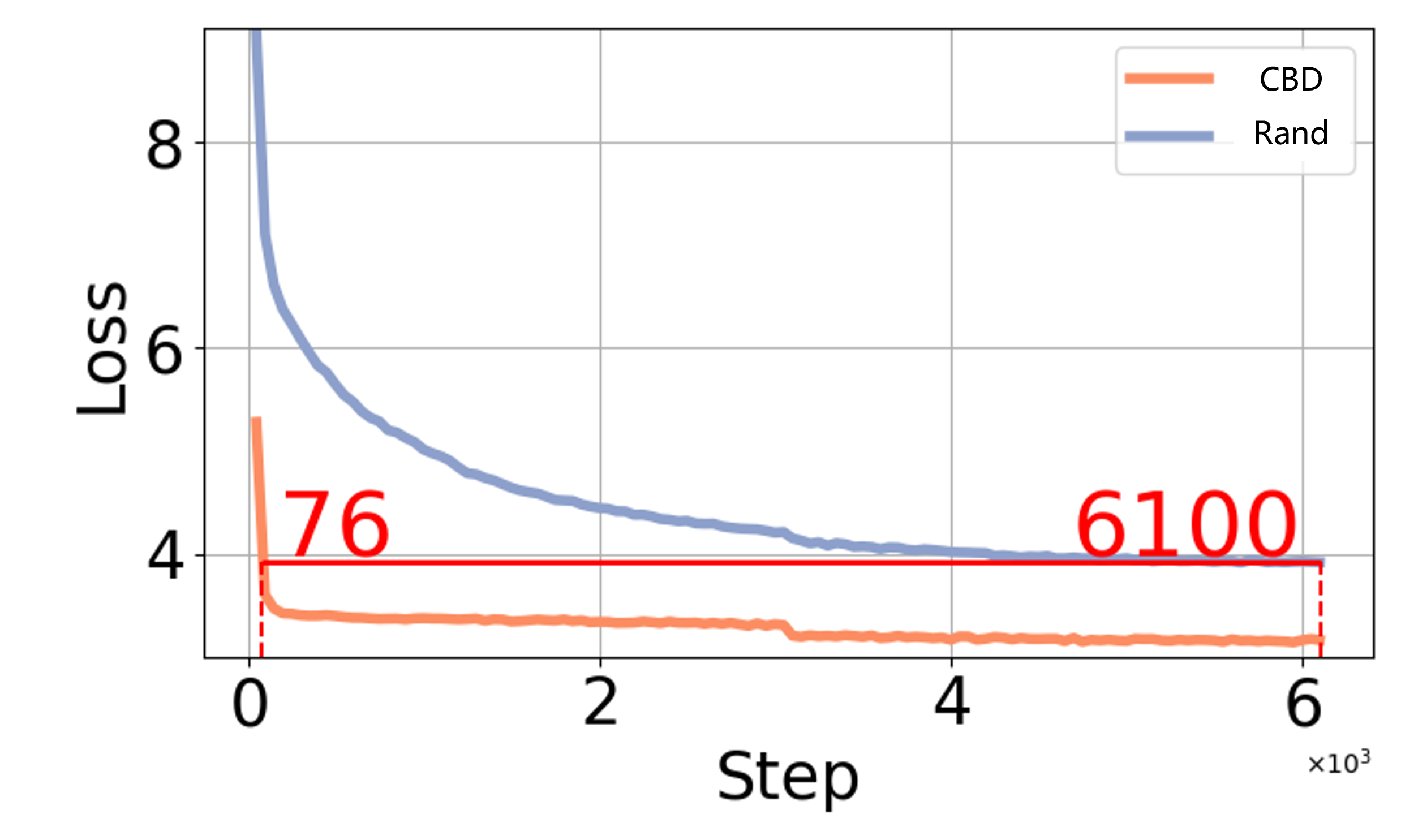}
    \caption{ GPT2-XL}    
    \end{subfigure}
\end{minipage}
\begin{minipage}{0.32\textwidth}
    \centering
    \begin{subfigure}{\linewidth}
        \includegraphics[width=\linewidth]{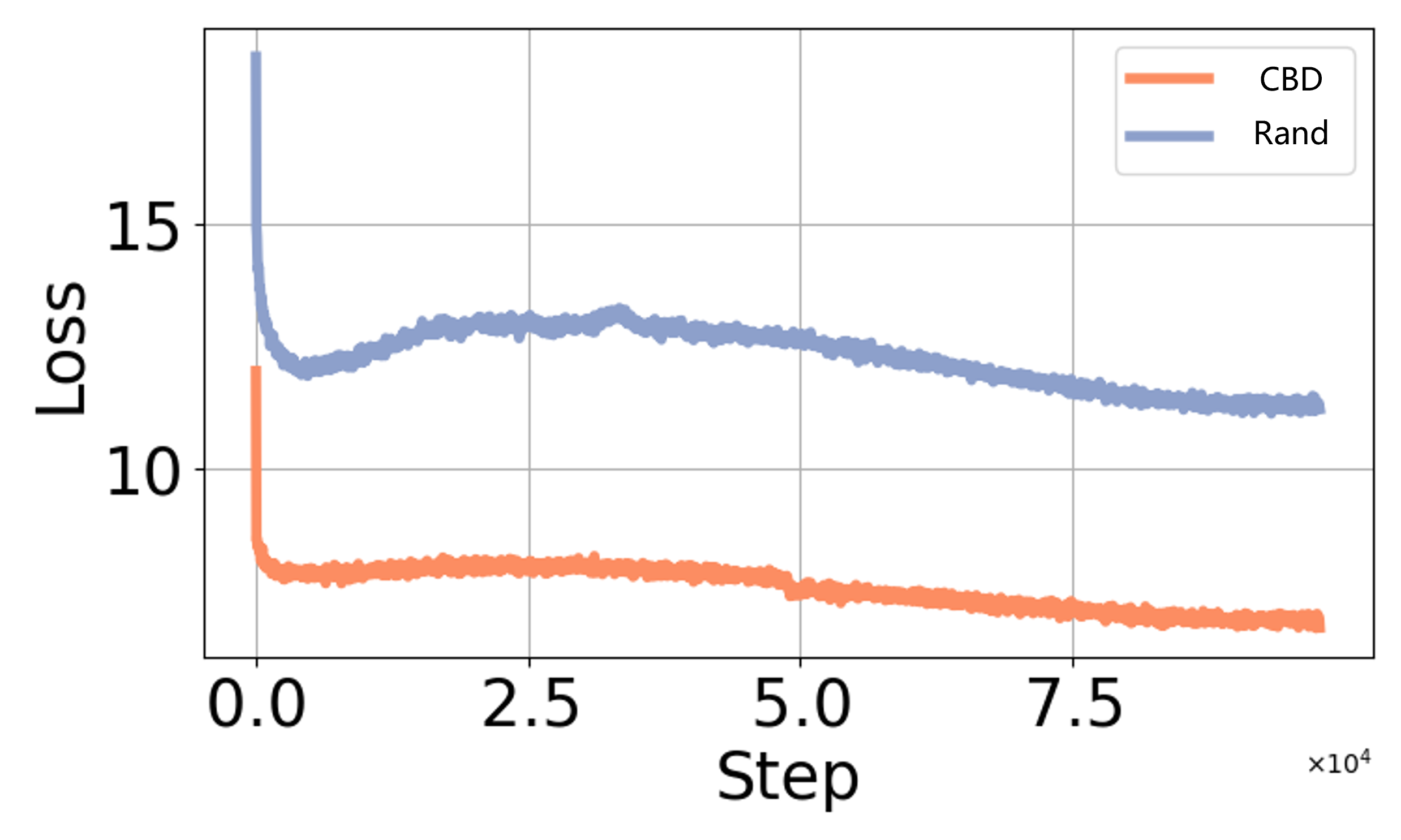}
    \caption{Llama3-8B}   
    \end{subfigure}
\end{minipage}
\begin{minipage}{0.32\textwidth}
    \centering
    \begin{subfigure}{\linewidth}
        \includegraphics[width=\linewidth]{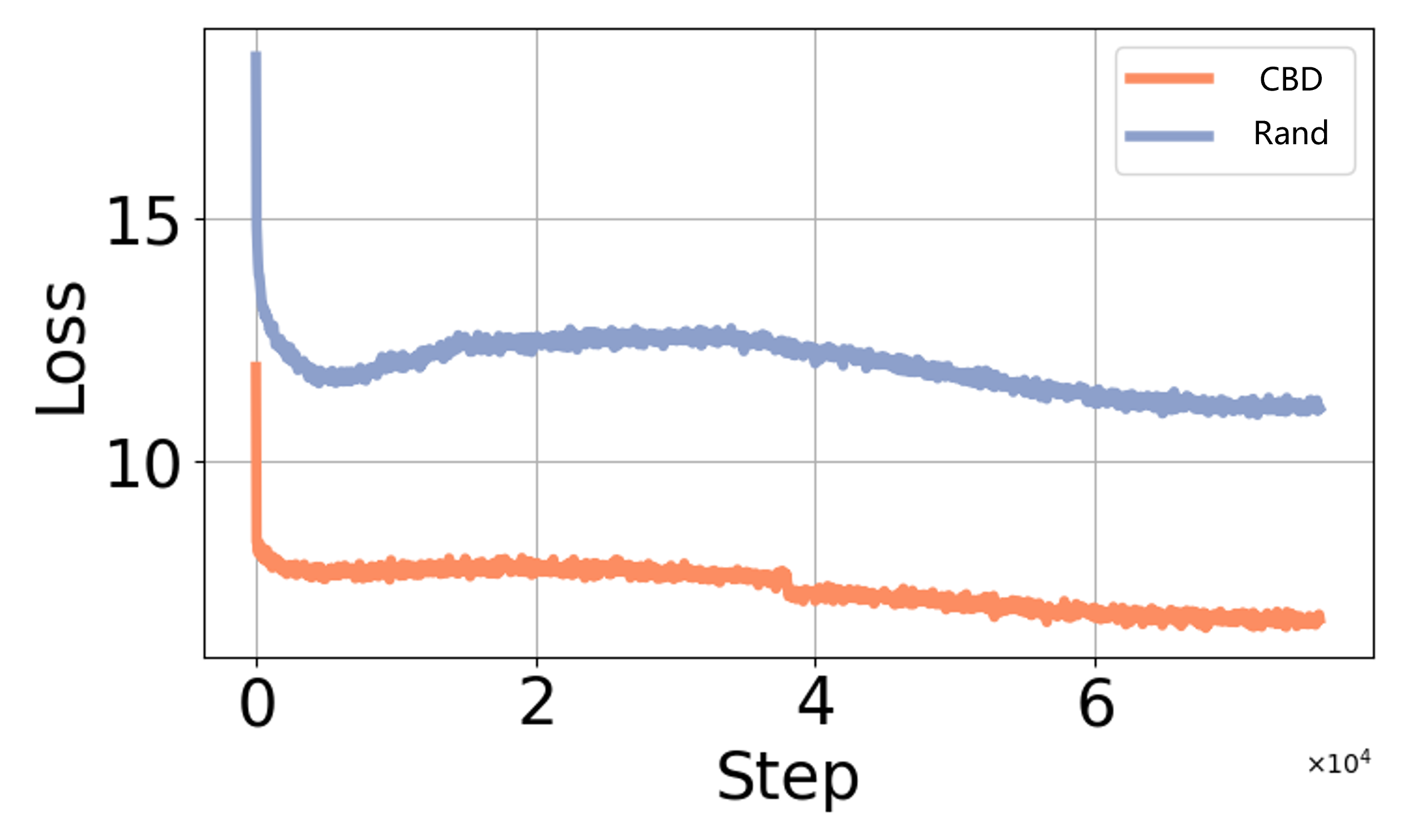}
        \caption{Qwen3-4B}
    \end{subfigure}
\end{minipage}
    \caption{Convergence speed of SLM-138M initialized by CBD and from scratch (Rand) on 100M pre-training tokens across three source LLMs.}
    \label{fig:convergence_100M}
\end{figure}

\begin{figure}[t]
\centering
\begin{minipage}{0.45\textwidth}
    \centering
    \begin{subfigure}{\linewidth}
        \includegraphics[width=\linewidth]{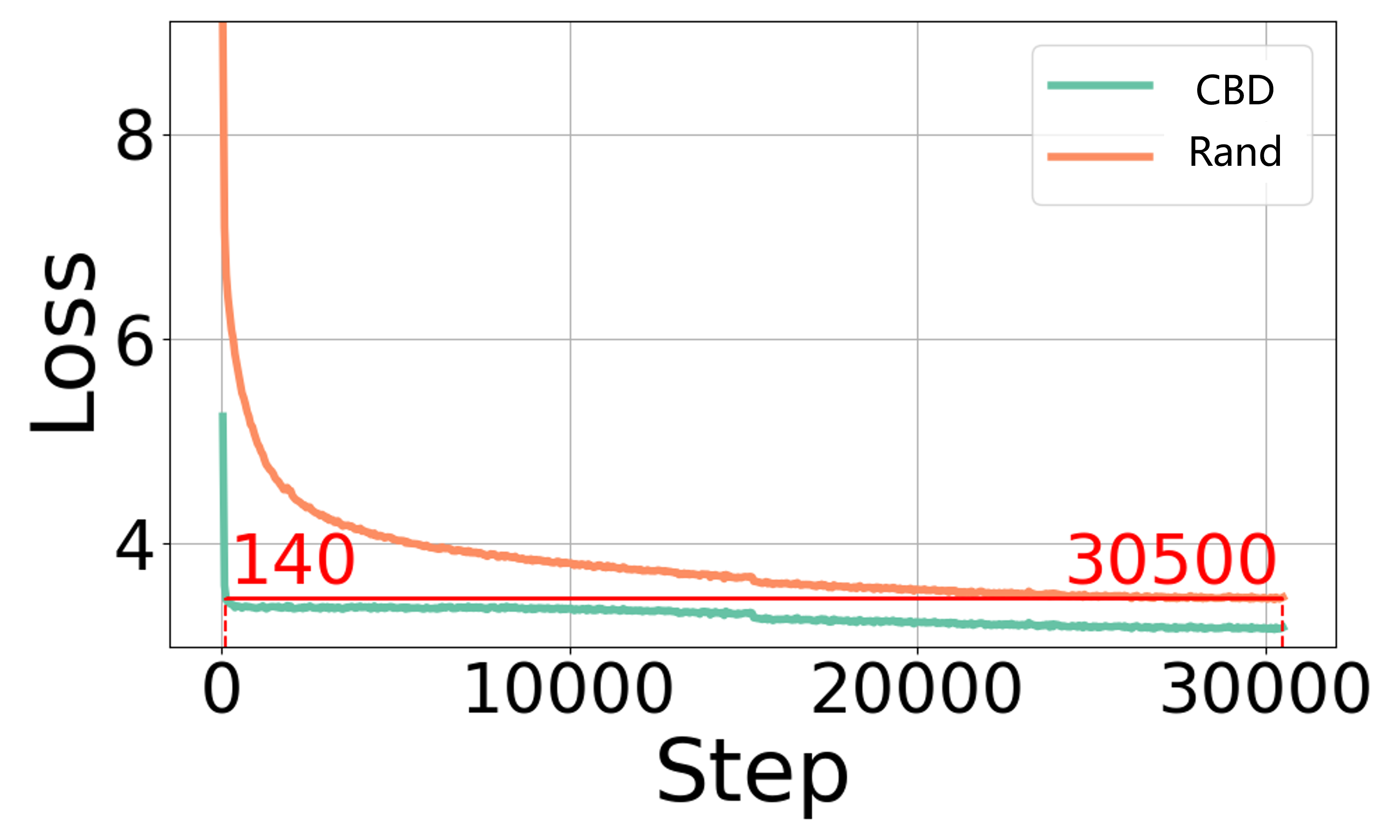}
    \caption{SLM-138M/500M Token}    
    \end{subfigure}
\end{minipage}
\begin{minipage}{0.45\textwidth}
    \centering
    \begin{subfigure}{\linewidth}
        \includegraphics[width=\linewidth]{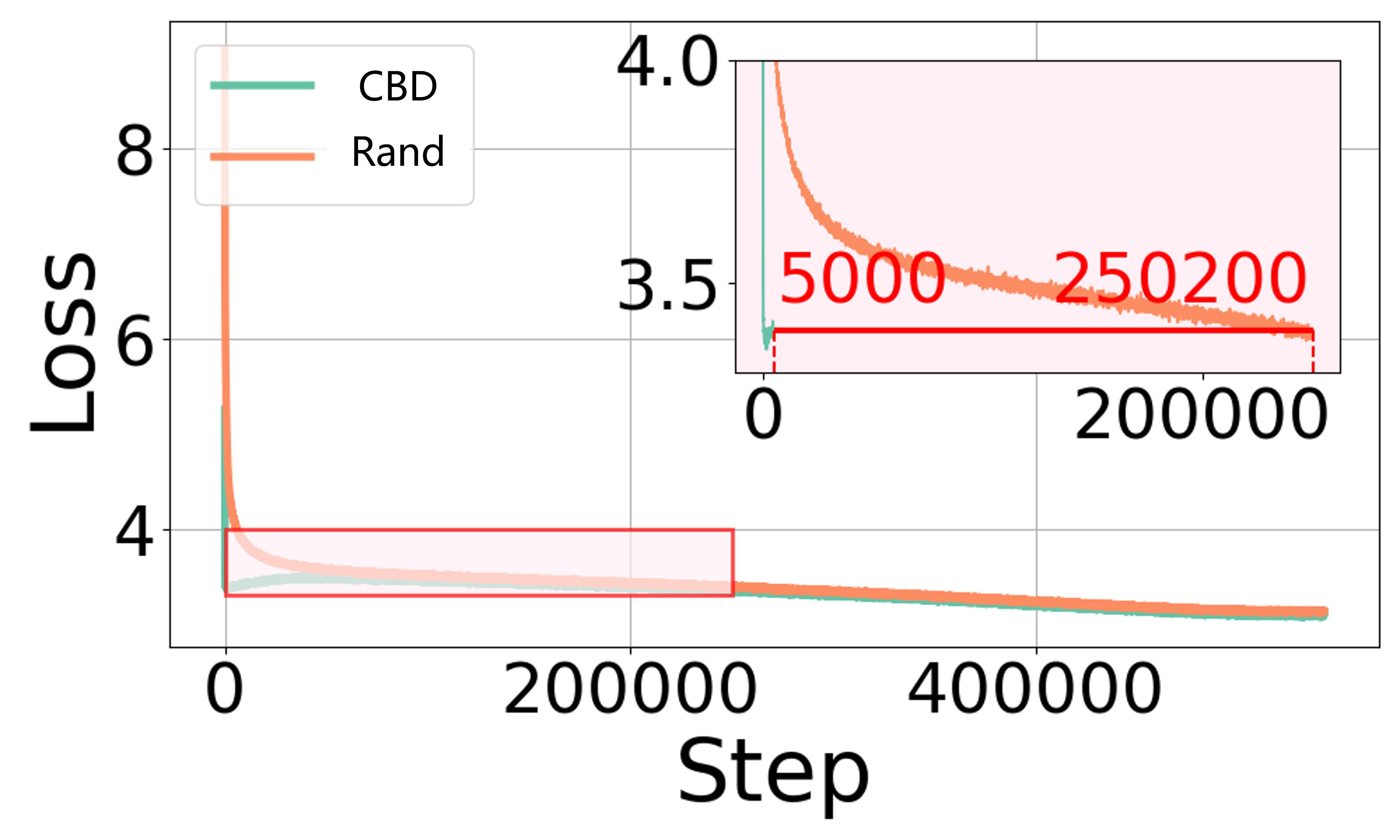}
    \caption{SLM-138M/10B Token}   
    \end{subfigure}
\end{minipage}
    \caption{Convergence speed of SLM-138M initialized by CBD and from scratch (Rand) on 500M and 10B pre-training tokens across the source LLM: GPT2-XL.}
    \label{fig:convergence_gptx_xl_1}
\end{figure}

\begin{figure}[t]
\centering
     \begin{subfigure}[b]{0.24\linewidth}
        \includegraphics[width=\linewidth]{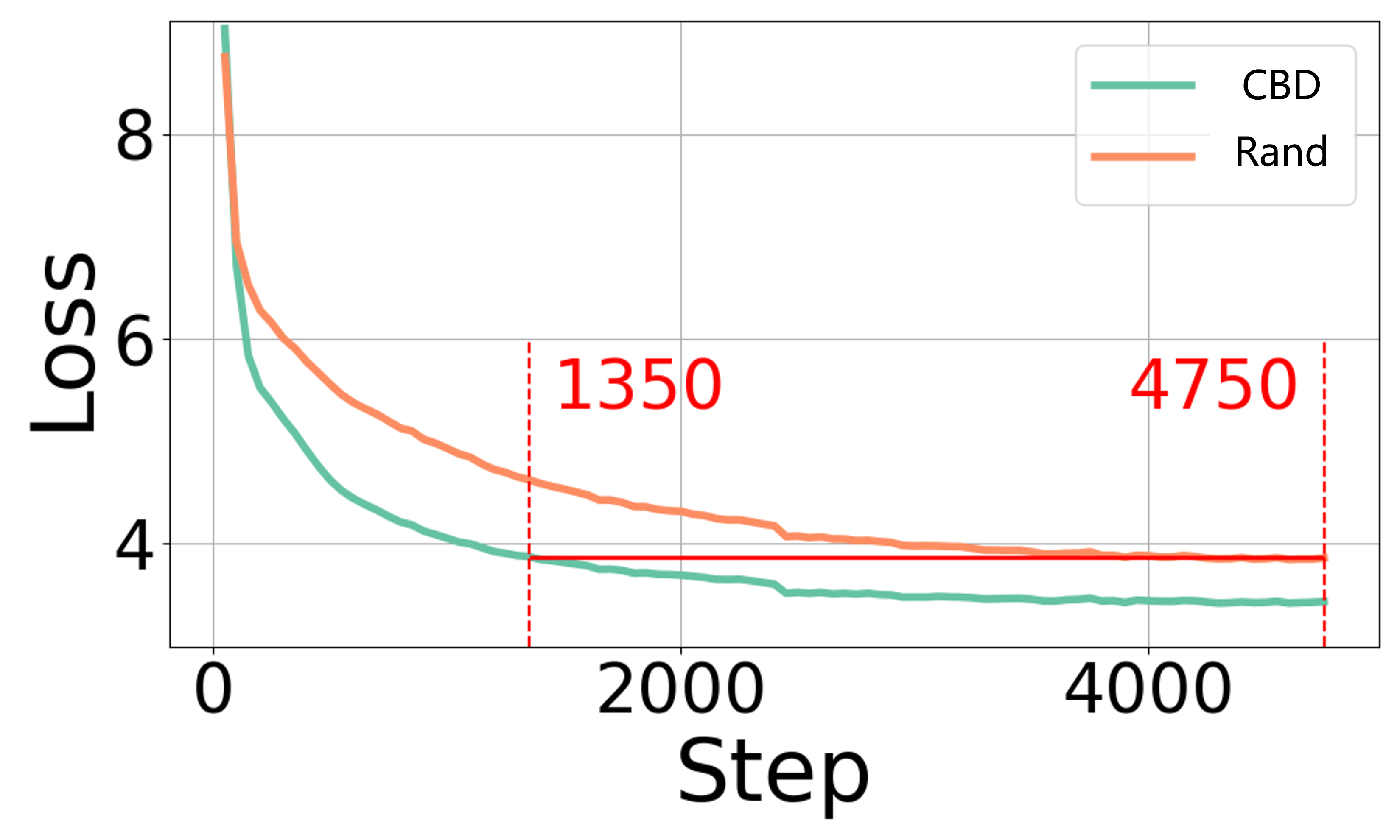}
        \caption{SLM-220M/78M T}
    \end{subfigure}
    \begin{subfigure}[b]{0.24\linewidth}
        \includegraphics[width=\linewidth]{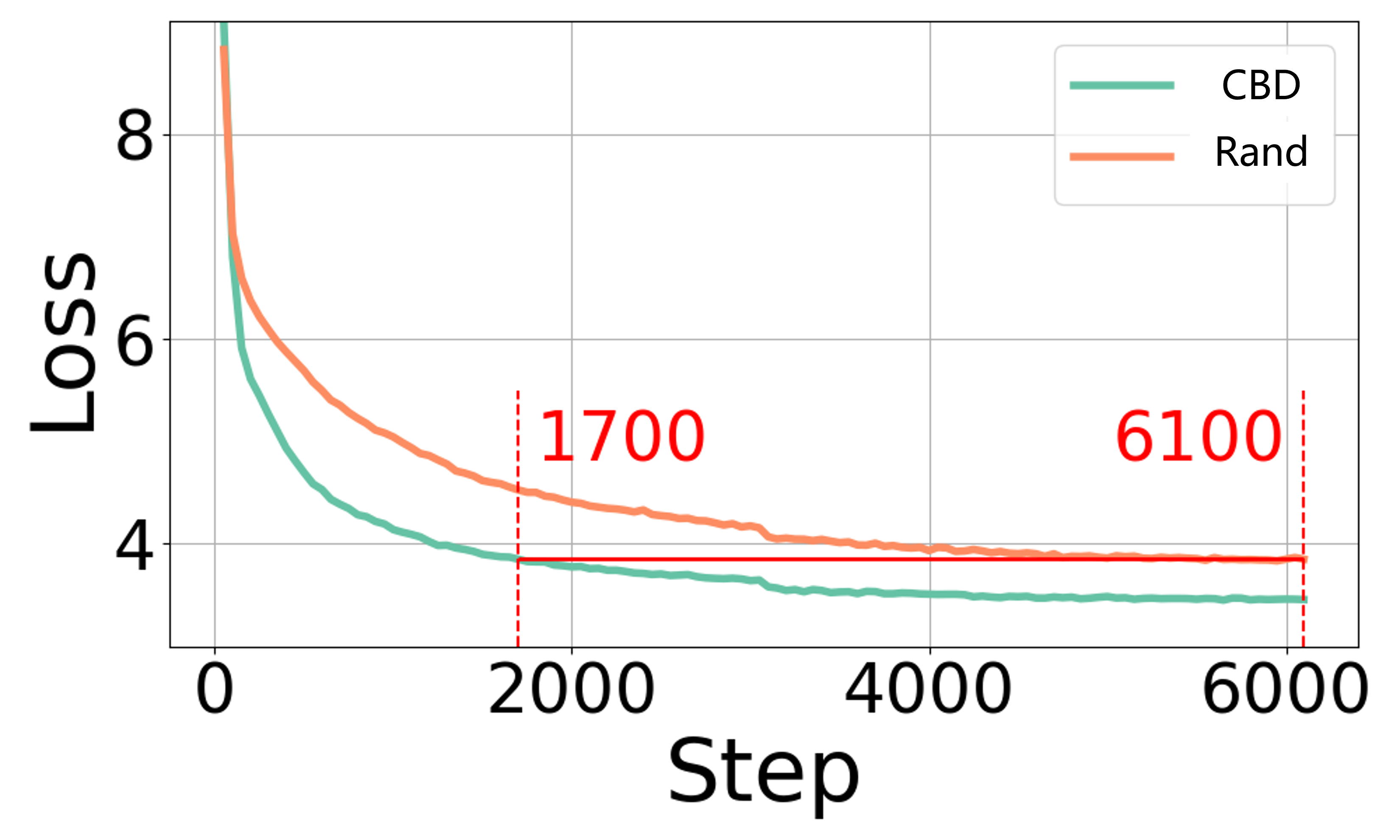}
        \caption{SLM-220M/100M T}
    \end{subfigure}
     \begin{subfigure}[b]{0.24\linewidth}
        \includegraphics[width=\linewidth]{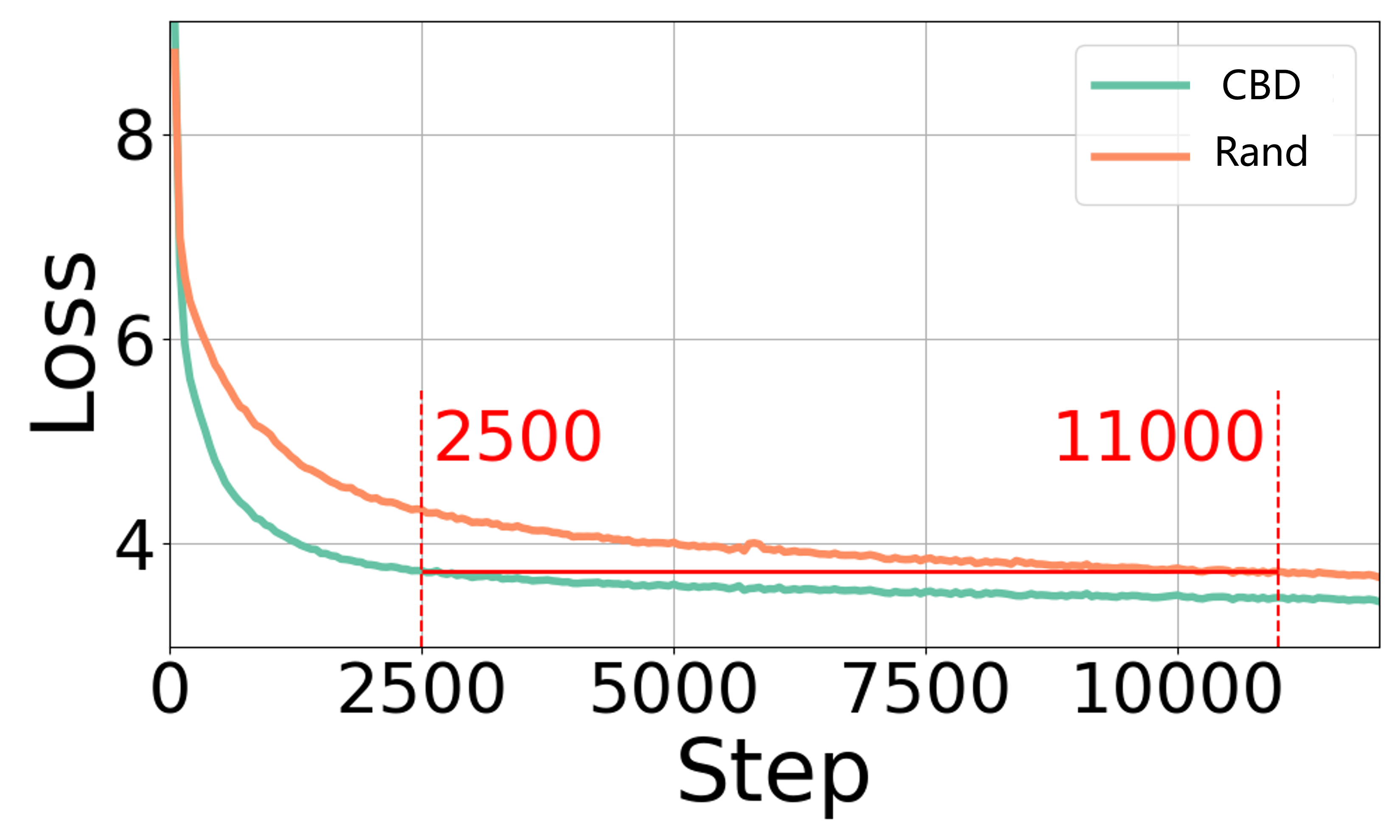}
        \caption{SLM-220M/500M T}
    \end{subfigure}
    \begin{subfigure}[b]{0.24\linewidth}
        \includegraphics[width=\linewidth]{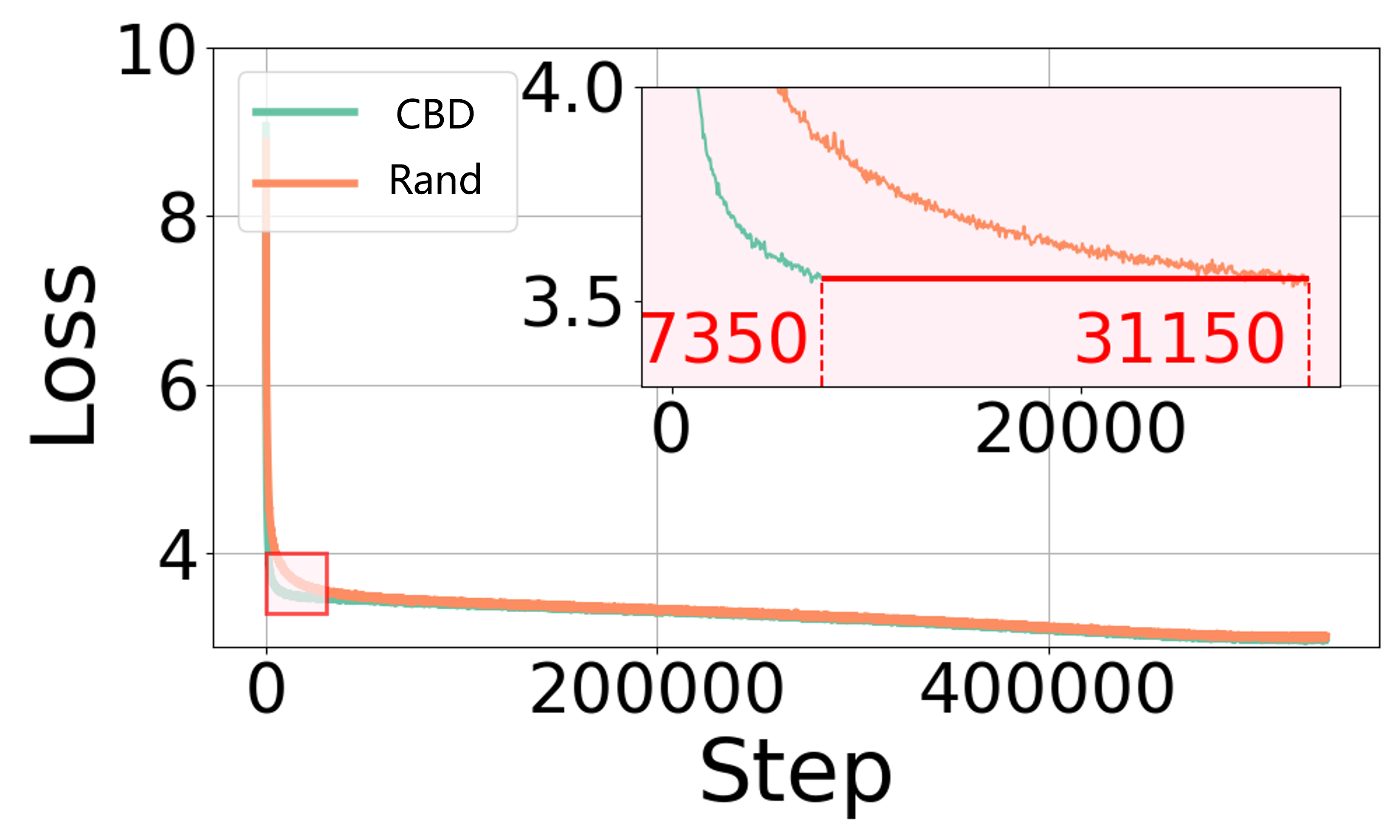}
        \caption{SLM-220M/10B T}
    \end{subfigure}
    \begin{subfigure}[b]{0.24\linewidth}
        \includegraphics[width=\linewidth]{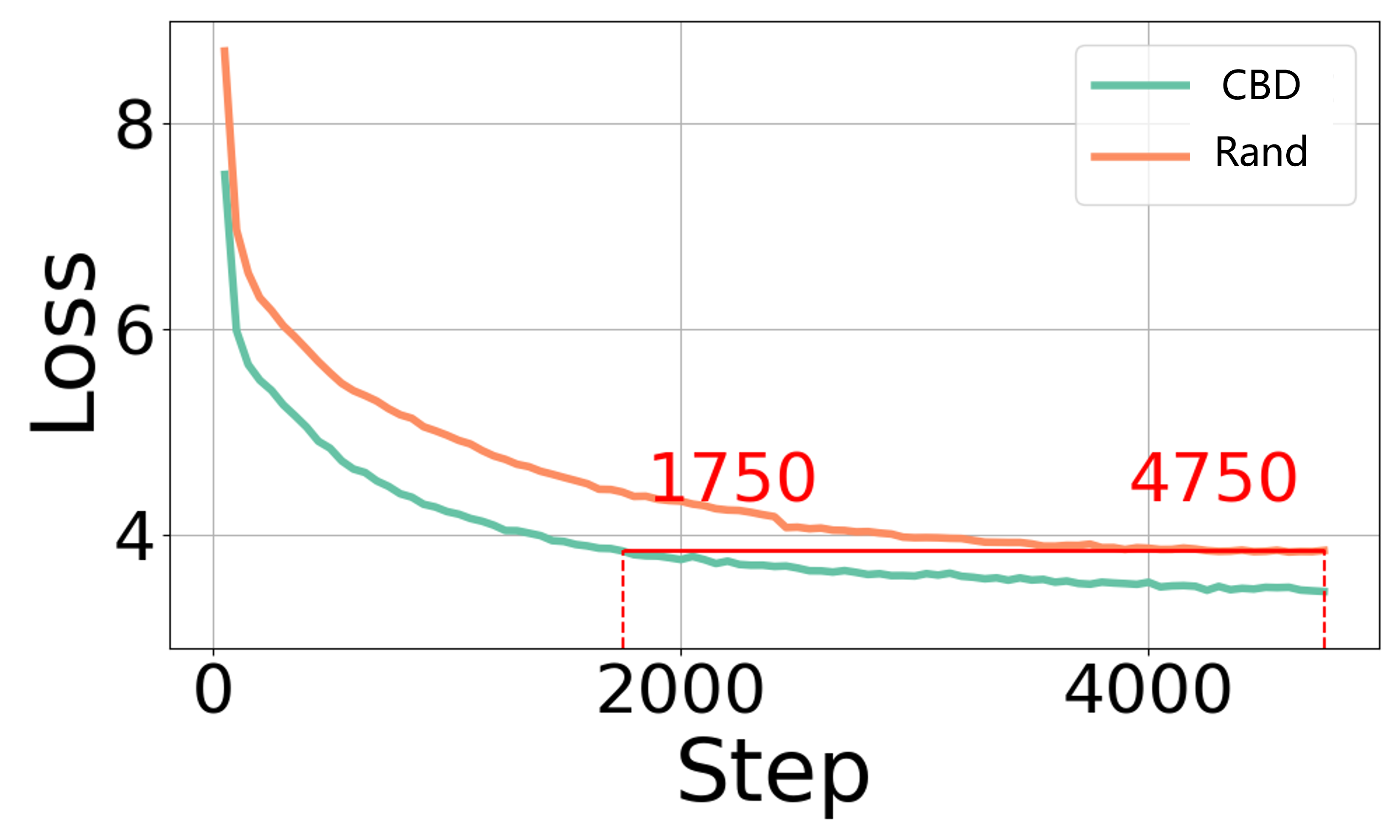}
        \caption{SLM-277M/78M T}
    \end{subfigure}
    \begin{subfigure}[b]{0.24\linewidth}
        \includegraphics[width=\linewidth]{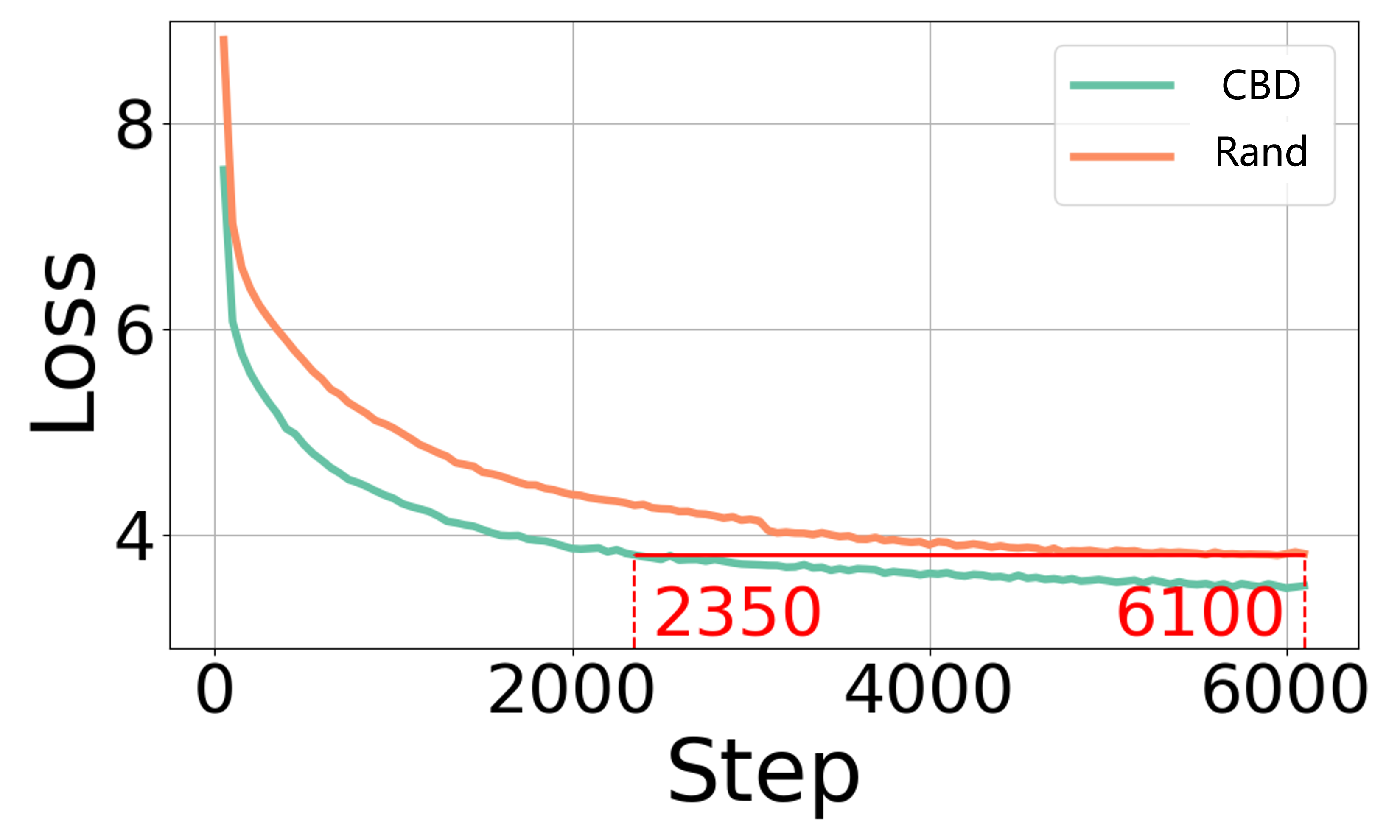}
        \caption{SLM-277M/100M T}
    \end{subfigure}
    \begin{subfigure}[b]{0.24\linewidth}
        \includegraphics[width=\linewidth]{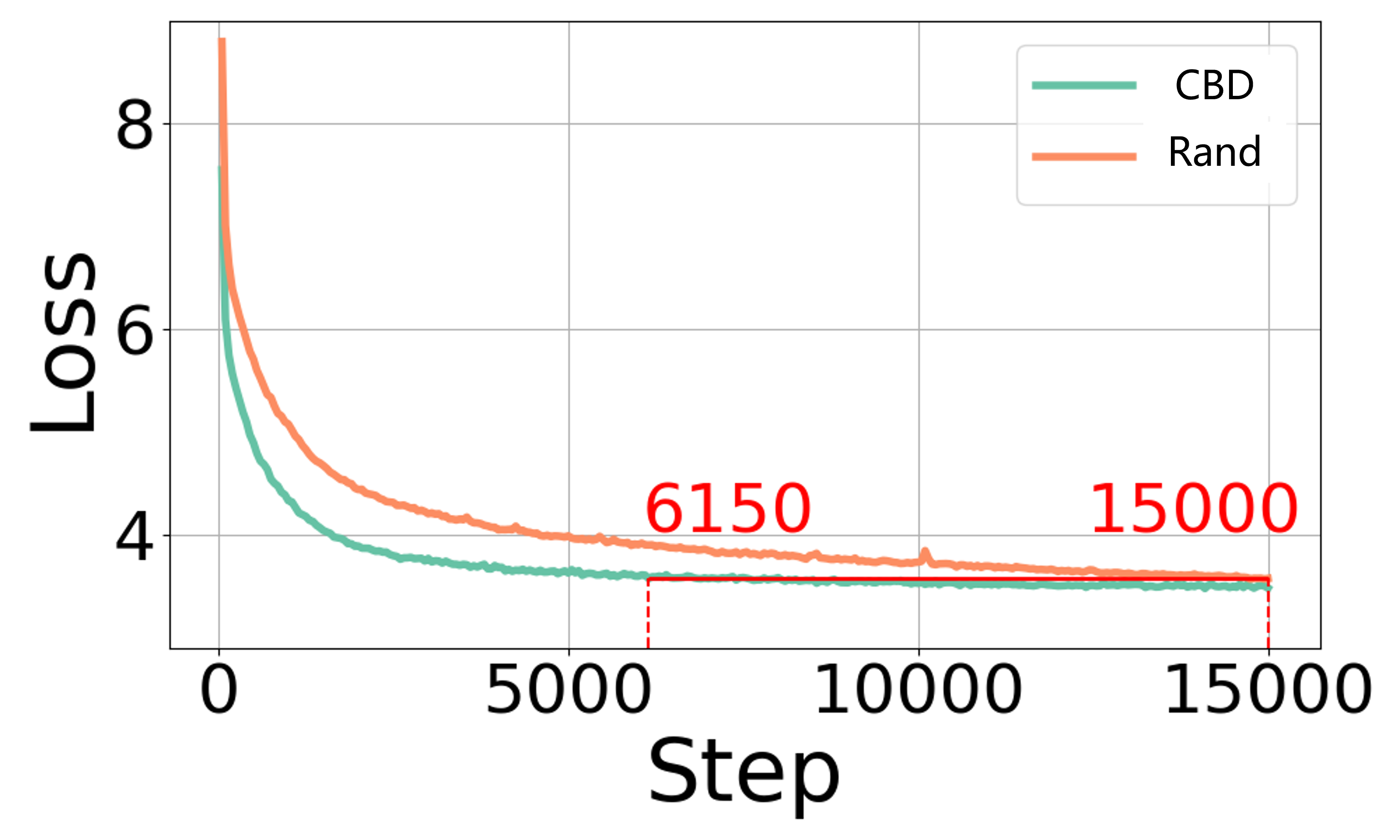}
        \caption{SLM-277M/500M T}
    \end{subfigure}
    \begin{subfigure}[b]{0.24\linewidth}
        \includegraphics[width=\linewidth]{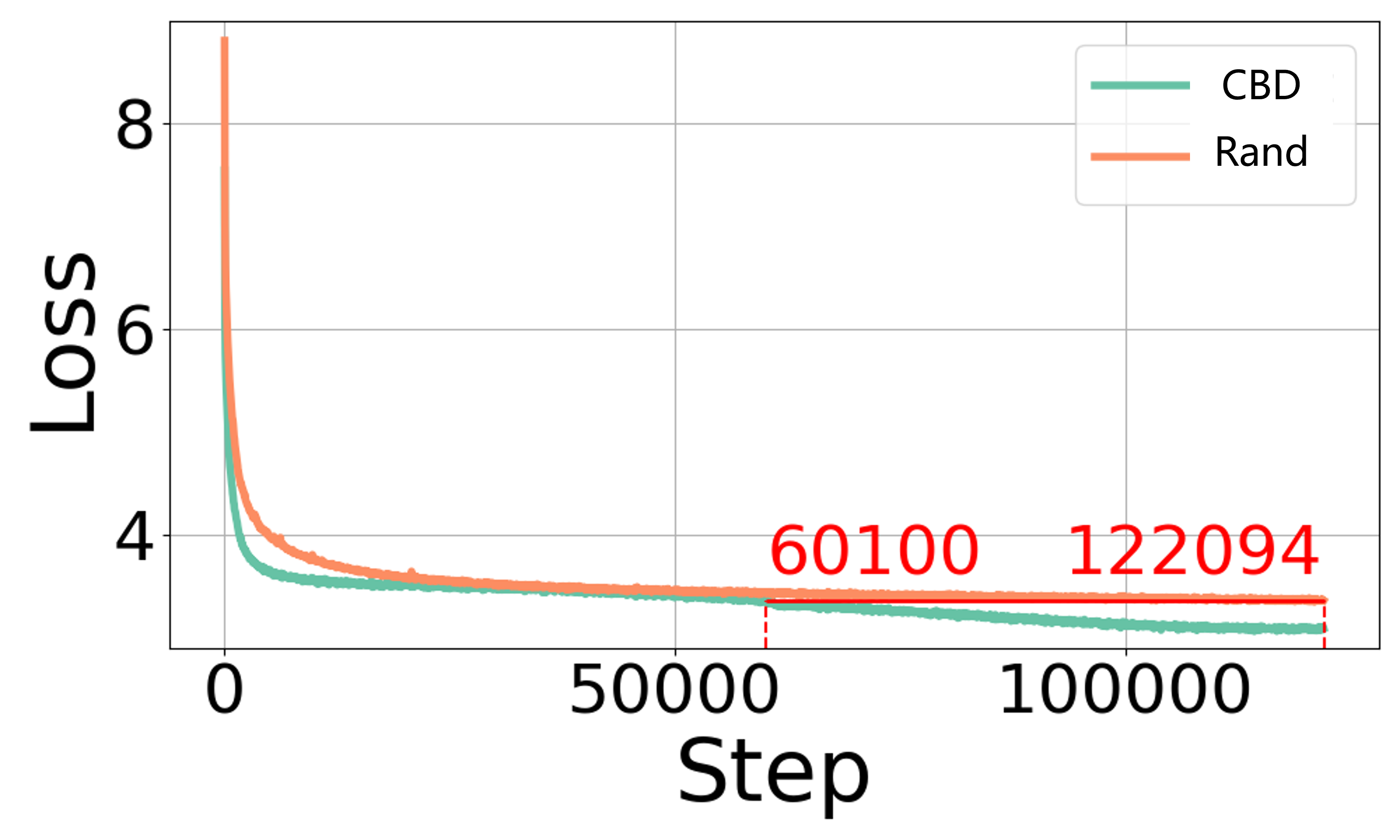}
        \caption{SLM-277M/10B T}
    \end{subfigure}
    \begin{subfigure}[b]{0.24\linewidth}
        \includegraphics[width=\linewidth]{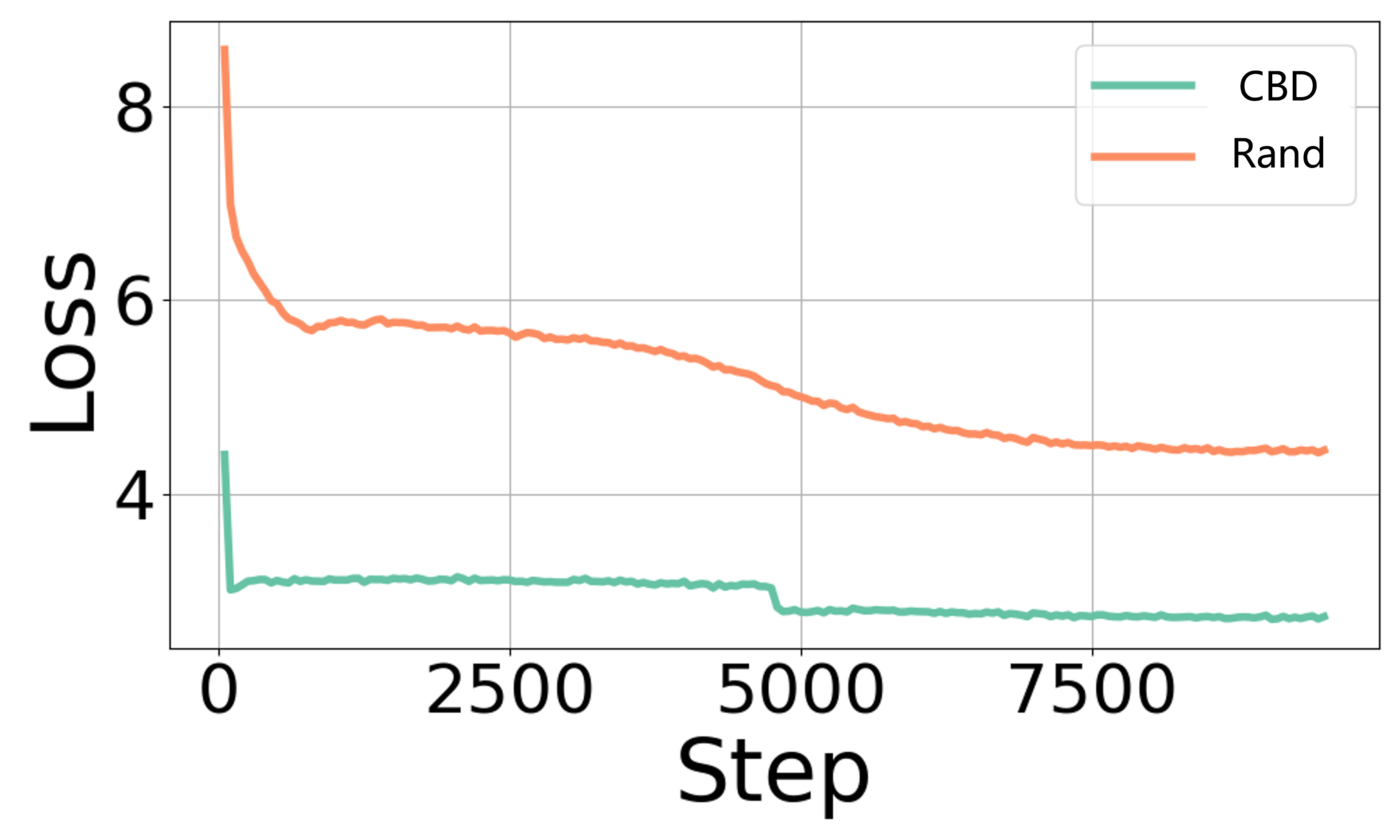}
        \caption{SLM-380M/78M T}
    \end{subfigure}
    \begin{subfigure}[b]{0.24\linewidth}
        \includegraphics[width=\linewidth]{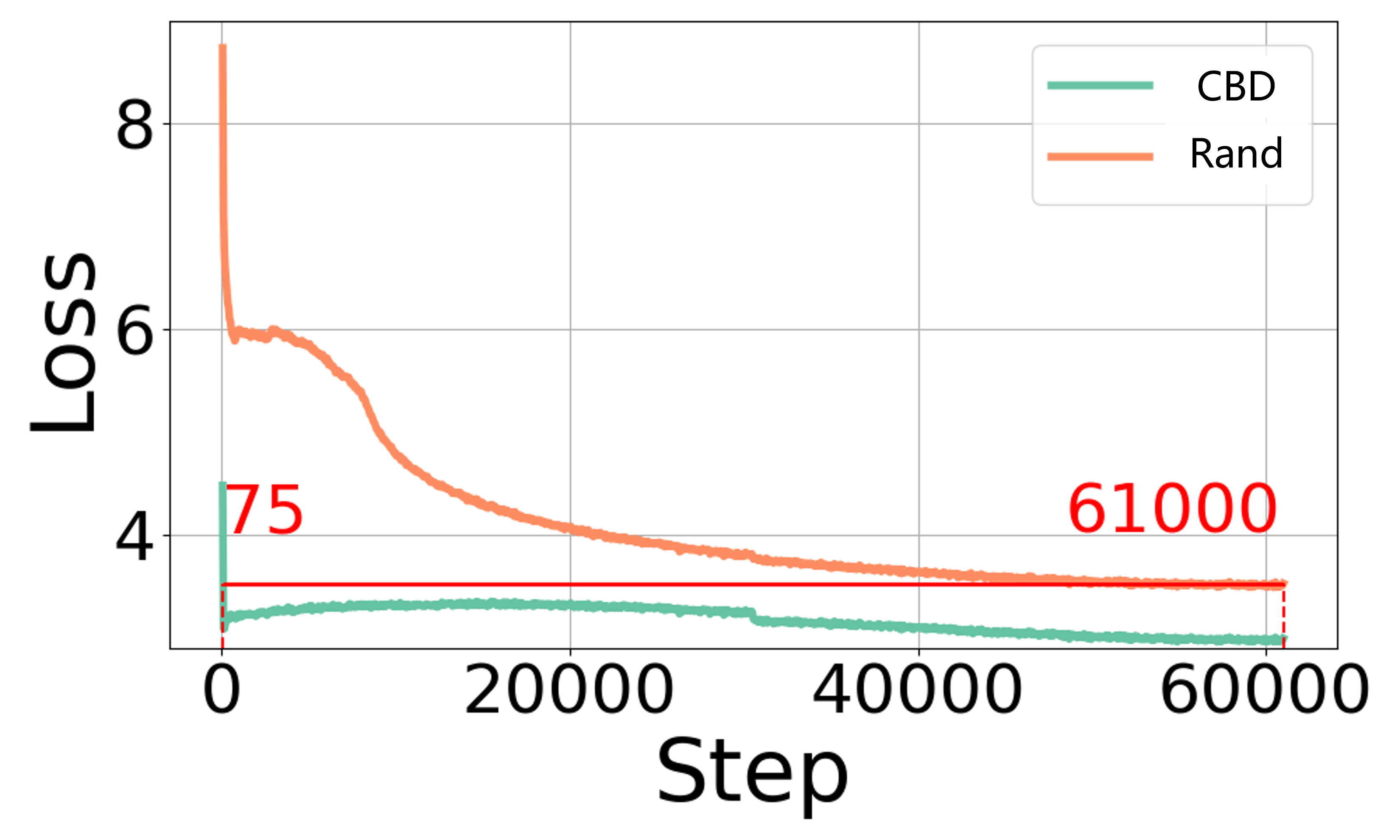}
        \caption{SLM-380M/500M T}
    \end{subfigure}
    \begin{subfigure}[b]{0.24\linewidth}
        \includegraphics[width=\linewidth]{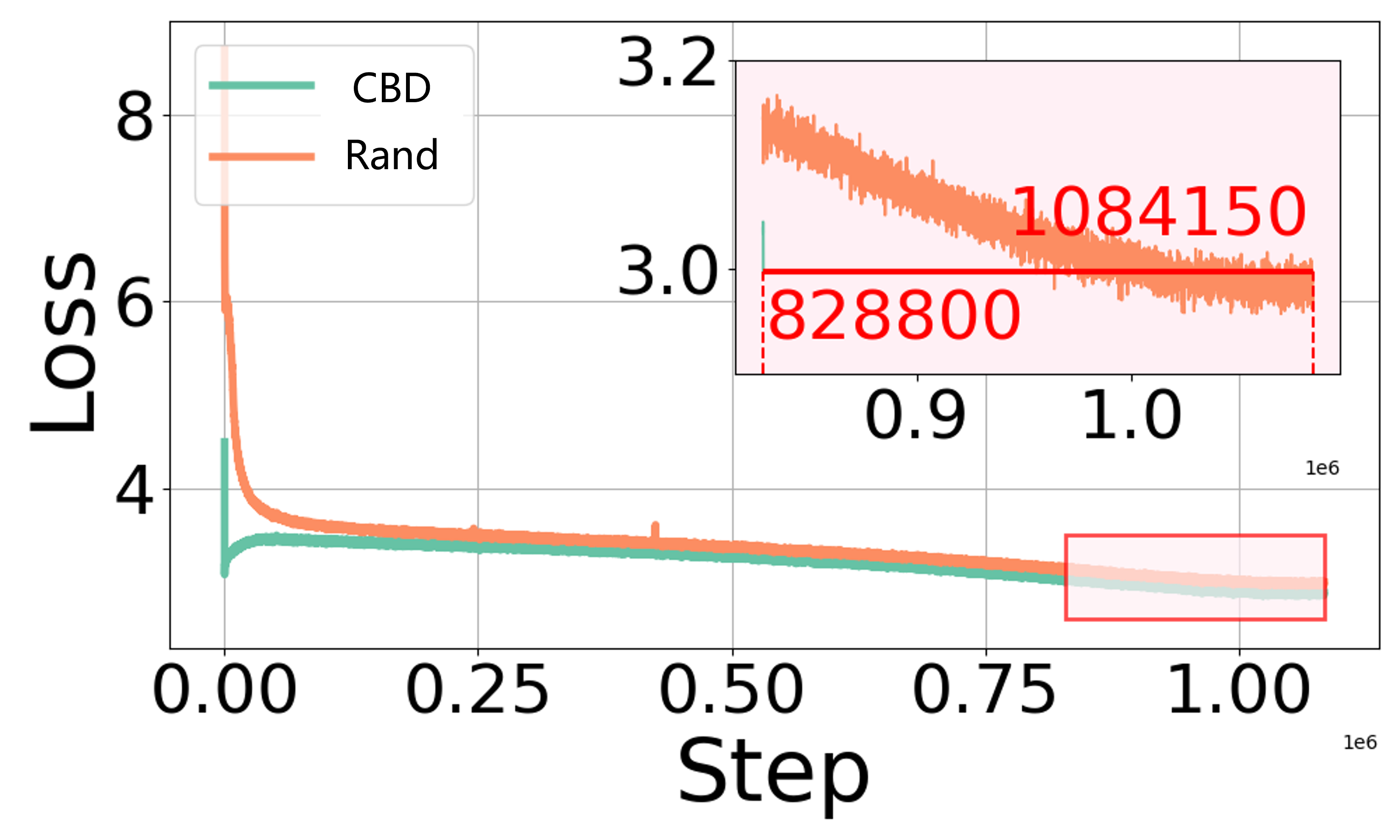}
        \caption{SLM-380M/10B T}
    \end{subfigure}
    \begin{subfigure}[b]{0.24\linewidth}
        \includegraphics[width=\linewidth]{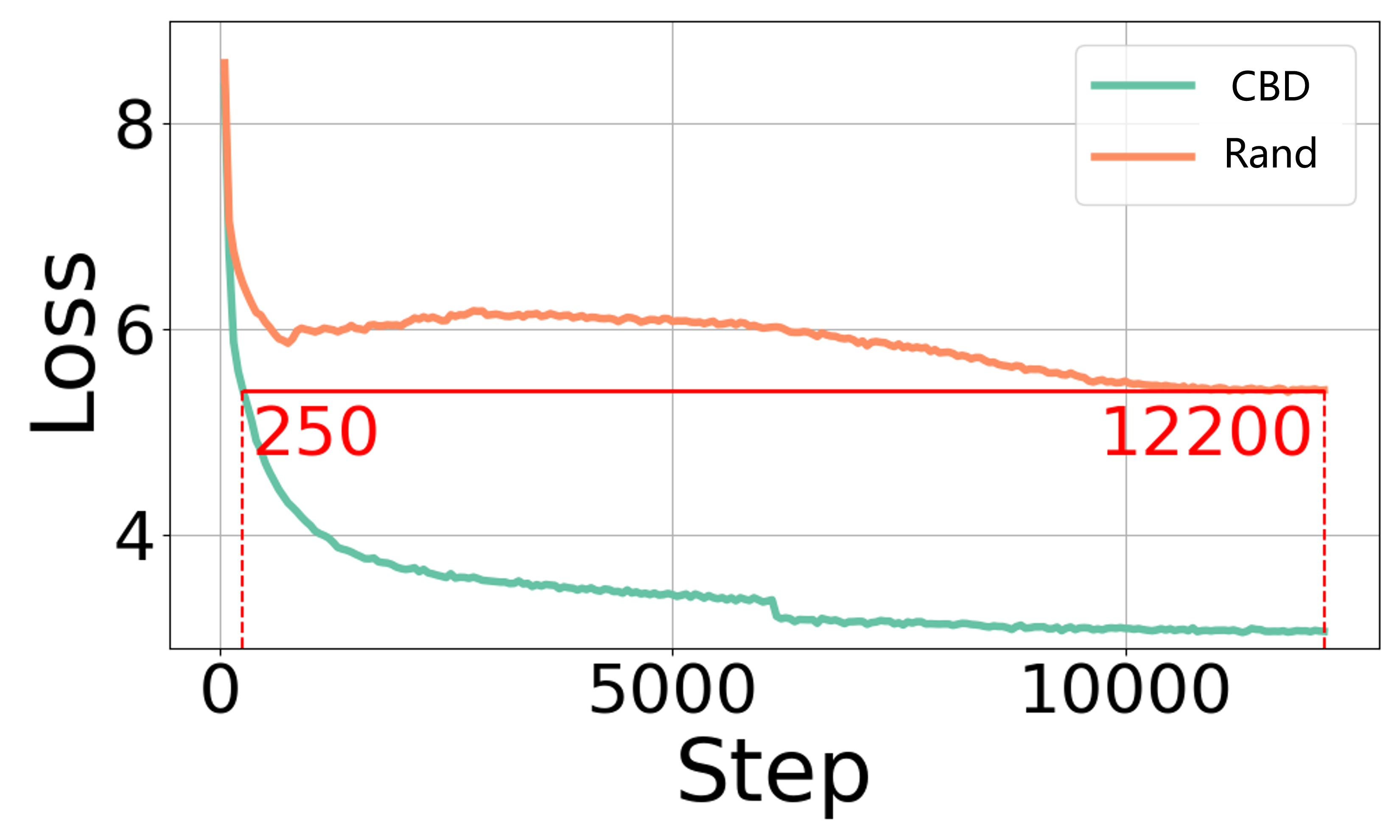}
        \caption{DN-537M/100M T}
    \end{subfigure}
    \caption{Convergence speed of variable-sized SLMs initialized by CBD and from scratch (Rand) on varying pre-training tokens across the source LLM: GPT2-XL. `T' means Token.}
    \label{fig:convergence_gptx_xl_2}
\end{figure}

In this section, we first compare the convergence speed of SLM-138M initialized by CBD and from scratch (Rand) on 100M pre-training tokens across three source LLMs: GPT2-XL, Llama3-8B, and Qwen3-4B, as shown in \cref{fig:convergence_100M}.
Then, we show the the convergence speed of SLM-138M initialized by CBD and Rand on 500M, 10B pre-training tokens, and the results are shown in \cref{fig:convergence_gptx_xl_1}.
Finally, we further show the convergence speed of SLM-220M, SLM-277M, SLM-380M, and SLM-537M initialized by CBD and Rand on 78M, 100M, 500M, 10B pre-training tokens across the source LLM: GPT2-XL. Both results are shown in \cref{fig:convergence_gptx_xl_2}.

From the both results, we can obviously observe that CBD has faster convergence speed than pre-training from scratch. 
For example, CBD achieves the same loss in only 140 and 5000 steps, compared to 30500, and 250200 steps for Rand on 500M, and 10B tokens, respectively, leading to a speedup of 80.26×, 217.86×, and 50.04×, as shown in \cref{fig:convergence_gptx_xl_1}. 
Besides, for SLM-220M, CBD reaches the same loss level using 1350, 1700, 2500, and 7350 steps under 78M, 100M, 500M, and 10B token budgets, respectively, while Pre requires 4750, 6100, 11000, and 31150 steps, as shown in \cref{fig:convergence_gptx_xl_2}. 
This corresponds to a convergence speedup of 3.52×, 3.59×, 4.4×, and 4.24×.
These results demonstrate that CBD consistently accelerates convergence across different token budgets and model sizes, with particularly significant gains in the low-token regime. 

\section{Performance Gap with the Increase of the Pre-training Token}
\label{app_performance_gap}
\begin{figure}[t]
\centering
\begin{minipage}{0.32\textwidth}
    \centering
    \begin{subfigure}{\linewidth}
        \includegraphics[width=\linewidth]{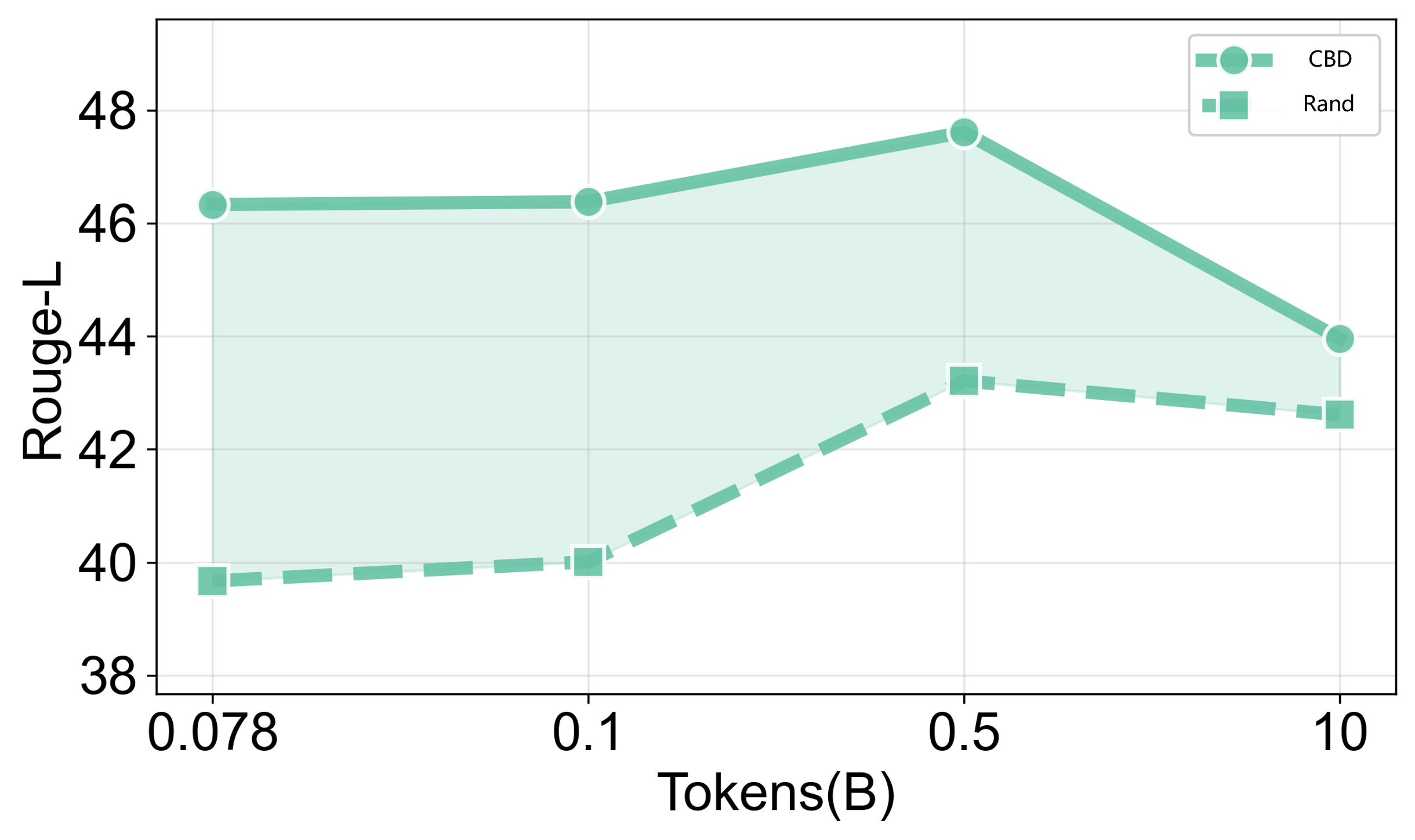}
        \caption{SLM-138M}
    \end{subfigure}
\end{minipage}
\begin{minipage}{0.32\textwidth}
    \centering
    \begin{subfigure}{\linewidth}
        \includegraphics[width=\linewidth]{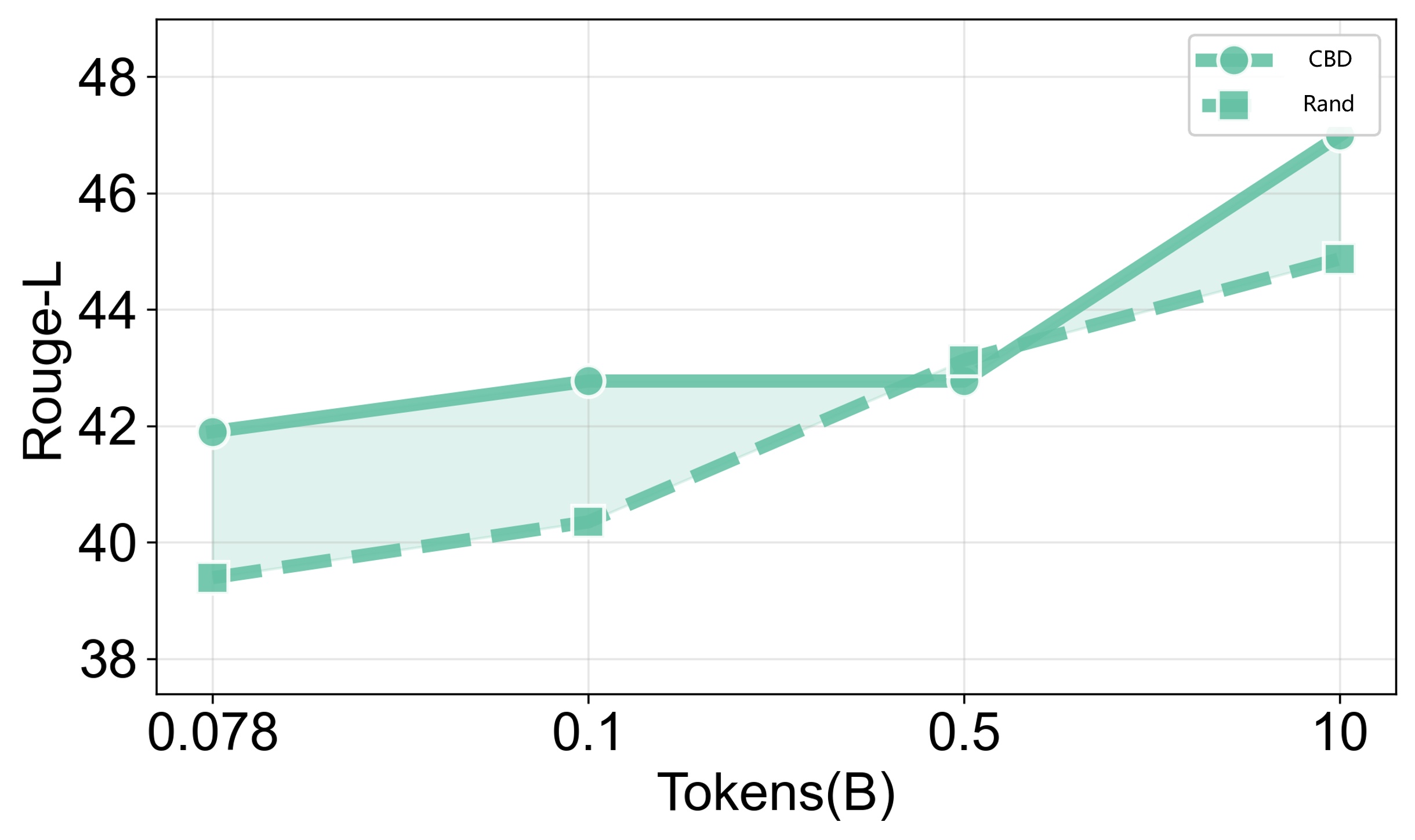}
        \caption{SLM-220M}
    \end{subfigure}
\end{minipage}
\begin{minipage}{0.32\textwidth}
    \centering
    \begin{subfigure}{\linewidth}
        \includegraphics[width=\linewidth]{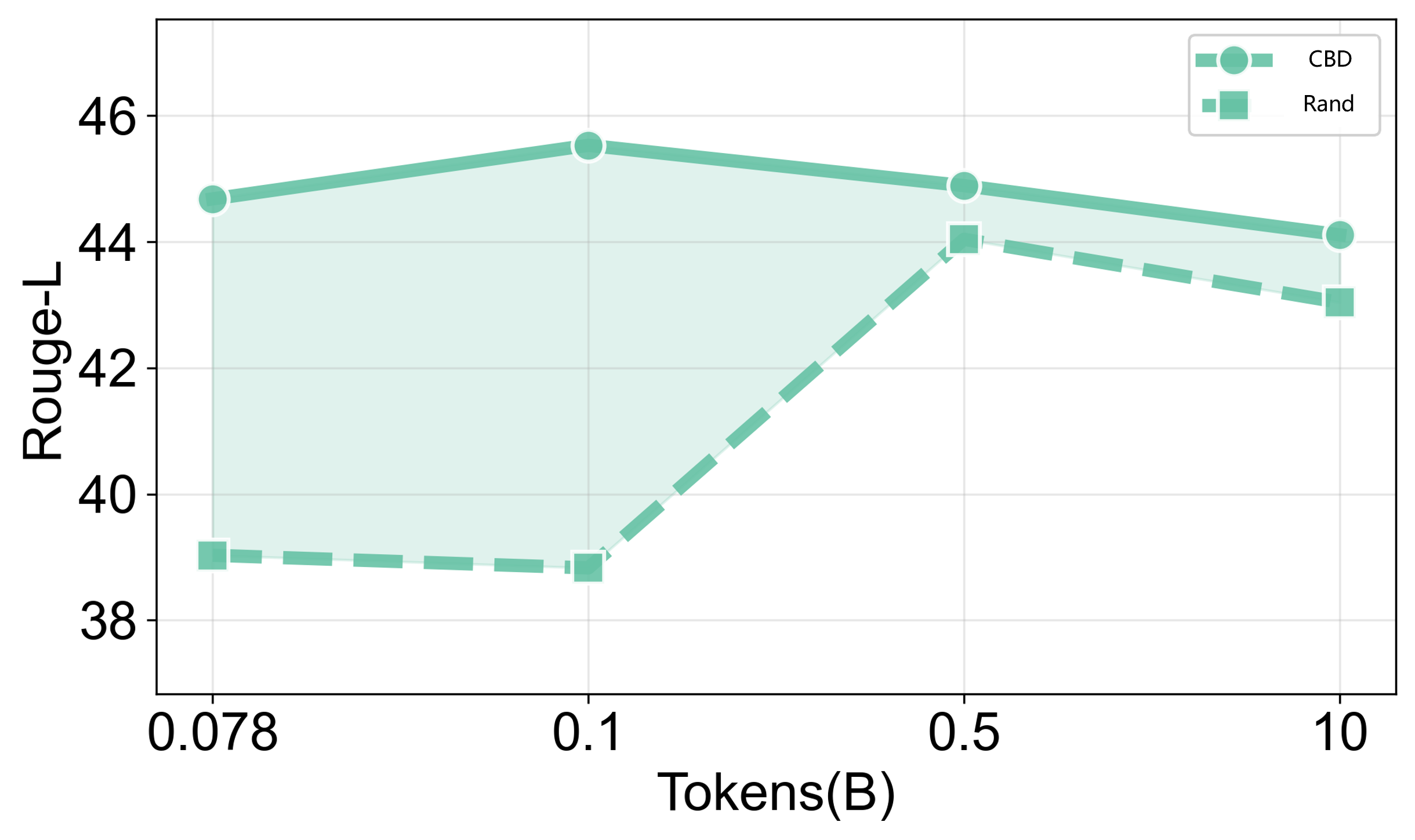}
        \caption{SLM-277M}
    \end{subfigure}
\end{minipage}
\begin{minipage}{0.32\textwidth}
    \centering
    \begin{subfigure}{\linewidth}
        \includegraphics[width=\linewidth]{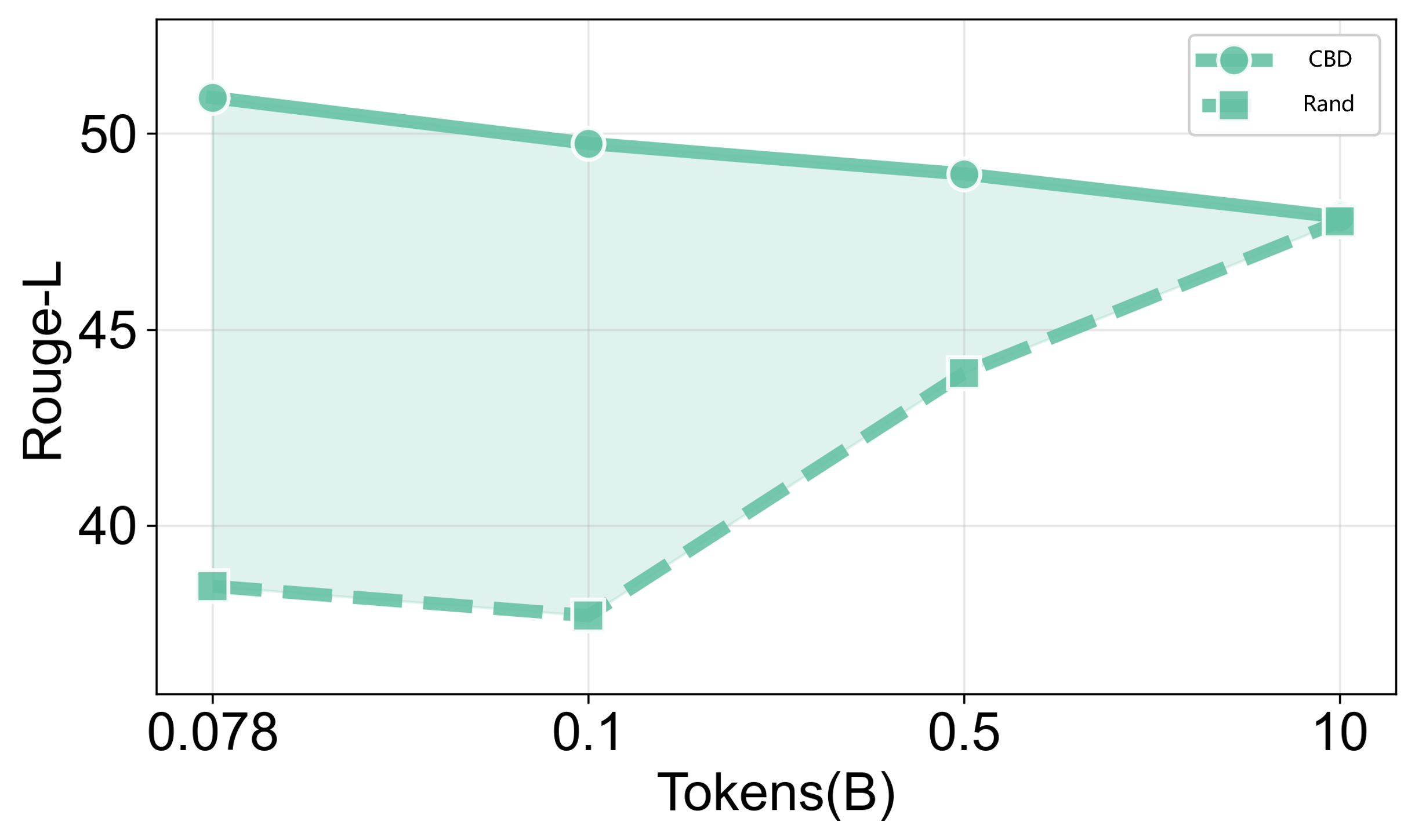}
        \caption{SLM-380M}
    \end{subfigure}
\end{minipage}
\begin{minipage}{0.32\textwidth}
    \centering
    \begin{subfigure}{\linewidth}
        \includegraphics[width=\linewidth]{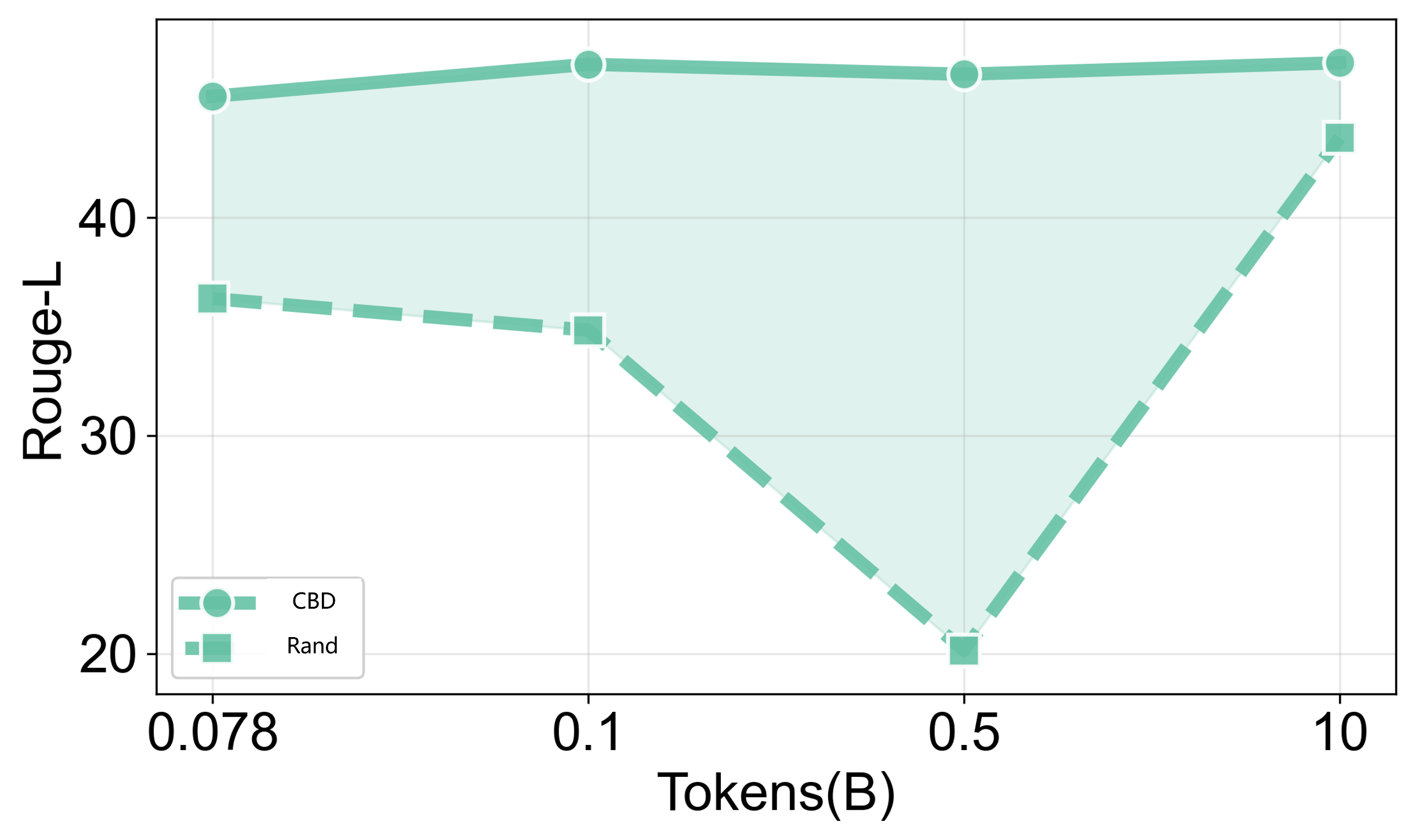}
        \caption{SLM-537M}
    \end{subfigure}
\end{minipage}
    \caption{Performance gap between the CBD and Rand on the Dolly task with the changing of the pretrain tokens.}
    \label{fig:token_acc_change}
\end{figure}

\cref{fig:token_acc_change} illustrates the performance change as the number of pre-training tokens increases. All SLMs are initialized from the source LLM GPT2-XL. The results show that the performance of pre-training from scratch (Rand) improves significantly with more pre-training data. This improvement is particularly substantial for SLMs with 380M and 537M parameters when the pre-training data reaches 10B tokens. In contrast, the CBD method proposed in this paper maintains outstanding performance under the same number of tokens. These results further demonstrate the superiority of the proposed CBD method.

\section{The Challenges of Transferring the Learngene Method to LLMs}
\label{app_previous_learngene}
We compare the proposed initialization way from the CBD against the representative previous learngene method, such as Auto-Learngene \citep{wang2023learngene}, and Vanilla-Learngene \citep{wang2022learngene} methods. 

Auto-Learngene proposes to automatically identify key layers of source model for Transformer architectures by introducing a MetaNet to guide the selection process. In experiments, Auto-Learngene observes that the lower layers of Transformers usually exhibit stronger generalization ability. Therefore, it directly selects the first N layers of the Transformer and stacks randomly initialized layers on top to construct models with different architectures. In contrast, Van-Learngene identifies key layers of source model in CNNs based on gradient signals. Empirically, Van-Learngene finds that the higher layers of CNNs tend to contain richer semantic information, and thus it extracts the last three layers as learngene and stacks N randomly initialized layers before them to build multiple models. 

Since neither of these approaches was originally designed for LLMs, we migrate them into the LLM setting. Specifically, for Auto-Learngene, we extract the first three layers of source LLM as learngene and stack 0, 2, or 4 randomly initialized layers after them to construct three SLMs of different sizes. For Van-Learngene, we extract the last three layers of source LLM as learngene and stack 0, 2, or 4 randomly initialized layers before them to construct three SLMs of different sizes. We use GPT2-XL as source LLM, because if Qwen3-4B or Llama3-8B were used, even extracting only one layer as learngene without stacking any random layers would already result in SLMs with over 500M and 1.4B parameters, respectively, which are unsuitable for our small language model setting. 

In addition, there exist other Learngene-based methods as introduced in the related work, such as Learngene Pool and SWS. These methods are both based on direct distillation with auxiliary models. In the ViT domain this strategy works because the capacity gaps are small, for example, DeiT-Base/ViT-Base (~86M) is distilled into 3M or 44M auxiliary models. However, directly transferring such designs to LLMs is infeasible. We fail to transfer those two methods into LLM, and the performance of the initialized models is even worse than that of the random initialization. This is due to the large gap between LLMs and small LMs. 
Moreover, they typically rely on multi-model distillation or staged expansion to form a learngene pool or modular structure. Such designs significantly increase computational costs and are difficult to scale to large LLMs. Therefore, in this study, we select Auto-Learngene and Van-Learngene as the most representative baseline methods to migrate and compare, as they both capture the core principles of learngene while aligning with our research objectives.

\section{Results of Qwen3-Based CBD}
\label{app_qwen3_col}
To further verify the generality of the proposed CBD methods, we conduct the experiments on Qwen3 \cite{yang2025qwen3} family. Here, we take Qwen3-4B as the source LLM, and the knowledge chain contains Qwen3-1.7B, Qwen3-0.6B. The results are shown in \cref{tab:col_qwen3}, and we find that Qwen3-based CBD still outperforms the random initialization on many tasks. This demostrates that the presented results and trends hold for more recent models like Qwen3.
\begin{table}[h]
\center
\caption{Performance comparison between the Qwen3-based CBD and Random initialization.}
\begin{tabular}{c|c|c|c|c}
\hline
       & BoolQ & Arc-E & WinoG & Avg            \\ \toprule
Rand & 64.94 & 25.26 & 43.80 & 44.67          \\ \midrule
CBD  & 65.76 & 25.44 & 44.44 & \textbf{45.21} \\ \bottomrule
\end{tabular}
\label{tab:col_qwen3}
\end{table}

\section{Comparison with HyperCloning and FSLM}
\label{app_compare_hyper_fslm}
In this section, we provide additional analysis comparing CBD with two related works, 
HyperCloning \cite{samragh2024scaling} and Stacked Small Language Models (FSLM) \cite{liang2024stacking}. 

\textbf{The difference between CBD and HyperCloning, FSLM}.
Although these methods share the broad goal of enhancing small language models (SLMs), 
their objectives, system designs, and scalability properties differ fundamentally from CBD. HyperCloning focuses on accelerating the training of a \emph{single} large model by progressively widening its architecture. FSLM constructs a multi-component system composed of several SLMs, but it does not offer a mechanism to initialize a large number of SLMs at multiple scales. In contrast, CBD is explicitly designed to generate and initialize \emph{multiple} SLMs of 
different sizes under strict compute and data budgets.

\textbf{Empirical Comparison with HyperCloning}.
To make the comparison favorable to HyperCloning, we follow its standard widening schedule and initialize the HyperCloning model from a strong CBD-initialized SLM-138M, rather than from random initialization. Using 500M pretraining tokens, we obtain HyperCloning models at two scales: 645M and 2.42B parameters. SLMs are constructed by CBD at comparable scales, and the results are shown in \cref{tab:hyper_vs_col}.

Across all fairness categories, CBD consistently outperforms HyperCloning, even when trained with significantly fewer tokens. This highlights CBD’s efficiency and its ability to transfer useful knowledge with minimal computation.
\begin{table}[h!]
\centering
\caption{Downstream performance comparison between HyperCloning and CBD under different scaling and data regimes.}
\begin{tabular}{cccccccc}
\toprule
Line & Method & Tokens & Param & BoolQ & MMLU & WinoG & Avg \\
\midrule
1 & HyperCloning & 500M & 2423M & 63.55 & 24.21 & 45.88 & 38.28 \\
2 & HyperCloning & 500M & 645M  & 63.90 & 23.99 & 46.47 & 38.00 \\
3 & CBD          & 500M & 277M  & 67.80 & 27.10 & 63.74 & 43.87 \\
4 & CBD          & 500M & 537M  & 67.80 & 25.30 & 62.30 & 43.69 \\
5 & CBD          & 100M & 537M  & 71.90 & 26.30 & 60.20 & 44.06 \\
\bottomrule
\end{tabular}
\label{tab:hyper_vs_col}
\end{table}

\textbf{Comparison with FSLM.}
We additionally compare CBD with the official results of FSLM (Stacked Small Language Models).
FSLM produces a 640M-parameter multi-SLM system using 5K supervised training samples. 
In contrast, CBD generates a single 138M SLMs (only 21.25\% of FSLM’s size) 
with approximately half as many SFT samples (2420 vs.\ 5000).

\begin{table}[h!]
\centering
\caption{Comparison between FSLM and SBD under strict resource constraints.}
\begin{tabular}{ccccc}
\toprule
Method & SFT Samples (K) & Param (M) & TinyARC/ARC & TinyMMLU/MMLU \\
\midrule
FSLM & 5.0 & 640  & 33.49 & 32.08 \\
CBD  & 2.139+0.285 & 138 (21.25\%) & 29.50 (88.08\%) & 28.20 (87.91\%) \\
\bottomrule
\end{tabular}
\label{tab:fslm}
\end{table}

Even under such a large scale and data disadvantage, CBD achieves 
\textbf{88.08\%} of FSLM’s performance on TinyARC and  
\textbf{87.91\%} on TinyMMLU. This shows that FSLM is not well suited for 
resource-limited scenarios, while CBD provides strong initialization quality 
even at very small model sizes.

Overall, these findings demonstrate that HyperCloning and FSLM differ fundamentally 
from CBD in goals and scalability, and CBD provides superior performance, 
faster convergence, and substantially better resource efficiency across a range 
of evaluation settings.

\section{CBD Performance Across Extreme Model Scales}
\label{app_extreme_scales}
To examine the applicability limits of CBD, we evaluate its performance on two extreme  model-size regimes: very small SLMs ($<$50M parameters), where representational capacity is highly constrained.

We construct a 51.47M-parameter descendant model (SLM-51.47M) and compare CBD 
initialization with random initialization under an identical 100M-token training budget. \cref{tab:col_50m} shows that CBD consistently outperforms random initialization across all downstream tasks, despite the extremely limited model capacity.

\begin{table}[h!]
\centering
\caption{Performance of SLM-51.47M initialized with CBD and random initialization under the same 100M-token budget.}
\begin{tabular}{lccccccc}
\toprule
Method & BoolQ & MMLU & WinoG & XSum & HellaS & Arc-E & Avg \\
\midrule
Rand & 51.55 & 24.29 & 41.23 & 33.11 & 35.56 & 27.36 & 35.52 \\
CBD  & 65.25 & 24.20 & 52.42 & 34.91 & 39.05 & 25.82 & 40.27 \\
\bottomrule
\end{tabular}
\label{tab:col_50m}
\end{table}

These results confirm that CBD remains effective even in the extremely small-model regime, providing substantial performance gains over random initialization.



\section{The Role of Bridge Distillation}
\label{app_role_bridge}
We compare heterogeneous bridge distillation with the homogeneous GPT2-based CBD on several generation-style benchmarks, including dolly \citep{DatabricksBlog2023DollyV2}, self-inst \cite{wang2023self}, sinst \cite{wang2022benchmarking}, uinst \cite{honovich2023unnatural}, and vicuna \cite{chiang2023vicuna}. The results for SLM-138M and SLM-380M are reported in \cref{tab:bridge_138m} and \cref{tab:bridge_380m}. Across all tasks and both model sizes, heterogeneous paths (Qwen3-4B or Llama3-8B as the source LLMs) achieve consistently higher scores than the GPT2-XL-based CBD.

These results differ from the multiple-choice benchmarks in the main paper, where homogeneous chains perform better. This contrast reflects the intrinsic properties of different task types. Multiple-choice tasks benefit more from architectural homogeneity and token-level alignment, whereas open-ended generation tasks rely more on high-level semantic priors that large modern source LLMs provide. Thus, heterogeneous bridge distillation is not weaker; it offers clear advantages in generative settings and highlights the task-dependent behavior of CBD.

\begin{table}[h]
\centering
\caption{Performance of bridge distillation on generation tasks for SLM-138M.}
\label{tab:bridge_138m}
\begin{tabular}{lccccc}
\toprule
 & dolly & self-inst & sinst & vicuna & Avg \\
\midrule
GPT2-XL      & 22.10 & 9.35  & 14.26 & 15.31 & 12.97 \\ 
Qwen3-4B     & 25.25 & 11.79 & 15.64 & 15.53 & \textbf{14.32} \\
Llama3-8B    & 22.83 & 10.99 & 15.48 & 15.86 & 14.11 \\
\bottomrule
\end{tabular}
\end{table}

\begin{table}[h]
\centering
\caption{Performance of bridge distillation on generation tasks for SLM-380M.}
\label{tab:bridge_380m}
\begin{tabular}{lccccc}
\toprule
 & dolly & self-inst & sinst & vicuna & Avg \\
\midrule
GPT2-XL      & 26.20 & 11.01 & 16.90 & 15.77 & 17.47 \\
Qwen3-4B     & 27.55 & 13.58 & 21.51 & 17.49 & \textbf{20.03} \\
Llama3-8B    & 26.63 & 12.95 & 21.65 & 16.75 & 19.49 \\
\bottomrule
\end{tabular}
\end{table}\textbf{}


\section{Limitations and Future Work}
\label{app_limit}

\textbf{Limitations.} In this work, we demonstrated the effectiveness of the proposed CBD framework primarily through averaged performance across multiple datasets. While CBD consistently achieves strong overall results, it does not yield the best performance on every individual task. This limitation arises because CBD emphasizes task-agnostic knowledge transfer and does not explicitly incorporate task-specific adaptations, which may lead to suboptimal outcomes in certain cases.

\textbf{Future Work.} A promising direction for future research is to enhance CBD with task-aware initialization strategies. By incorporating characteristics of the downstream application scenarios, CBD could selectively adapt and initialize parameters that are most sensitive to the target task. Such extensions have the potential to yield more comprehensive improvements in the performance of SLMs on specific tasks, further broadening the applicability of CBD.


\end{document}